\documentclass[review]{elsarticle}

\usepackage{hyperref}

\journal{Pattern Recognition}

\usepackage[tight,normalsize,sf,SF]{subfigure}
\usepackage{dsfont}

\usepackage[nodots]{numcompress}

\usepackage{lineno}
\modulolinenumbers[5]

\usepackage{url}
\usepackage{footnote}
\usepackage{color}
\usepackage{rotating}
\usepackage{multirow}
\usepackage{tabularx}
\usepackage{booktabs}
\newcommand{\myparagraph}[1]{\vspace{-0.0cm}\paragraph{\textbf{#1}}}

\def\R{\mathds{R}}
\def\S{$\mathcal{\mathbf{S}}$}
\def\A{$\mathbf{A}_i$}
\def\p{$\mathbf{p}_i$}

\newcommand{\nrd}{\textit{NRD}}
\newcommand{\scapeModelInit}{\textit{S-SCAPE + NRD}}
\newcommand{\jain}{\textit{S-SCAPE}}
\newcommand{\scapeModelInitcw}{\textit{S-SCAPE + NRD CW}}
\newcommand{\WSX}{\textit{WSX}}
\newcommand{\NH}{\textit{NH}}
\newcommand{\ours}{\textit{ours}}

\def\Rotation{\textbf{R}}

\def\Deformation{\textbf{Q}}
\def\Pose{\chi}
\def\Mesh{\textbf{M}}
\def\TemplateMesh{\textbf{T}}
\def\TriangleIndex{m}
\def\ScanIndex{i}

\def\Rmi{\Rotation_{\TriangleIndex,\ScanIndex}}
\def\Qmi{\Deformation_{\TriangleIndex,\ScanIndex}}
\def\Cmi{\textbf{C}_{\TriangleIndex,\ScanIndex}}
\newcommand{\SO}[1]{\mathrm{SO}(#1)}
\def\Shape{\varphi}
\def\AverageMesh{\overline{\Mesh}}
\def\PersonalizedMesh{{\Mesh}_{\Shape,\Pose_0}}
\def\PCAMat{\textbf{C}}
\def\PCADim{D}
\def\NumberVertices{N}
\def\NumberBones{B}

\begin{document}

\begin{frontmatter}

\title{Building Statistical Shape Spaces \\for 3D Human Modeling}

%% Group authors per affiliation:
%\author{Leonid~Pishchulin\fnref{myfootnote}}
%\address{Radarweg 29, Amsterdam}
%\fntext[myfootnote]{}

%% or include affiliations in footnotes:
\author[mymainaddress]{Leonid~Pishchulin\corref{mycorrespondingauthor}}
\cortext[mycorrespondingauthor]{Corresponding author}
%\ead[url]{www.elsevier.com}

\author[mysecondaryaddress]{Stefanie~Wuhrer}
\author[myotheraddress]{Thomas~Helten}
\author[mymainaddress]{Christian~Theobalt}
\author[mymainaddress]{Bernt~Schiele}

%\ead{support@elsevier.com}

\address[mymainaddress]{Max Planck Institute for Informatics, Germany}
\address[mysecondaryaddress]{INRIA Grenoble Rh\^{o}ne-Alpes, France}
\address[myotheraddress]{GoalControl GmbH, Germany}

\begin{abstract}
  Statistical models of 3D human shape and pose learned from scan
  databases have developed into valuable tools to solve a variety of
  vision and graphics problems. Unfortunately, most publicly available
  models are of limited expressiveness as they were learned on very
  small databases that hardly reflect the true variety in human body
  shapes. In this paper, we contribute by rebuilding a widely used
  statistical body representation from the largest commercially
  available scan database, and making the resulting model available to
  the community (visit \emph{\url{http://humanshape.mpi-inf.mpg.de}}). As
  preprocessing several thousand scans for learning the model is a
  challenge in itself, we contribute by developing robust best
  practice solutions for scan alignment that quantitatively lead to
  the best learned models. We make implementations of these
  preprocessing steps also publicly available. We extensively evaluate
  the improved accuracy and generality of our new model, and show its
  improved performance for human body reconstruction from sparse input
  data.
\end{abstract}


\end{frontmatter}

%\linenumbers

\section{Introduction}

Statistical human shape models represent variations in human physique
and pose using low-dimensional parameter spaces, and are valuable
tools to solve difficult vision and graphics problems, e.g. in pose
tracking or animation.  Despite significant progress in modeling the
statistics of the complete 3D human shape and
pose~\cite{AllenTSO03,AnguelovSCA05,Guan2012,ChenTBH13,Neophytou2013,HaslerBM2009,Jain:2010:MovieReshape}
only few publicly available statistical 3D body shape spaces
exist~\cite{HaslerBM2009,Jain:2010:MovieReshape}. Further on, the
public models are often learned on only small datasets with limited
shape variations~\cite{HaslerBM2009}. The reason is a lack of large
representative \textit{public} datasets and the significant effort
required to process and align raw laser scans prior to learning a
statistical shape space.

This paper contributes by systematically constructing a model of 3D
human shape and pose from the largest \textit{commercially} available
dataset of 3D laser scans~\cite{RobinetteTCP99} and making it publicly
available to the research community (Section
~\ref{sec:shape_space}). Our model is based on a simplified and
efficient variant of the SCAPE model~\cite{AnguelovSCA05} (henceforth
termed \emph{\jain~space}) that was described by Jain et
al.~\cite{Jain:2010:MovieReshape} and used for different applications
in computer vision and
graphics~\cite{Jain:2010:MovieReshape,pishchulin11bmvc,pishchulin12cvpr,HeltenPAE13,conf/cvpr/MundermannCA07},
but was never learned from such a complete dataset. This compact shape
space learns a probability distribution from a dataset of 3D human
laser scans. It models variations due to changes in identity using a
principal component analysis (PCA) space, and variations due to pose
using a skeleton-based surface skinning approach. This representation
makes the model versatile and computationally efficient.%. compared to
% SCAPE.

Prior to statistical analysis, the human scans have to be processed
and aligned to establish correspondence. We contribute by evaluating
different variants of the state-of-the-art techniques for non-rigid
template fitting and posture normalization to process the raw
data~\cite{AllenTSO03,HaslerBM2009,WuhrerPIS12,Neophytou2013}. Our
findings are not entirely new methods, but best practices and specific
solutions for automatic preprocessing of large scan databases for
learning the \jain~model in the best way
(Section~\ref{sec:data_processing}). First, shape and posture fitting
of an initial shape model to a raw scan prior to non-rigid deformation
considerably improves the results. Second, multiple passes over the
dataset improve initialization and thus increase the overall fitting
accuracy and statistical model qualities. Third, posture normalization
prior to shape space learning leads to much better generalization and
specificity.

The main contribution of our work is a set of \jain~spaces learned
from the largest database that is currently commercially
available~\cite{RobinetteTCP99}. The differences in our \jain~spaces
stem from differences in the registration and pre-alignment of the
human body scans. We evaluate different data processing techniques in
Section~\ref{sec:eval_template_fitting} and the resulting shape spaces
in Section~\ref{sec:eval_shape_space}. Finally, in
Section~\ref{sec:eval_reconstruction} we compare our \jain~spaces to
the state of the art \jain~space learned from a publicly available
database~\cite{HaslerBM2009} for the application of reconstructing
full 3D body models from partial depth data. Experimental evaluation
clearly demonstrates the advantages of our more expressive shape
models in terms of shape space quality and performance on the task of
reconstructing 3D human body shapes from partial depth observations.

We release the new shape spaces with code to (1) pre-process raw scans
and (2) fit a shape space to a raw scan for public usage. We believe
this contribution is required for future development in human body
modeling. Visit \emph{\url{http://humanshape.mpi-inf.mpg.de}} to
download code and models.

\subsection{Related work}

%\subsubsection{Datasets} 
\myparagraph{Datasets} Several datasets have been collected to analyze
populations of 3D human bodies. Many publicly available research
datasets allow for the analysis of shape and posture variations
jointly; unfortunately they feature data of at most 100
individuals~\cite{AnguelovSCA05,HaslerBM2009,Bogo:CVPR:2014}, which
limits the range of shape variations. We therefore use CAESAR
database~\cite{RobinetteTCP99}, the largest \emph{commercially}
available dataset to date that contains 3D scans of over $4500$
American and European subjects in a standard pose, as it represents
much richer sample of the human physique.

%\subsubsection*{Statistical shape spaces of 3D human bodies}
\myparagraph{Statistical shape spaces of 3D human bodies} Building
statistical shape spaces of human bodies is challenging, as there is
strong and intertwined 3D shape and posture variability yielding a
complex function of multiple correlated shape and posture
parameters. Methods to learn this shape space usually follow one of
two routes. The first group of methods learn shape- and
posture-related deformations separately and combine them
afterwards~\cite{AnguelovSCA05,Guan2012,ChenTBH13,Jain:2010:MovieReshape,Neophytou2013,Loper2015}. These
methods are inspired by SCAPE model~\cite{AnguelovSCA05} that couples
a shape space learned from variations in body shape with a posture
space learned from deformations of a single subject. This method has
recently been enhanced to capture deformations related to
breathing~\cite{Tsoli2014} and dynamic
motions~\cite{Pons-Moll2015}. Most SCAPE-like models use a set of
transformations per triangle to encode shape variations in a shape
space. Hence, to convert between the vertex coordinates of a processed
scan and its representation in shape space, a computationally
demanding optimization problem needs to be solved. To overcome this
difficulty, a simplified version of SCAPE model (\jain) was
proposed~\cite{Jain:2010:MovieReshape}. \jain~operates on vertex
coordinates directly and models pose variation with an efficient
skeleton-based surface skinning
approach~\cite{Jain:2010:MovieReshape,HeltenPAE13,conf/cvpr/MundermannCA07}. Recently,
two alternative multi-linear shape spaces have been proposed that also
operate directly on vertex coordinates~\cite{Neophytou2013,Loper2015}.

Another group of methods intends to perform simultaneous analysis of
shape and posture variations~\cite{AllenLAC06,HaslerBM2009}. These
methods learn skinning weights for corrective enveloping of
posture-related shape variations, which allows to explore both shape
and posture variations using a single shape space. Furthermore, it
allows for realistic muscle bulging as shape and posture are
correlated~\cite{NeumannEG2013}. It has been shown, however, that for
many applications in computer vision and graphics this level of detail
is not required and simpler and computationally more efficient shape
spaces can be
used~\cite{Jain:2010:MovieReshape,HeltenPAE13,pishchulin11bmvc,pishchulin12cvpr}.

%\subsubsection*{Mesh registration}
\myparagraph{Mesh registration}
Mesh registration is performed on the scans to bring them in
correspondence for statistical analysis.
Two surveys~\cite{vanKaickASO11,TamRO313} review such techniques, and a full
review is beyond the scope of this paper. Allen et
al.~\cite{AllenTSO03} use non-rigid template fitting
to compute correspondences between human body shapes in similar
posture. This technique has been extended to work for varying
postures~\cite{AnguelovSCA05,AllenLAC06,HaslerBM2009} and in
scenarios where no landmarks are
available~\cite{WuhrerLFP11}. In this work, we evaluate a non-rigid
template fitting approach inspired by Allen et al.~\cite{AllenTSO03}.

%\subsubsection*{Applications} 
\myparagraph{Applications} 
Statistical spaces of human body shape and posture are applicable in
many areas including computer vision, computer graphics, and ergonomic
design; our new model that was learned on a large commercially
available dataset is beneficial in each of these
applications. Statistical shape spaces have been used to predict body
shapes from partial data, such as image sequences and depth
images~\cite{Seo2006, balan07CVPR, Sigal2007, Xi2007a, GuanEHS09,
  HaslerMPA10, WeissH3D11, Boisvert2013, HeltenPAE13} and semantic
parameters~\cite{Seo2003, AllenTSO03, Chu2010, Baek2012, WuhrerE3D13,
  Rupprecht3DS13}. Furthermore, they have been used to estimate body
shapes from images~\cite{BalanTNT08} and 3D
scans~\cite{HaslerEBS09,Wuhrer14CVIU} of dressed subjects. Given a 3D
body shape, statistical shape spaces can be used to modify input
images~\cite{ZhouPRO10} or videos~\cite{Jain:2010:MovieReshape}, to
automatically generate training sets for people
detection~\cite{pishchulin11bmvc,pishchulin12cvpr}, or to simulate
clothing on people~\cite{Guan2012}.

\section{Statistical modeling with SCAPE}
\label{sec:shape_space}

We briefly recap the efficient version of the SCAPE
model~\cite{Jain:2010:MovieReshape} we build on and discuss its
differences to the original SCAPE model~\cite{AnguelovSCA05} in more
detail. For learning the model, both methods assume that a template
mesh $\TemplateMesh$ containing $\NumberVertices$ vertices has been
deformed to each raw scan in a database. All scans of the database are
assumed to be rigidly aligned, e.g. by Procrustes
Analysis~\cite{goodall91jrss}.

\subsection{Original SCAPE model}

In the original SCAPE model, the transformation of each triangle of
$\TemplateMesh$ is modeled as combination of three linear
transformations $\Rmi\in\SO{3}$ and $\Qmi \in\R^{3\times 3}$
controlling posture, and $\Cmi\in\R^{3 \times 3}$ controlling body
shape.  Index $\ScanIndex$ indicates one particular scan
$\TemplateMesh$ is fitted to. Fitting result after rigid alignment
with $\TemplateMesh$ is denoted as instance mesh $\Mesh_\ScanIndex$.

Shape deformations $\Cmi$ encode per-triangle deformations that can be
applied to change the body shape of the person in the same standard
posture. A low-dimensional space of plausible shape deformations
$\Cmi$ is computed by performing PCA on the training dataset captured
in standard posture.

To represent posture changes, two transformations are used: $\Rmi$
represents the posture of the person as rotation induced by the
deformation of an underlying rigid skeleton, and $\Qmi$ encodes the
individual deformations of each triangle that originates from varying
body shape or non-rigid posture dependent surface deformations such as
muscle bulging. Computing $\Qmi$ for each triangle separately is an
under-constrained problem. Therefore, smoothing is applied such that
$\Qmi$ of neighboring triangles become dependent. Finally, the
dimensionality of the transformations $\Rmi$ and $\Qmi$ is reduced
with the help of a kinematic chain model.

In this way, SCAPE obtains a flexible model that covers a wide range
of possible shape and posture deformations. However, as the model does
not explicitly encode vertex positions, a computationally expensive
optimization problem needs to be solved in order to reconstruct the
mesh surface.

\subsection{Simplified SCAPE (\jain) space}

The aforementioned computational overhead is often prohibitive in
applications where speed is more important than the overall
reconstruction quality, or when many samples need to be drawn from the
shape space. \jain~space~\cite{Jain:2010:MovieReshape} reconstructs
vertex positions in a given posture and shape without need of solving
a Poisson system. To learn the model, only laser scans in a standard
posture $\Pose_0$ are used. Meshes $\Mesh_\ScanIndex$ are used to
learn a PCA model that represents each shape using a parameter vector
$\Shape \in \R^\PCADim$ and can generate new models (represented in
homogeneous coordinates) with body shape $\Shape$ in posture $\Pose_0$
as
\begin{equation}
	\Mesh_{\Shape,\Pose_0} = \PCAMat \Shape + \AverageMesh.
	\label{eq:PCA-shape-space}
\end{equation}
$\PCAMat \in \R^{4 \NumberVertices \times \PCADim}$ with $\PCADim$ the
dimension of the PCA space is the matrix computed using PCA and
$\AverageMesh$ is the mean body shape of the training database.

This shape space only covers variations in body shape but not in
posture. To enable the latter an articulated skeleton is fitted to the
average human shape and linear blend skinning weights are used to
attach surface to bones. This allows to deform a body with fixed shape
$\Shape_0$ into an arbitrary posture $\Pose$ as
\begin{equation}
	\mathbf{p}_i\left(\Mesh_{\Shape_0,\Pose}\right) = \sum_{j=1}^{\NumberBones} w_{i,j}\Rotation_j(\Pose) \mathbf{p}_i\left(\Mesh_{\Shape_0,\Pose_0}\right),
	\label{eq:LBS}
\end{equation}
where $\mathbf{p}_i\left(\Mesh_{\Shape_0,\Pose}\right)$ is the
homogeneous coordinate of the $i$-th vertex of
$\Mesh_{\Shape_0,\Pose}$, $\NumberBones$ is the number of bones used
for the rigging, $\Rotation_j\in\R^{4\times 4}$ is the transformation
of the $j$-th bone, and $w_{i,j}$ are the rigging weights. We use the
rigging and skeleton consisting of $B=15$ bones proposed by Jain et
al.~\cite{Jain:2010:MovieReshape}. The skeleton is controlled by $30$
pose parameters corresponding to rigid transformations, joint angles
and scale. %This allows to express joint locations relative to nearby
%surface vertex locations.

For reconstructing a model of shape $\Shape$ in skeleton posture
$\Pose$, the method first calculates a personalized mesh
$\PersonalizedMesh$ using $\Shape$, and subsequently applies linear
blend skinning to the personalized mesh to obtain the final mesh
$\Mesh_{\Shape,\Pose}$. This can be expressed in matrix notation as
\begin{equation}
	\Mesh_{\Shape,\Pose} = \Rotation(\Pose) \PCAMat \Shape + \Rotation(\Pose) \AverageMesh,
	\label{eq:S-SCAPE-reconstruct}
\end{equation}
where $\Rotation(\Pose) \in \R^{4 \NumberVertices \times 4
  \NumberVertices}$ is a block-diagonal matrix containing the
per-vertex transformations. While decoupling of shape and posture
modeling by \jain~results in lower level of details
(e.g. posture-specific deformations such as muscle bulging may be
missing), it leads much faster reconstruction speed, especially when
the personalized mesh and skeleton can be precomputed. We argue that
in many applications speed may be much more important than the overall
reconstruction quality and build on this simple and efficient shape
space in this work.

\section{Data processing}
\label{sec:data_processing}

This section describes our pre-processing procedure that allows to
establish correspondences between raw laser scans. We demonstrate
best-practice solutions for non-rigid template fitting, effective
initialization strategies, introduce novel human-in-the-loop
bootstrapping approach that allows to improve the correspondences, and
finally explore postures normalization strategies. Tools to reproduce
these steps are made publicly available.

\subsection{Non-rigid template fitting}
\label{sec:nrd}
Our method to fit a human shape template $\TemplateMesh$ to a human
scan \S~is inspired by Allen et al.~\cite{AllenTSO03}. In non-rigid
template fitting (henceforth abbreviated \nrd), each vertex \p~of
$\TemplateMesh$ is transformed by a $4\times4$ affine matrix \A, which
allows for twelve degrees of freedom during the transformation. The
aim is to find a set of matrices \A~that align vertices of the
deformed template $\Mesh$ to the corresponding points of \S~in the
best possible way. The fitting is done by minimizing a combination of
data, smoothness and landmark errors.
 
%\subsubsection*{Data term} 
\myparagraph{Data term} 
The data term requires each vertex of the transformed template to be
as close as possible to its corresponding vertex of \S, and takes the form
\begin{equation}
  E_d=\sum_{i=1}^\NumberVertices
  w_i||\mathbf{A}_i\mathbf{p}_i-NN_i(\mathbf{S})||_{F}^2, 
	\label{eq:data-term}
\end{equation}
where $w_i$ weights the error contribution of each vertex, $||.||_{F}$
denotes the Frobenius norm, and $NN_i$ is a closest compatible point
in \S. If surface normals of closest points are less than $60^\circ$
apart and the distance between the points is less than $20$~mm, we set
$w_i$ to 1, otherwise to 0.

%\subsubsection*{Smoothness term} 
\myparagraph{Smoothness term} Fitting using $E_d$ only may lead to
situations where neighboring vertices of $\TemplateMesh$ match to
disparate vertices in \S. To enforce smooth surface deformations we
use a smoothness term $E_s$ that requires affine transformations
applied to connected vertices to be similar, i.e.
\begin{equation}
  E_s=\sum_{\{i,j|(\mathbf{p}_i,\mathbf{p}_j)\in \mathrm{edges}(\mathcal{\mathbf{T}})\}}||\mathbf{A}_i-\mathbf{A}_j||_{F}^2.
\end{equation}

%\subsubsection*{Landmark term} 
\myparagraph{Landmark term} Although using $E_d$ and $E_s$ would
suffice to fit two surfaces that are close to each other, the
optimization may get stuck in a local minimum when $\TemplateMesh$ and
\S~are far apart. A remedy is to identify a set of points on
$\TemplateMesh$ corresponding to known anthropometric landmarks on
\S. In each CAESAR scan these are obtained by placing markers on each
subject prior to scanning.  Our landmark term penalizes misalignments
between landmark locations
\begin{equation}
  E_l=\sum_{i=1}^M ||\mathbf{A}_{k_i}\mathbf{p}_{k_i}-\mathbf{l}_i||_{F}^2,
	\label{eq:landmark-term}
\end{equation}
where $k_i$ is the landmark index on $\TemplateMesh$, and
$\mathbf{l}_i$ is the landmark point on \S.  Although there are only
64 landmarks compared to the total number of $6449$ vertices, good
landmark fitting is enough to get the deformed surface of
$\TemplateMesh$ close to \S and avoid local convergence.

%\subsubsection*{Combined energy} 
\myparagraph{Combined energy} 
The three terms are combined into a single objective
\begin{equation}
  E=\alpha E_d + \beta E_s + \gamma E_l.
	\label{eq:combined_energy}
\end{equation}
For optimization we use L-BFGS-B \cite{zbMATH01235219}.  We vary the
weights $\alpha$, $\beta$ and $\gamma$ according to the following
empirically found schedule. We first perform a single iteration of
optimization without data term by setting $\alpha=0$, $\beta=10^6$,
$\gamma=10^{-3}$, which allows to bring the surfaces into a rough
correspondence. We then allow the data term to contribute by setting
$\alpha=1$, $\beta=10^6$, $\gamma=10^{-3}$. In addition, we relax
smoothness and landmark weights after each iteration of fitting to
$\beta:=0.25\beta$ and $\gamma:=0.25\gamma$, thus allowing the data
term to dominate. This is repeated until $\beta\le 10^3$. Reducing
$\beta$ increases the flexibility of deformation and allows
$\TemplateMesh$ to better reproduce fine details, while reducing
$\gamma$ is necessary due to unreliable placement of landmarks in some
scans.

\subsection{Initialization}

For non-rigid template fitting to succeed, $\TemplateMesh$ should be
pre-aligned to \S. We explore two initialization strategies.

%\subsubsection*{Static template} 
A first standard way to initialize \nrd~is to use a static template with annotated landmarks. Corresponding landmarks are then used to rigidly align \S~to $\TemplateMesh$.

%\subsubsection*{Simplified SCAPE space} 
A second way to initialize the fitting is to start with a \jain~space
that was learned from a small registered dataset. Fitting the shape
space to a scan is achieved by finding shape and posture parameters
$\Shape$ and $\Pose$ such that $\Mesh_{\Shape,\Pose}$ (see
Eq.~\ref{eq:S-SCAPE-reconstruct}) is close to \S. To this end, $E$
(Eq.~\ref{eq:combined_energy}) is minimized with respect to $\Shape$
and $\Pose$. To minimize $E$ depending on $\Shape$ and $\Pose$, we use
the vertices \p~of $\Mesh_{\Shape,\Pose}$, and set the per-vertex
deformations \A~to the identity. That is, the deformation of the body
shape $\Mesh_{\Shape,\Pose}$ is exclusively controlled by the
parameters $\Shape$ and $\Pose$. As in this case neighboring vertices
do not move independently, the term $E_s$ is not required, and we set
$\beta=0$.

To find a good local minimum, good initialization is required. We
found a two-step optimization approach to work well in
practice. First, we set $\alpha=0$ and $\gamma=1$ and optimize $E$
with respect to $\Pose$ while fixing $\Shape$, which fits the posture
of $\Mesh_{\Shape,\Pose}$ to \S~with the help of landmarks. Second, we
set $\alpha=1$ and $\gamma=0$ and optimize $E$ with respect to
$\Shape$ and $\Pose$ iteratively. For increased efficiency, each
iteration optimizes $E$ with respect to $\Pose$ in a first step and
with respect to $\Shape$ in a second step. After each iteration, the
set $NN_i(\mathbf{S})$ is recomputed. This iterative procedure is
repeated until $E$ does not change significantly. Iterative interior
point method is used for optimization.

\subsection{Bootstrapping}

%% In many cases, even after \nrd, $\TemplateMesh$ is far from \S. Using
%% registered scans with a high fitting error for shape space learning
%% may lead to unrealistic shape deformations in the learned space. A
%% remedy is to visually examine each fitting, discard fittings of low
%% quality, and learn a \jain~space using the samples that passed the
%% visual inspection.  This \jain~space is then used as
%% initialization to perform a fitting during the next pass. This
%% bootstrapping process is performed until nearly all registered scans
%% pass the visual inspection. Note that visual inspection is required,
%% as low average fitting errors do not always correspond to good
%% results, since the fitting of localized areas may be inaccurate.

In many cases, even after non-rigid template fitting (\nrd), fitted
mesh $\Mesh$ is far from the target human scan \S. Learning from
registered scans with a high fitting error may capture unrealistic
shape deformations. We thus propose the following human-in-the-loop
bootstrapping learning process: after each fitting iteration we
visually examine each registered scan, discard registered scans of low
quality, and learn a \jain~space using the registered scans that
passed the visual inspection; learned \jain~space is then used during
initialization of the next fitting pass and the process is
repeated. This bootstrapping process is performed for multiple
iterations until nearly all registered scans pass the visual
inspection. Note that visual inspection is required, as low average
fitting errors do not always correspond to good results, since the
fitting of localized areas may be inaccurate.

\subsection{Posture normalization}

The \jain~space used in this study decouples learning of shape and
posture variations and learns shape variations via PCA on the
registered scans captured in a standard posture. However even standard
postures may still contain slight posture variations, mostly due to
movements of arms. Thus PCA may learn global shape variations due to
variation in posture. In order to address this issue we perform
posture normalization of the registered scans based on two
approaches~\cite{WuhrerPIS12,Neophytou2013}, as explained in the
following.

Wuhrer et al.~\cite{WuhrerPIS12} factor out variations due to posture
changes by performing PCA on localized Laplacian coordinates.  While
this approach leads to better shape spaces, it is difficult to
directly apply this approach to the \jain~spaces learned using
Cartesian coordinates. We therefore compute a posture-normalized
version of each fitted mesh $\Mesh_\ScanIndex$ in the following way:
we start with a mean shape $\AverageMesh$ computed over all
$\Mesh_\ScanIndex$ and use~\cite{WuhrerPIS12} to optimize the
localized Laplacian coordinates of $\AverageMesh$ to be as close as
possible to $\Mesh_\ScanIndex$. This leads to fittings that have the
body shape of $\Mesh_\ScanIndex$ in the common normalized posture of
$\AverageMesh$.

Neophytou and Hilton~\cite{Neophytou2013} normalize the posture of
each processed scan using a skeleton model and Laplacian surface
deformation. While such normalization may introduce artifacts around
joints when the posture is changed significantly, this approach is
suitable to account for minor posture variations of CAESAR scans. We
use this method to modify the posture of each fitted mesh
$\Mesh_\ScanIndex$.

\section{Evaluation of template fitting}
\label{sec:eval_template_fitting}
We now evaluate different components of our registration procedure on
CAESAR dataset~\cite{RobinetteTCP99}. Each CAESAR scan contains $73$
manually placed landmarks. We exclude several landmarks located on
open hands, as those are missing for our template, resulting in $64$
landmarks used for registration. Furthermore, we remove all laser
scans without landmarks and corrupted scans, resulting in 4308 scans.

\subsection{Implementation details} 
Non-rigid template fitting requires a human shape template as input,
and the initialization procedure requires an initial shape space. We
use registered scans of 111 individuals in neutral posture of the MPI
Human Shape dataset~\cite{HaslerBM2009} to compute these
initializations.

However, MPI scans have artifacts such as spiky non-smooth surfaces in
the areas of head and neck. We smooth these areas by identifying
problematic vertices and by iteratively recomputing their positions as
an average position of direct neighbors.  Furthermore, due to privacy
reasons, head vertices of each human scan were replaced by the same
dummy head, which is not representative and of low quality at the
backside. We adjust the vertex compatibility criteria to compute
nearest neighbors during \nrd~by allowing $30^\circ$ deviation of the
head face normals while increasing the distance threshold to $50$~mm.

We employ the algorithm from Section~\ref{sec:nrd} to compute
correspondences for the CAESAR dataset. One inconsistency between the
datasets is that the hands in the MPI Human Shape dataset are closed,
while they are open in the CAESAR dataset. As remedy, we set $\alpha$
and $\gamma$ to zero for hand vertices in
Eq.~\ref{eq:combined_energy}, thus only allowing $E_s$ to
contribute. Prior to fitting, we sub-sample each CAESAR scan to have a
total number of vertices that exceeds the number of vertices of
$\TemplateMesh$ by a factor of three ($6449$ vertices in
$\TemplateMesh$ vs. $19347$ vertices in \S). This provides a good
trade-off between fitting quality and computational efficiency.

\subsection{Quality measure} 
Measuring the accuracy of surface fitting is not straightforward, as
no ground truth correspondence between \S~and $\TemplateMesh$ is
available. We evaluate the fitting accuracy by finding the nearest
neighbor in \S~for each fitted template vertex. If this neighbor is
not further than $50$ mm from its correspondence in $\TemplateMesh$
and its face normals do not deviate more than $60^\circ$, the
Euclidean vertex-to-vertex distance is computed. In our experiments we
report both the proportion of vertices falling below a certain
threshold and the distance per vertex averaged over all fitted
templates.  In the following, we first show the effects of various
types of initialization and weighting schemes in the \nrd~procedure on
the fitting error. Second, we show the effect of performing multiple
bootstrapping rounds.  %% \renewcommand{\subfigtopskip}{0pt}%
\begin{figure*}[tbp]
\centering
\bgroup
\tabcolsep 0.0pt
\renewcommand{\arraystretch}{0.0}
\subfigure[Total fitting error]{
\begin{tabular}{c}
\includegraphics[height=0.18\linewidth]{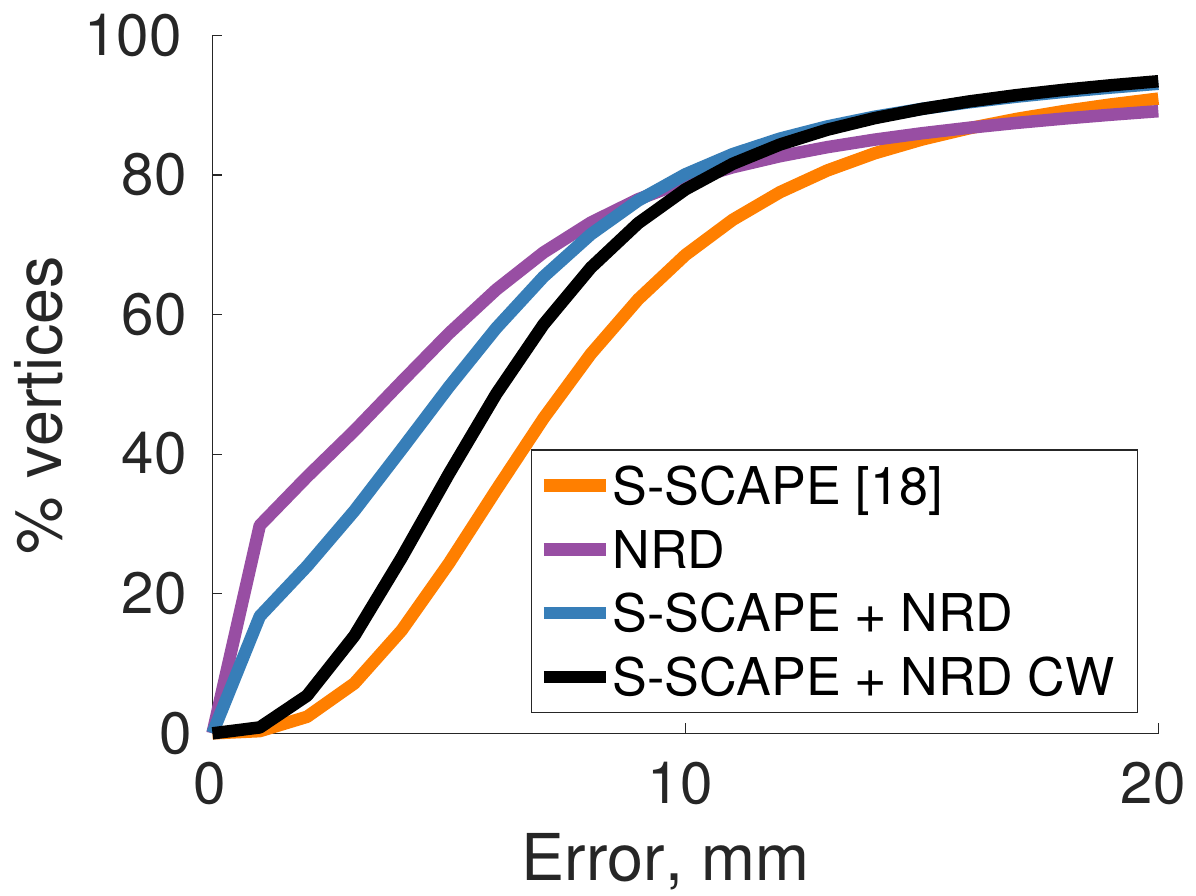}\\
\\
\vspace{0.8em}
\end{tabular}
}
\hspace{1em}
\subfigure[Mean fitting error]{
\tabcolsep 0.0pt
\begin{tabular}{ccccc}
\includegraphics[trim=10cm 0cm 0cm 0cm, clip=true, height=0.15\linewidth]{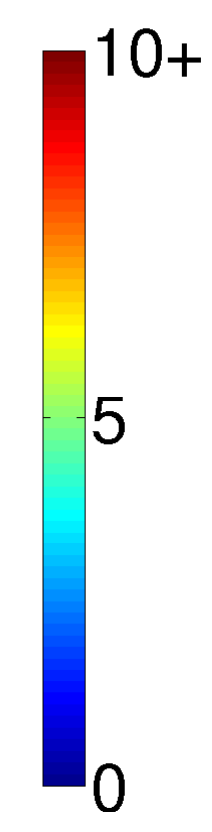}&\quad
\includegraphics[trim=0cm 1.8cm 5cm 1.8cm, clip=true, height=0.15\linewidth]{mpii-model-validnn-thresh1-0-thresh2-10-meanDist-45.pdf}&\quad
\includegraphics[trim=0cm 1.8cm 5cm 1.8cm, clip=true, height=0.15\linewidth]{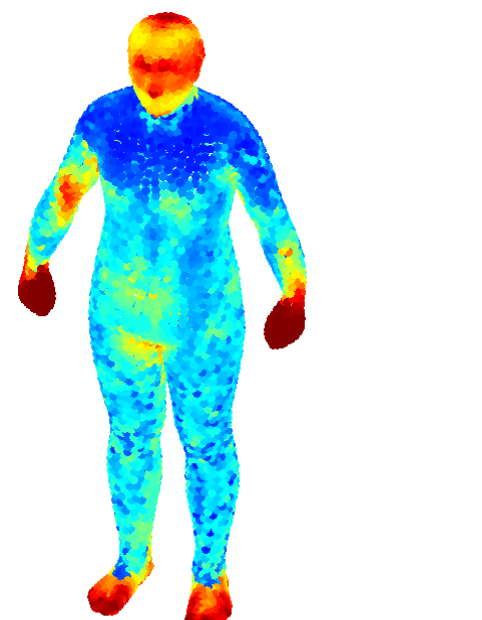}&\quad
\includegraphics[trim=0cm 1.8cm 5cm 1.8cm, clip=true, height=0.15\linewidth]{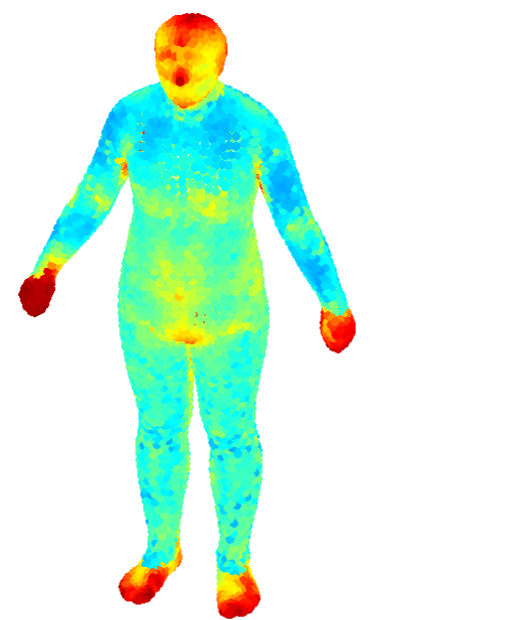}&\quad
\includegraphics[trim=0cm 1.8cm 5cm 1.8cm, clip=true, height=0.15\linewidth]{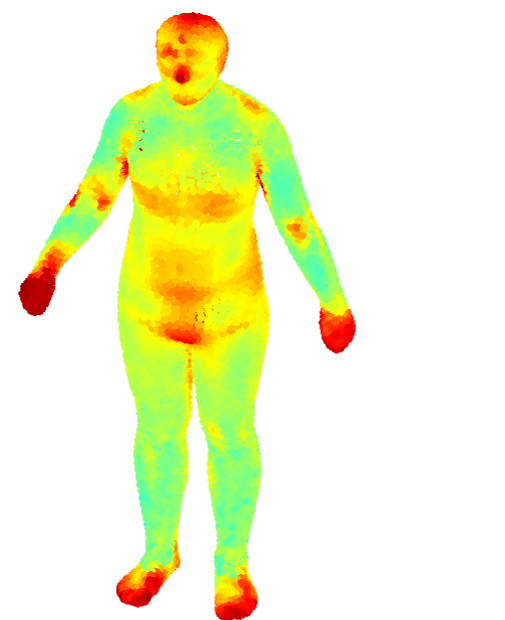}\\[10pt]
&\tiny \jain~\cite{Jain:2010:MovieReshape}& \tiny \nrd &\tiny \scapeModelInit&\tiny \scapeModelInitcw\\
\vspace{0.5em}
\end{tabular}
}
\caption{Fitting error on the CAESAR dataset when using the
  \jain~space~\cite{Jain:2010:MovieReshape} alone, \nrd~alone, and
  initializing using \jain~prior to \nrd~with different weighting
  schedules (\scapeModelInit, \scapeModelInitcw). Shown are (a) the
  proportion of vertices [\%] with fitting error below a threshold and
  (b) the average fitting error per vertex.}
  \label{fig:init-avg-dist}
\egroup
\vspace{-1.0em}
\end{figure*}

\subsection{Initialization} 
First, we evaluate two different initialization strategies used in our
fitting procedure. We compare the results when using an average human
template (\nrd) to the case when additionally using the \jain~space
learned on the MPI Human Shape dataset (\scapeModelInit) for
initialization. We also demonstrate the effects of both non-rigid
deformation schemes on the fitting accuracy and finally compare to the
results when using the publicly available \jain~space by Jain et
al.~\cite{Jain:2010:MovieReshape} alone (\jain).

The results are shown in Fig.~\ref{fig:init-avg-dist}. The
total fitting error in Fig.~\ref{fig:init-avg-dist}(a)
shows that \nrd~achieves good fitting results in the low error range
of 0~--~10~mm, as it can produce good template fits for the areas
where $\TemplateMesh$ is close to \S. However, as \nrd~is a model-free
method, the smooth topology of $\TemplateMesh$ may not be preserved
during the deformation, e.g. convex surfaces of $\TemplateMesh$
may be deformed into non-convex surfaces after \nrd. This leads
to large fitting errors for areas of $\TemplateMesh$ that are far from
\S. \scapeModelInit~uses a shape space fitting prior to NRD, 
which allows for a better initial alignment of
$\TemplateMesh$ to \S. Note that \scapeModelInit~results in
a better fitting accuracy in the high error range of
$10-20~mm$. 
%\todo{why $0-10~mm$ fitting errors are worse?}
The fitting result by \scapeModelInit~favorably compares against using
\jain~alone. Although \jain~results into deformations preserving the
human body shape topology, the shape space is learned from the
relatively specialized MPI Human Shape dataset containing mostly young
adults and thus cannot represent all shape variations.

We also analyze the differences in the mean fitting errors per vertex
in Fig.~\ref{fig:init-avg-dist}(b). \nrd~achieves good fitting results
for most of the vertices. However, the arms are not fitted well due to
differences in body posture of $\TemplateMesh$ and
\S. Furthermore, the average fitting error is not smooth, which shows
that despite using $E_s$, \nrd~may produce non-smooth
deformations. In contrast, the result of \scapeModelInit~is smoother
and has a lower fitting error for the arms. Clearly, the average
fitting error of \jain~is much higher, with notably worse fitting
results for arms, belly and chest.

\subsection{\nrd~parameters}
Second, we evaluate the influence of the weight relaxation during
\nrd~on the fitting accuracy. Specifically, we compare
the standard weighting scheme where weights are relaxed in
each iteration (\scapeModelInit) to the case where the weights stay
constant (\scapeModelInitcw). Fig.~\ref{fig:init-avg-dist}(a) shows
that the total fitting error of \scapeModelInit~is lower than
\scapeModelInitcw. This is because \scapeModelInitcw~enforces higher
localized rigidity by keeping weights constantly high,
while \scapeModelInit~relaxes the weights so that $\TemplateMesh$ can
fit more accurately to \S. This explanation is supported by
consistently higher per-vertex mean fitting errors in case of
\scapeModelInitcw~compared to \scapeModelInit, as shown in
Fig.~\ref{fig:init-avg-dist}(b). The highest differences are
in the areas of high body shape variability, such as belly and
chest. Different weight reduction schemes such as $\beta:=0.5\beta$,
$\gamma:=0.5\gamma$ and $\beta:=0.25\beta$, $\gamma:=0.25\gamma$ lead
to better fitting accuracy compared to constant weights, with the
latter scheme achieving slightly better results and faster convergence
rates. We thus use the proposed weight reduction scheme in the
following.

\begin{figure*}[tbp]
\centering
\bgroup
\tabcolsep 1.5pt
\renewcommand{\arraystretch}{0.2}
\subfigure[Total fitting error]{
\begin{tabular}{c}
\includegraphics[height=0.24\linewidth]{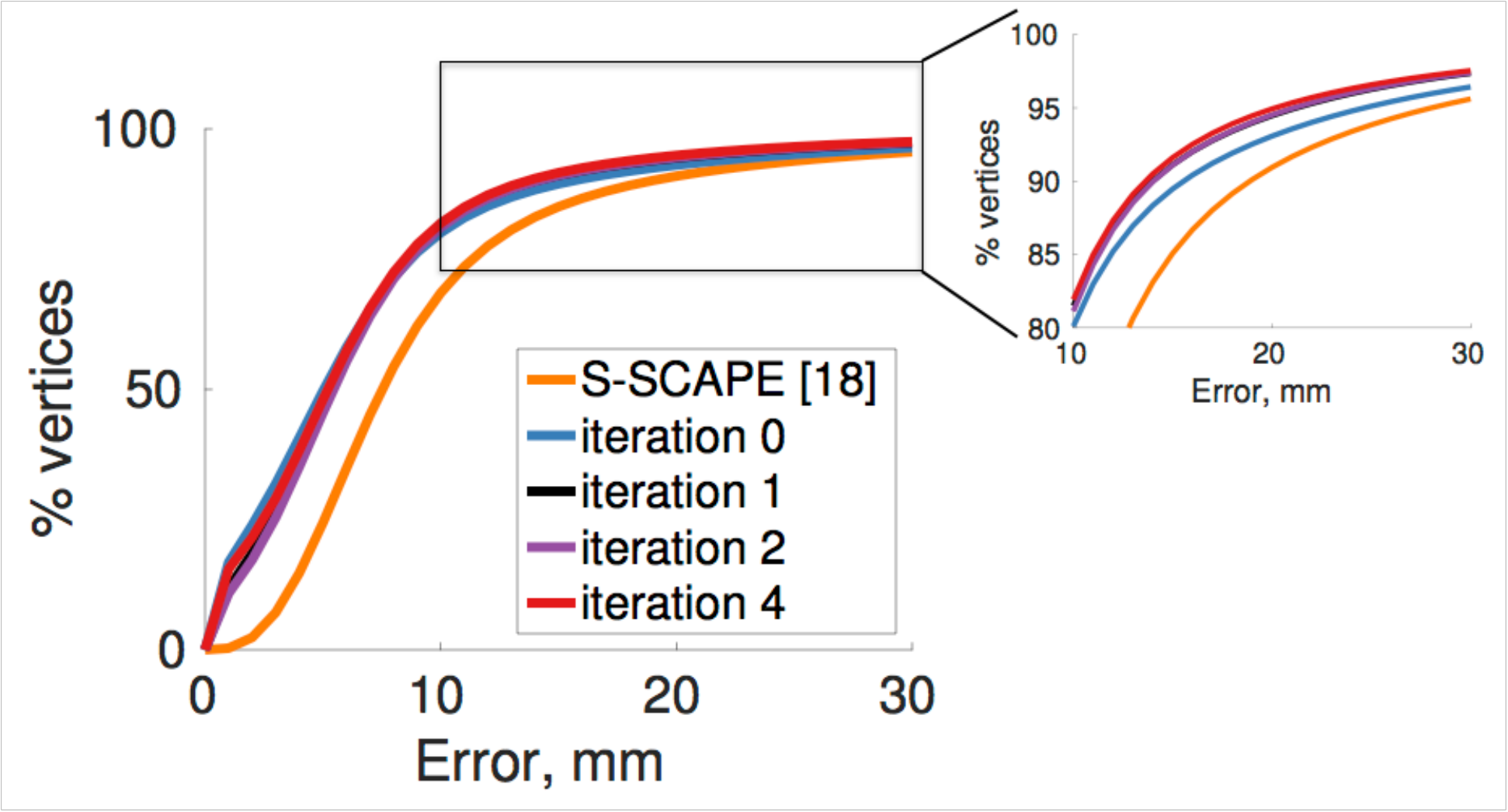}\\
\\
%\vspace{0.8em}
\end{tabular}
}
\subfigure[Mean fitting error on a sample scan]{
\begin{tabular}{cccc}
%\includegraphics[trim=6cm 1cm 6cm 1cm,clip=true,height=0.2\linewidth]{model-mpii-all-samples-warm-sched-3-scanidx-1-thresh1-0-thresh2-10-scan-45.png}
%}
\includegraphics[trim=10cm 0cm 0cm 0cm, clip=true,
height=0.16\linewidth]{mpii-model-validnn-thresh1-0-thresh2-10-meanDist-45.pdf}
&
\includegraphics[trim=6cm 2cm 6cm
1cm,clip=true,height=0.16\linewidth]{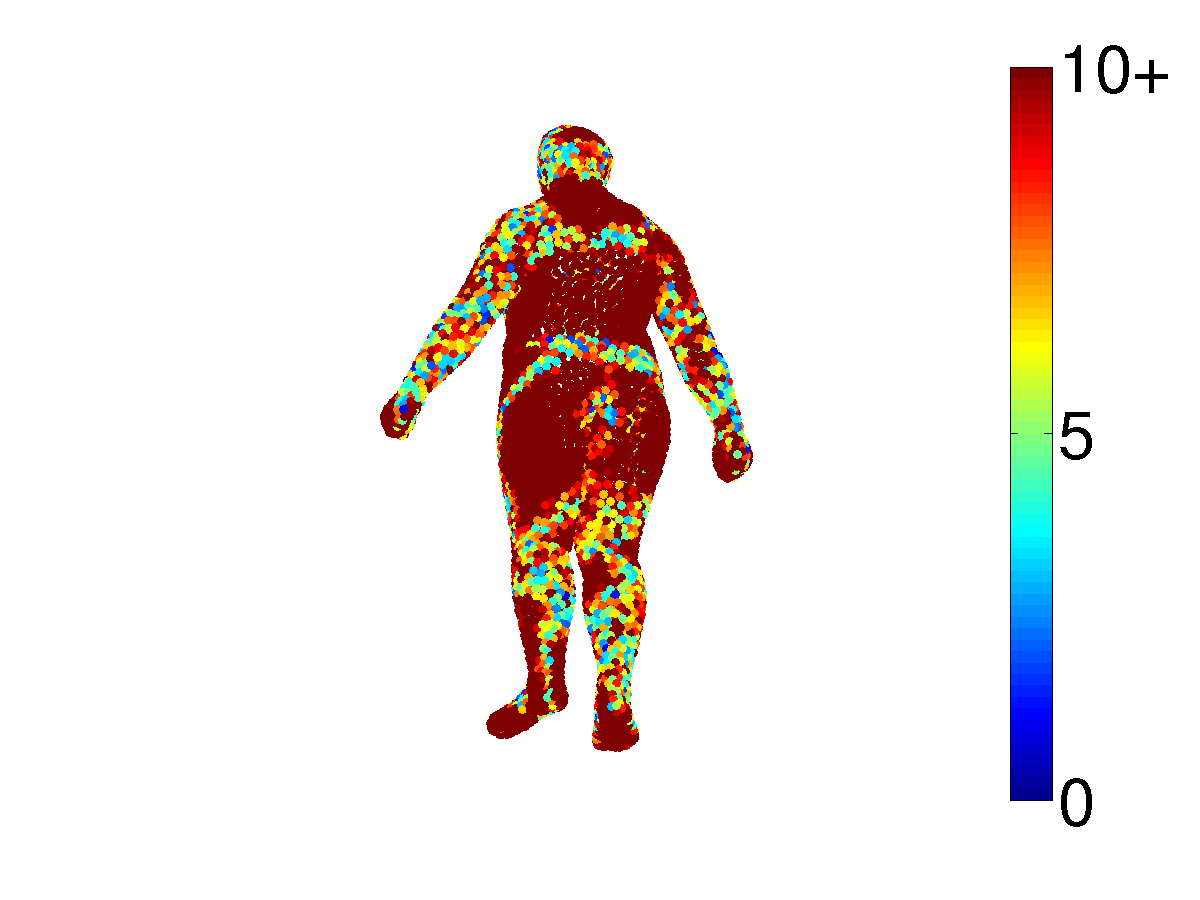}
&
\includegraphics[trim=6cm 2cm 6cm
1cm,clip=true,height=0.16\linewidth]{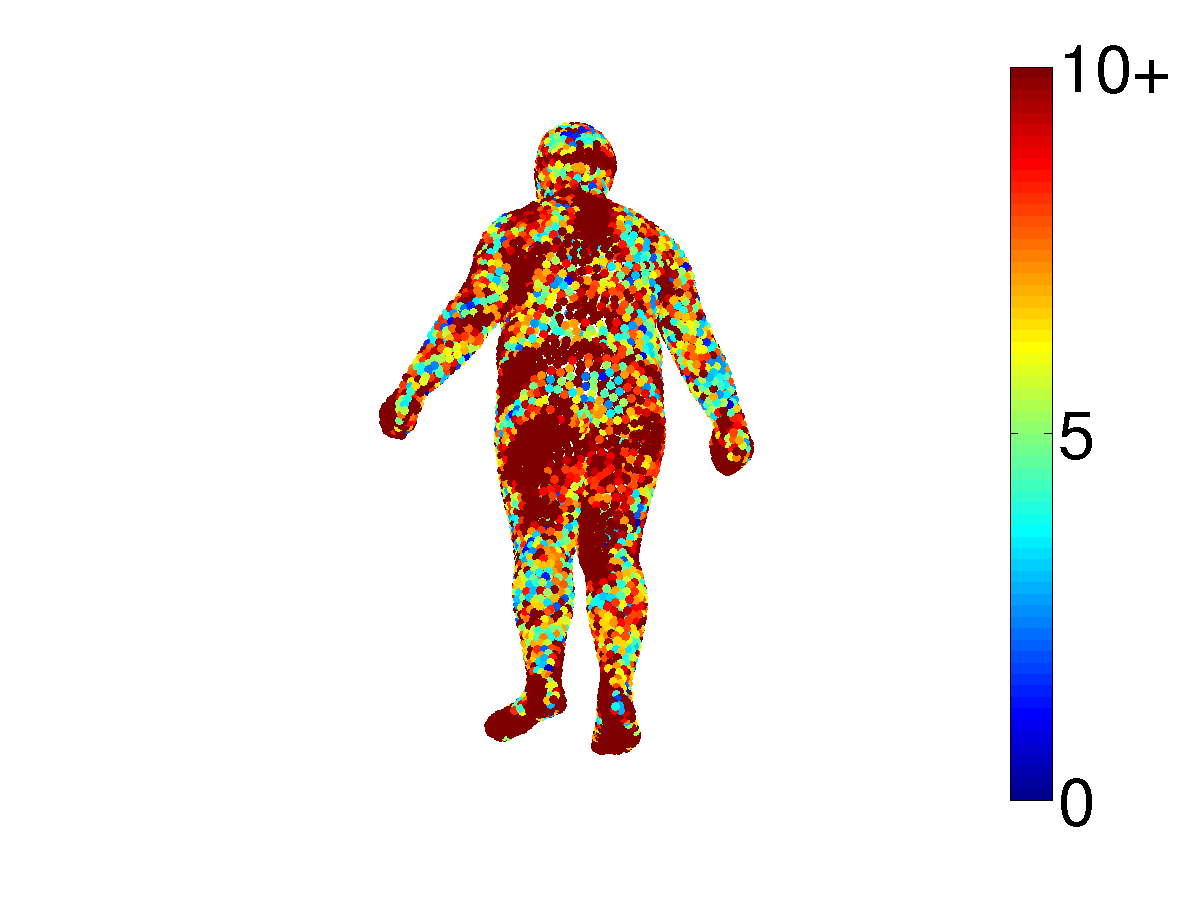}
&
\includegraphics[trim=6cm 2cm 6cm 1cm,clip=true,height=0.16\linewidth]{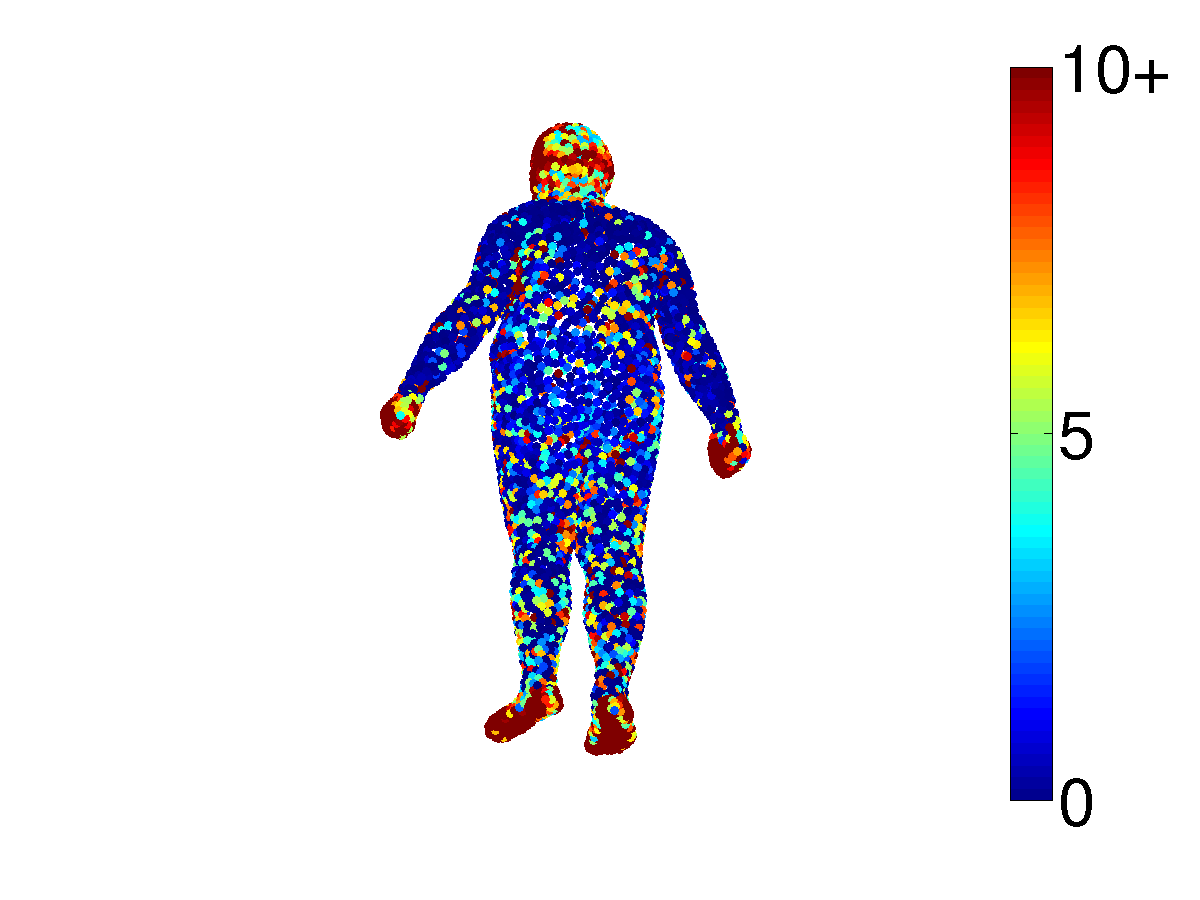}
\\
&iteration 0 & iteration 2& iteration 4\\[15pt]
\vspace{0.5em}
\end{tabular}
}

% \caption{Fitting error after a single (init-NRD 3), three (bootsrap 2)
%   and five (bootstrap 4) passes through ``CAESAR'' database. Shown
%   are (a) proportion of vertices [\%] which fitting error falls below
%   a certain threshold and (b) average fitting error per
%   vertex.}
%   \label{fig:init-mean-dist}
% \egroup
% \end{figure*}

% \begin{figure*}[tbp]
% \centering
% \bgroup
% \tabcolsep 1.5pt
% \includegraphics[trim=0cm 0cm 0cm 0cm,clip=true,width=0.50\linewidth]{bootstrap-all-dist-validNN-nrd-0-30-combined.pdf}\\

\caption{Fitting error after up to four bootstrapping rounds over the
  CEASAR database when using the \jain~\cite{Jain:2010:MovieReshape}
  as initialization for iteration 0. Shown are (a) the proportion of
  vertices [\%] with fitting error below a threshold and (b) the
  average fitting error per vertex.}
  \label{fig:boots-all-dist}
\egroup
\vspace{-0.5em}
\end{figure*}

\subsection{Bootstrapping} 

Third, we evaluate the fitting accuracy before and after performing
multiple rounds of bootstrapping. To that end, we use the output of
\scapeModelInit~(iteration 0) to learn a new statistical shape space,
which is in turn used to initialize \nrd~during the second pass over
the data (iteration 1). This process is repeated for five passes. The
number of registered scans that passed the visual inspection after
each round is $1771$, $3253$, $3641$, $4237$ and $4301$,
respectively. This results show that bootstrapping allows to register
and thus to learn from an increasing number of scans. Fitting results
are shown in Fig.~\ref{fig:boots-all-dist}. The close-up shows that
although the overall fitting accuracy before and after bootstrapping
is similar, bootstrapping allows to slightly improve the fitting
accuracy in the range of $10-30~mm$. Fitting results after three
passes over the dataset (iteration 2) are slightly better compared to
the initial fitting (iteration 0), and the accuracy is further
increased after five passes (iteration 4).
Fig.~\ref{fig:boots-all-dist}~(b) shows sample fitting results before
and after several bootstrapping rounds. Largest improvements are
achieved for the belly and chest - areas with large shape
variability. The fitting improves with an increasing number of
bootstrapping rounds. We use the fitting results after five passes
(iteration 4) to learn the \jain~space used in the following.

\section{Evaluation of statistical shape space}
\label{sec:eval_shape_space}

In this section, we evaluate the \jain~space using the statistical
quality measures of generalization and specificity~\cite{Styner2003}.

\subsection{Quality measure} 

We use two complementary measures of shape statistics. Generalization
evaluates the ability of a shape space to represent unseen instances
of the object class. Good generalization means the shape space is
capable of learning the characteristics of an object class from a
limited number of training samples, poor generalization indicates
overfitting of the training set. Generalization is measured using
leave-one-out cross reconstruction of training samples, i.e. the shape
space is learned using all but one training sample and the resulting
shape space is fitted to the excluded sample.  The fitting error is
measured using the mean vertex-to-vertex Euclidean
distance. Generalization is reported as mean fitting error averaged
over the complete set of trials, and plotted as a function of the
number of shape space parameters. It is expected that the mean error
decreases until convergence as the number of shape space parameters
increases.

Specificity measures the ability of a shape space to generate
instances of the object class that are similar to the training
samples. The specificity test is performed by generating a set of
instances randomly drawn from the learned shape space and by comparing
them to the training samples. The error is measured as average
distance of the generated instances to their nearest neighbors in the
training set. It is expected that the mean distance increases until
convergence with increasing number of shape space parameters.  We
follow Styner et al.~\cite{Styner2003} and generate 10,000 random
samples.

\begin{figure*}[tbp]
\centering
\bgroup
\tabcolsep 1.5pt
\renewcommand{\arraystretch}{0.2}
%\subfigure[Bootstrapping]{
\begin{tabular}{cccc}
\begin{sideways} \quad
 \bf Generalization \end{sideways}
&
\includegraphics[width=0.29\linewidth]{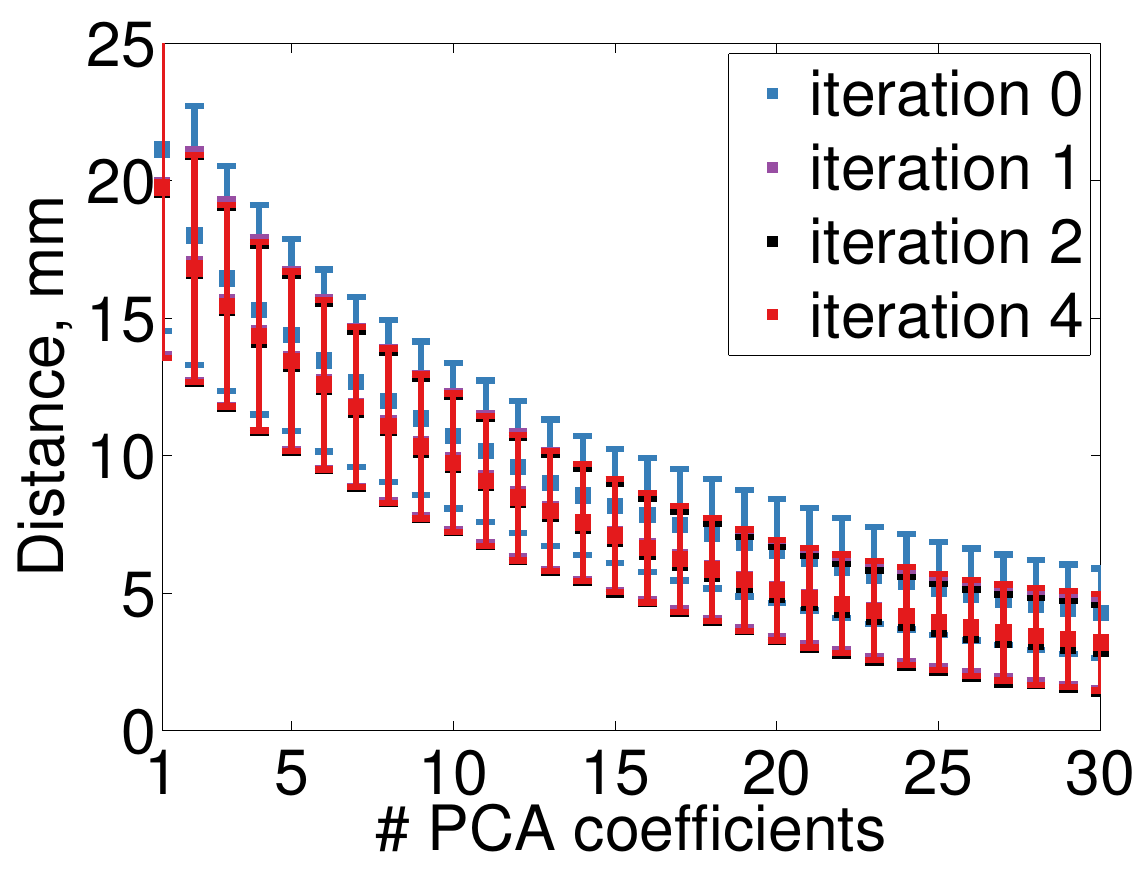}
&
\includegraphics[width=0.29\linewidth]{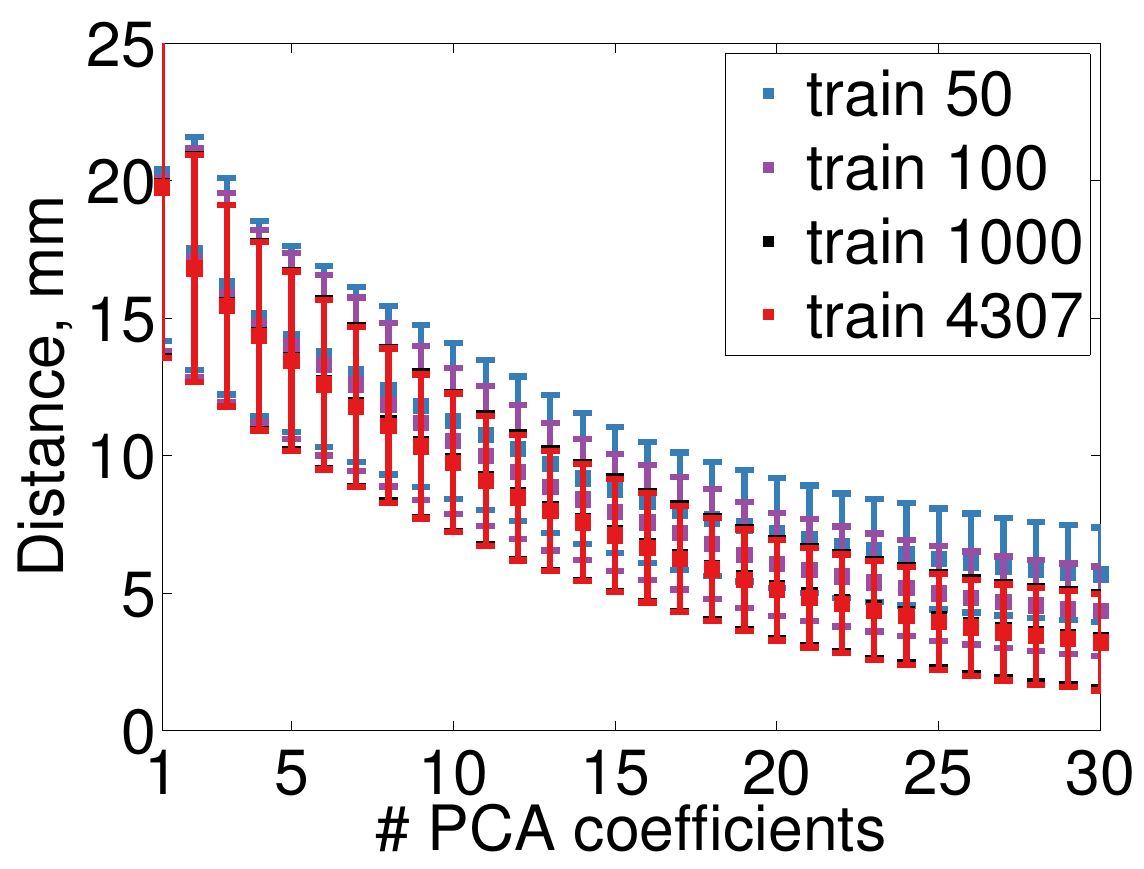}
&
\includegraphics[width=0.29\linewidth]{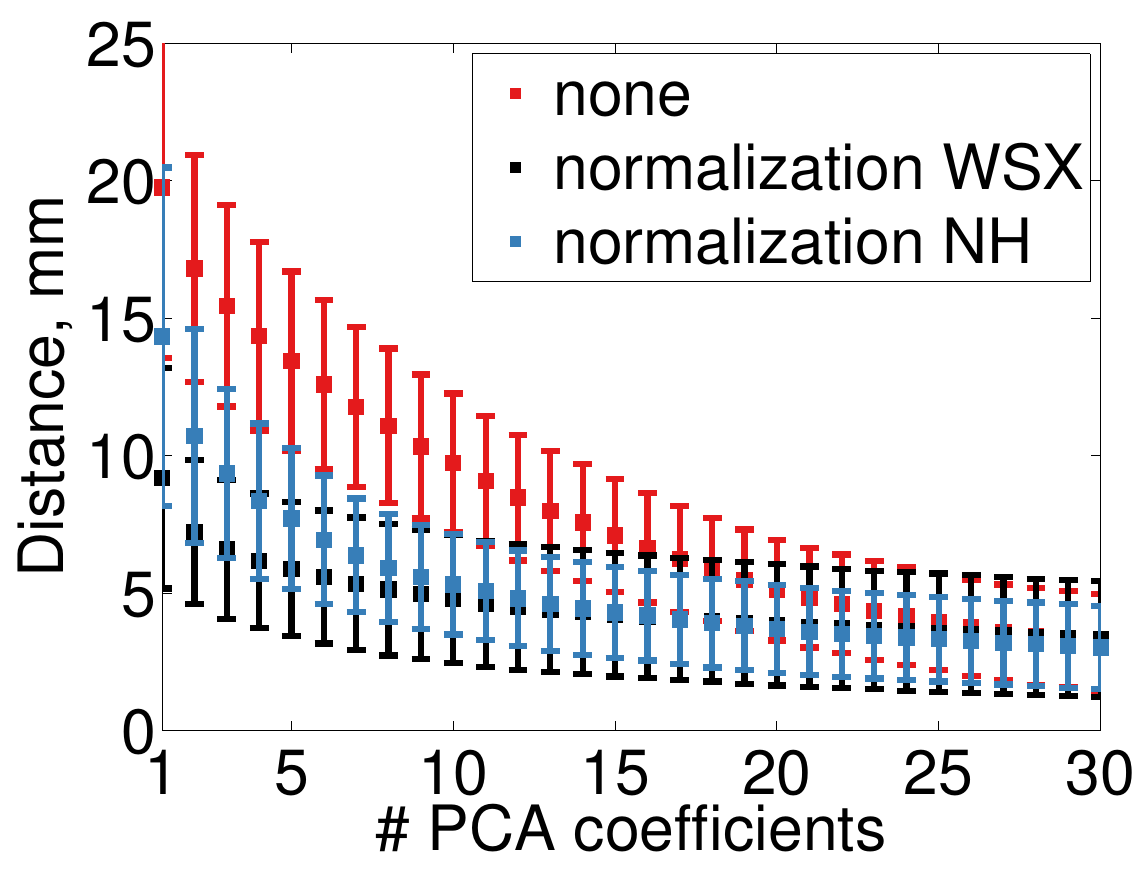}
\\[2pt]
\begin{sideways} \quad\quad
 \bf Specificity \end{sideways}
&
\includegraphics[width=0.29\linewidth]{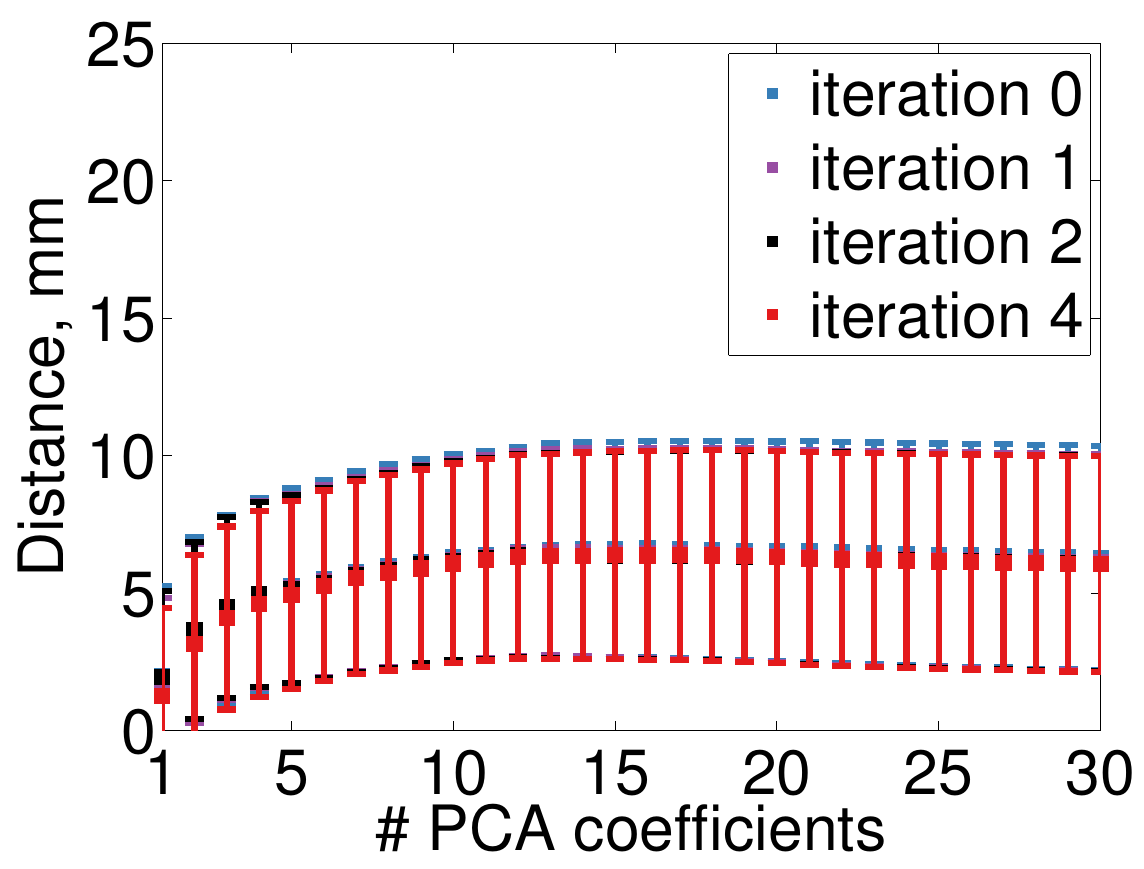}
&
\includegraphics[width=0.29\linewidth]{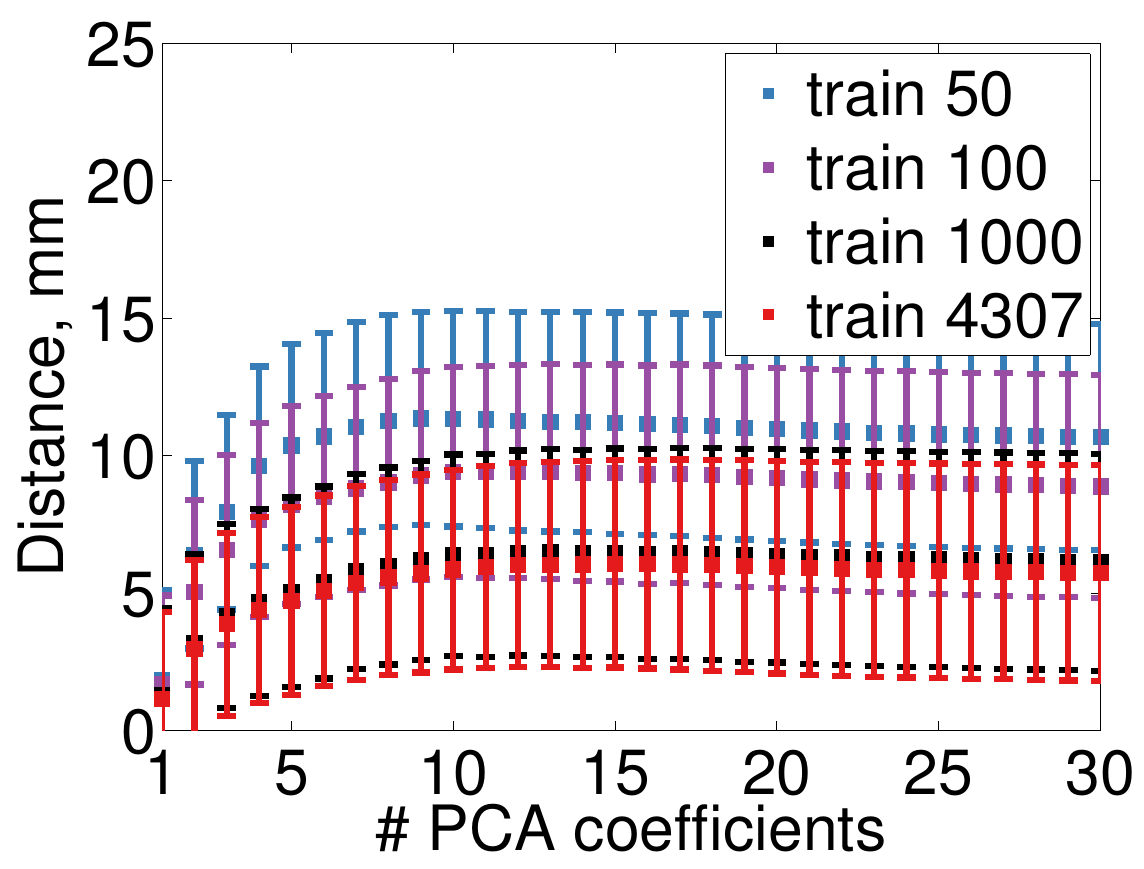}
&
\includegraphics[width=0.29\linewidth]{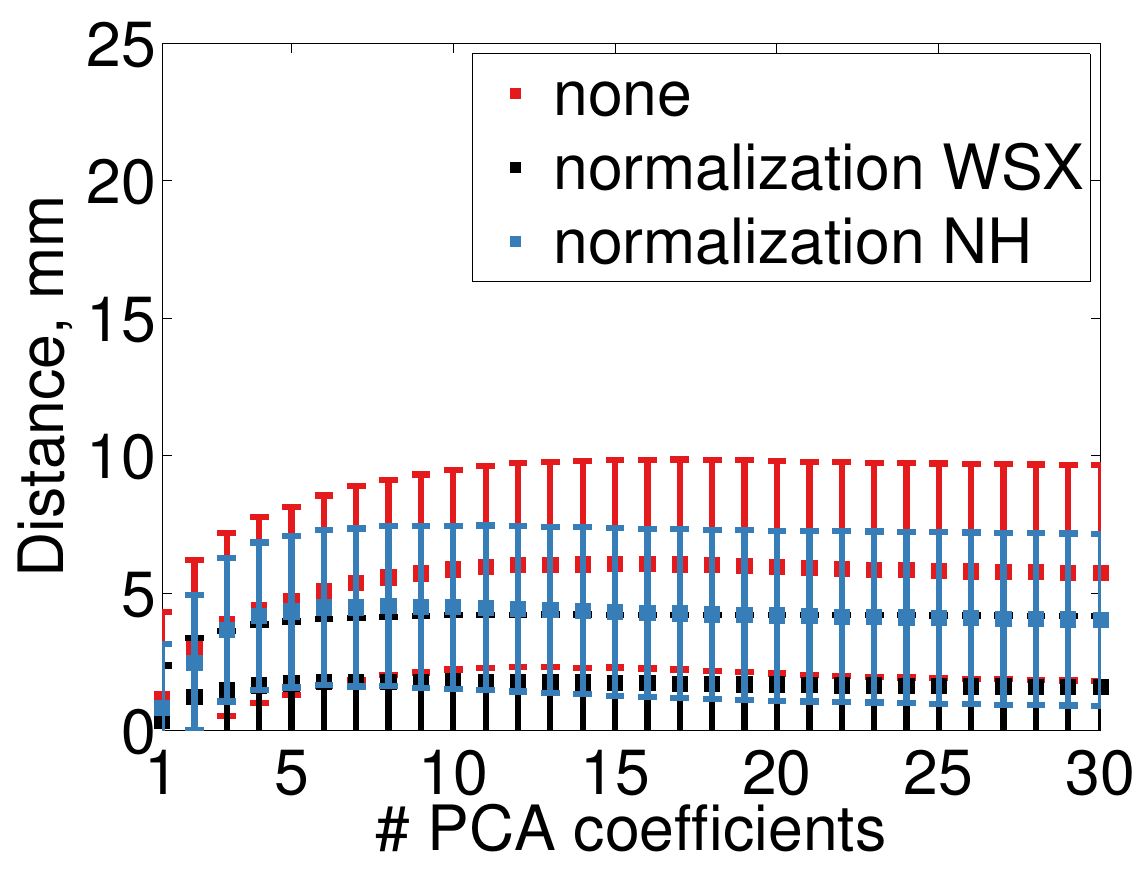}
\\
& (a) bootstrapping & (b) \# training samples & (c) posture normalization\\
\end{tabular}

\caption{Influence of different design choices on statistical quality
  measures. Shown are influence of (a) bootstrapping, (b) number of
  training samples and (c) posture normalization on generalization
  (top row) and specificity (bottom row). Best viewed with zoom on the
  screen.}

\label{fig:model-quality}
\egroup
\vspace{-0.5em}
\end{figure*}

\subsection{Bootstrapping}
We evaluate the influence of bootstrapping on the quality of the
statistical shape space by comparing models obtained after zero, one,
two and four iterations of bootstrapping. The geometry of
the training samples changes in each bootstrapping round, which makes
the generalization and specificity results incomparable across
different shape spaces. We thus use the training samples obtained
after four iterations of bootstrapping as ``ground truth'', i.e.,
the reconstruction error of generalization and the nearest
neighbor distance of specificity for each shape
space is computed w.r.t. fitting results after four bootstrapping
rounds. This allows for a fair comparison across
different statistical shape spaces.

The results are shown in
Fig.~\ref{fig:model-quality}(a). Generalization error is already low
after a single iteration of bootstrapping because after one iteration,
the shape space is learned from a significantly larger number of
training samples, thereby using samples with higher shape variation
that were discarded in the $0^{th}$ iteration. The following rounds of
bootstrapping have little influence on generalization and specificity,
with the shape space after four iterations resulting in a slightly
lower specificity error than for previous iterations for a small
number of shape parameters.

\subsection{Number of training samples} 
To evaluate the influence of the number of training samples, we vary
the number of samples obtained after four bootstrapping
iterations. Specifically, we consider subsets of $50$, $100$, $1,000$
and $4,307$ ($all-1$) training samples. To compute a shape space, the
desired number of training shapes are sampled from the entire set of
training samples. For generalization, we cross-evaluate on all $4,308$
training samples by leaving one sample out and by sampling the desired
number of training shapes from the remaining samples. For specificity,
we compute the nearest-neighbor distances to all $4,308$ training
samples to find the closest sample.

The results are shown in Fig.~\ref{fig:model-quality}(b). The shape
space learned from the smallest number of samples performs
worst. Increasing the number of samples consistently improves the
performance with the best results achieved when using the maximum
number. Both generalization and specificity error reduction is most
pronounced when increasing the number of samples from $50$ to
$100$. Further increasing the number of samples to $1,000$ affects
specificity much stronger than generalization. This shows that the
shape space learned from only $100$ samples generalizes well, while
its generative qualities are poor. Increasing the number of samples
from $1,000$ to $4,307$ only slightly reduces both generalization and
specificity errors, which shows that a high-quality statistical shape
space can be learned from $1,000$ samples.

\subsection{Posture normalization}
Finally, we evaluate the generalization and specificity of the shape
space obtained when performing posture normalization using the methods
of Wuhrer~et~al. \cite{WuhrerPIS12} (\WSX) and Neophytou and
Hilton~\cite{Neophytou2013} (\NH). The results are shown in
Fig.~\ref{fig:model-quality} (c). Posture normalization significantly
improves generalization and specificity, with \WSX~achieving the best
result. The reduction of the average fitting error in case of
generalization is highest for a low number of shape parameters. This
is because both \WSX~and \NH~lead to shape spaces that are more
compact compared to the shape space obtained with unnormalized
training shapes. Additionally, both posture-normalized shape spaces
exhibit much better specificity. Compared to the shape space trained
before posture normalization, randomly generated samples from both
shape spaces trained after \WSX~and \NH~exhibit less variation in
posture and are thus more similar to their corresponding
posture-normalized training samples.

Finally, we qualitatively examine the first five PCA components
learned by the following \jain~spaces: the current state-of-the-art
shape space \jain~\cite{Jain:2010:MovieReshape}, our shape space
without posture normalization and with posture normalization using
\WSX~and \NH. The results are shown in
Fig.~\ref{fig:model_vis_pca}. Major modes of shape variation by
\jain~(row 1) are affected by global and local posture-related
deformations, such as moving of arms or tilting the body. In contrast,
the principal modes of variation by our shape space (row 2) are mostly
due to shape changes, which is achieved due to better template fitting
procedure and a more representative training set. However, small
posture variations are still part of the learned shape
space. Performing posture normalization of the training samples prior
to learning the shape space completely factors out changes due to
posture, as can be seen in the major principal components of $\ours+$\WSX~(row 3)
and $\ours+$\NH~(row 4).
\newcommand\h{0.36}

\begin{figure}[tbp]
\centering
\bgroup
\tabcolsep 0.0pt
\renewcommand{\arraystretch}{0.0}
\begin{tabular}{c cc c cc c cc c cc c cc}
\begin{sideways} \bf \textit \quad \quad \jain~\cite{Jain:2010:MovieReshape} \end{sideways}&
\includegraphics[trim=9.0cm 1cm 8.7cm 1cm, clip=true, height=\h\linewidth]{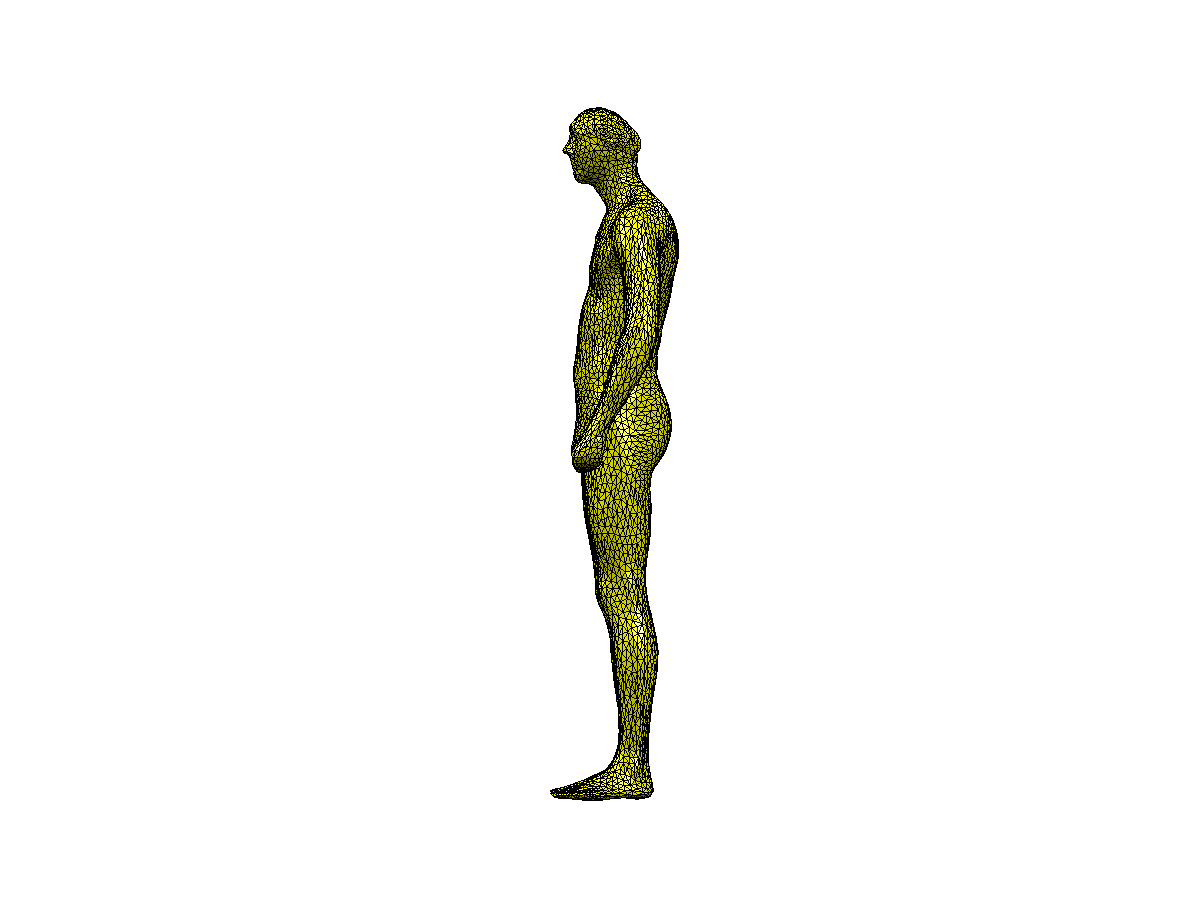}&
\includegraphics[trim=9.0cm 1cm 8.7cm 1cm, clip=true, height=\h\linewidth]{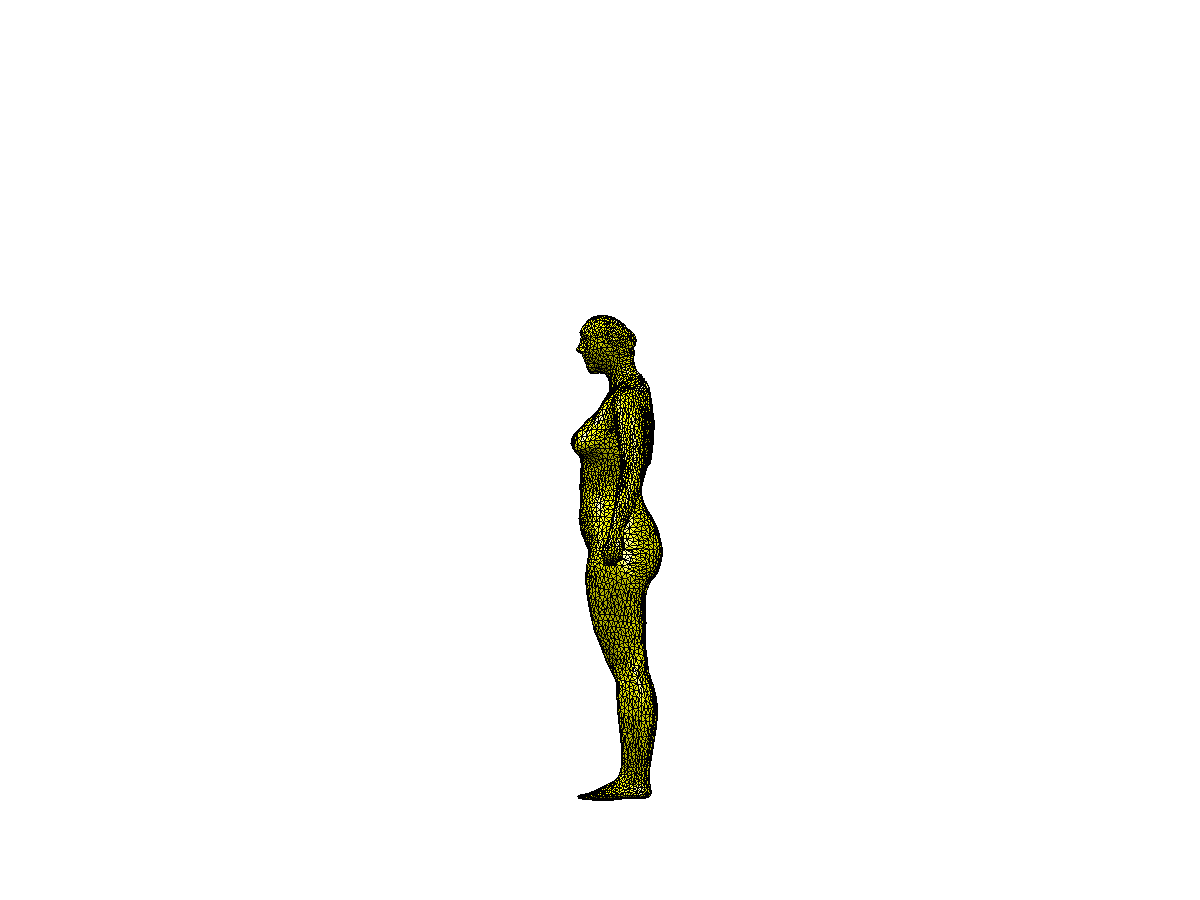}&
\quad\quad&
\includegraphics[trim=9.0cm 1cm 8.7cm 1cm, clip=true, height=\h\linewidth]{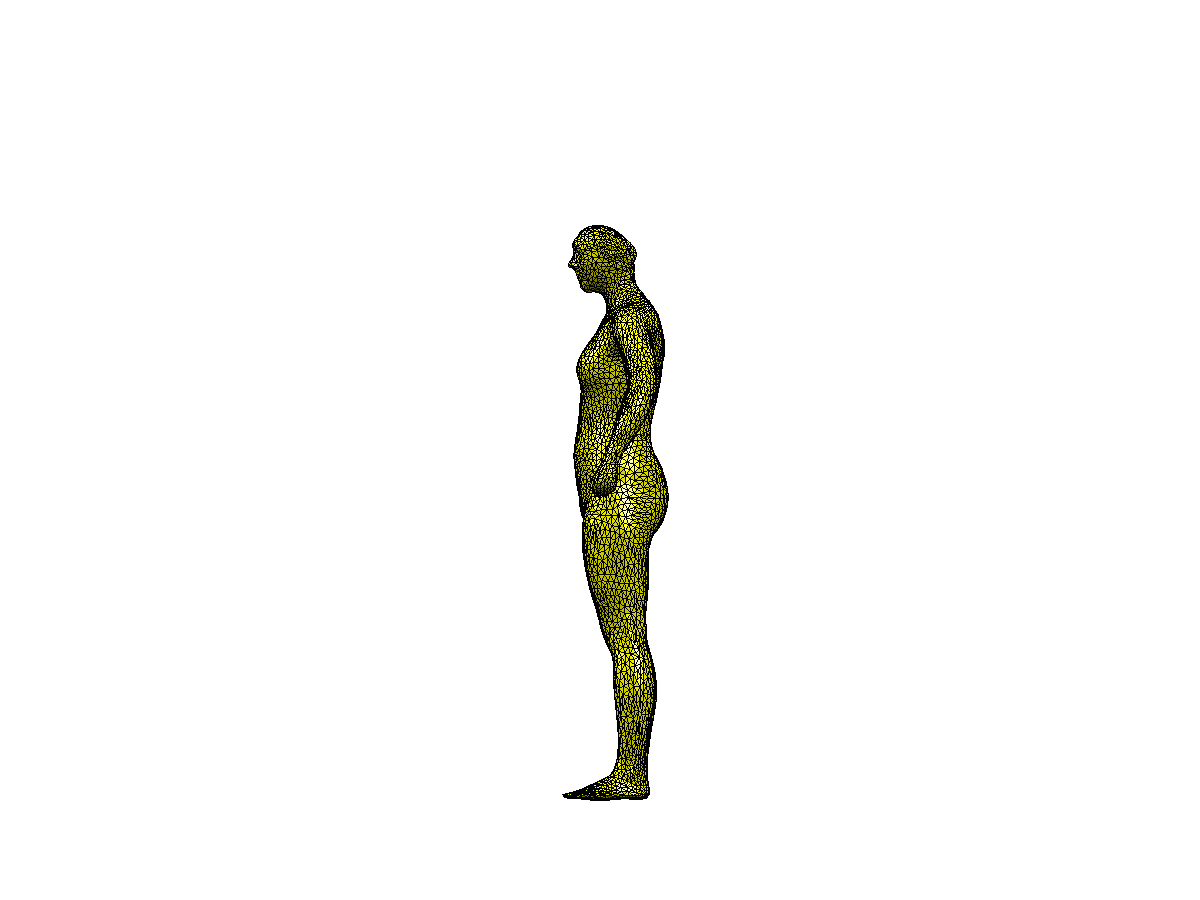}&
\includegraphics[trim=9.0cm 1cm 8.7cm 1cm, clip=true, height=\h\linewidth]{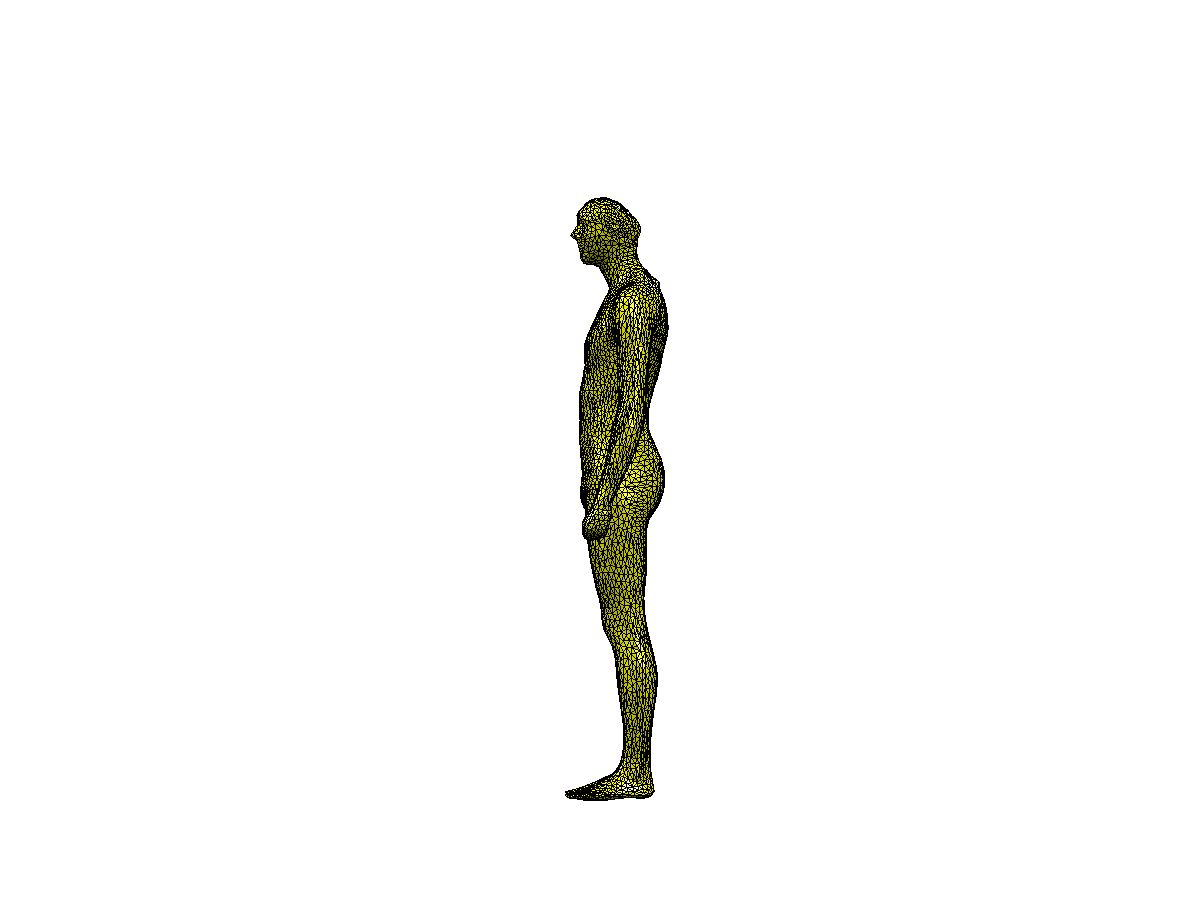}&
\quad\quad&
\includegraphics[trim=9.0cm 1cm 8.7cm 1cm, clip=true, height=\h\linewidth]{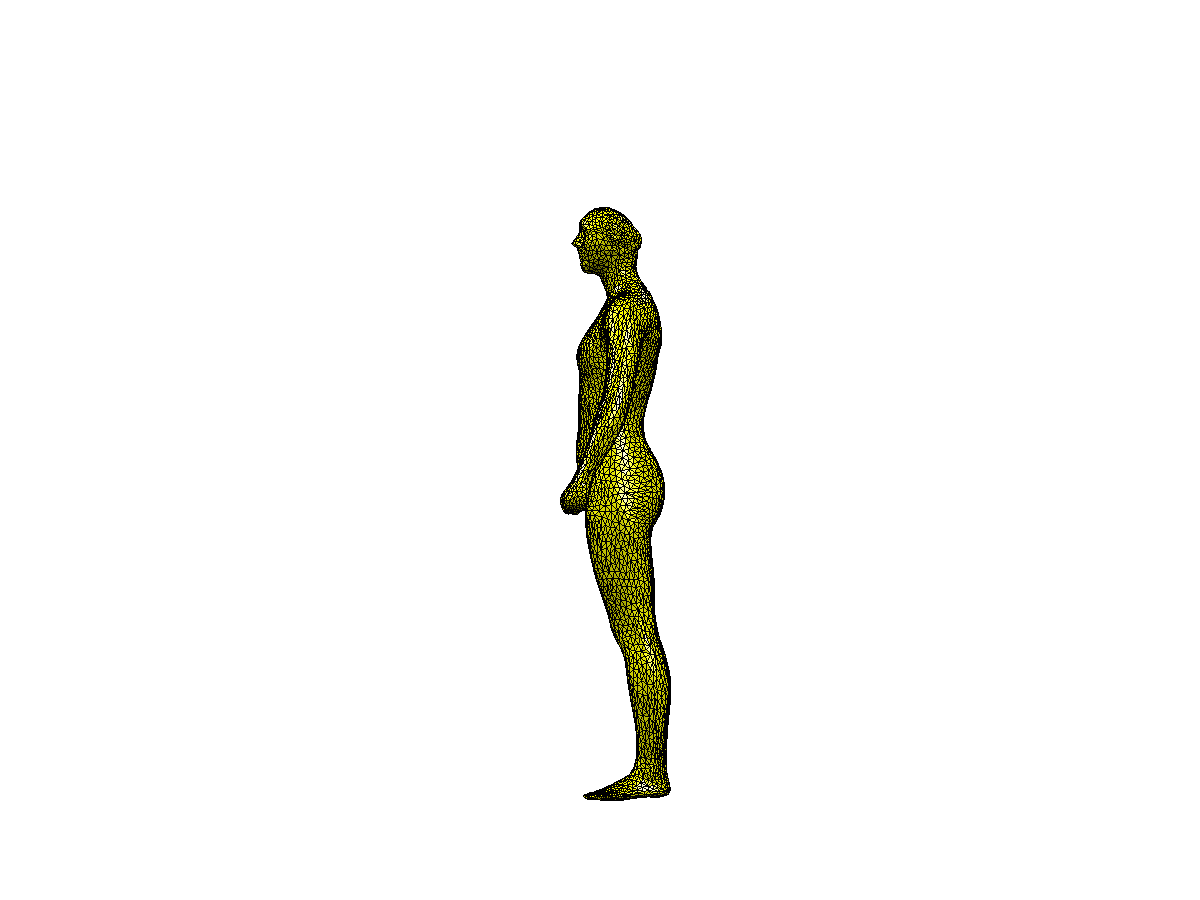}&
\includegraphics[trim=9.0cm 1cm 8.7cm 1cm, clip=true, height=\h\linewidth]{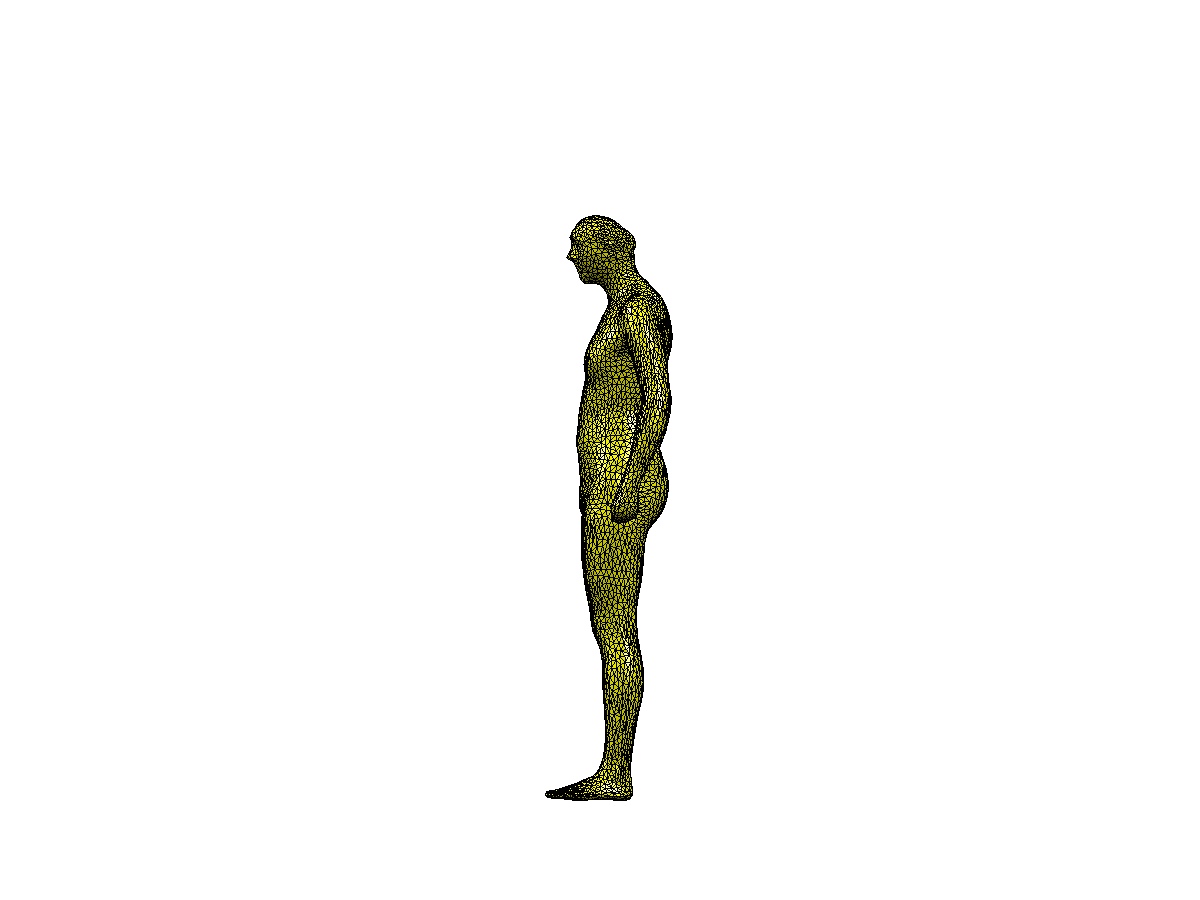}&
\quad\quad&
\includegraphics[trim=9.0cm 1cm 8.7cm 1cm, clip=true, height=\h\linewidth]{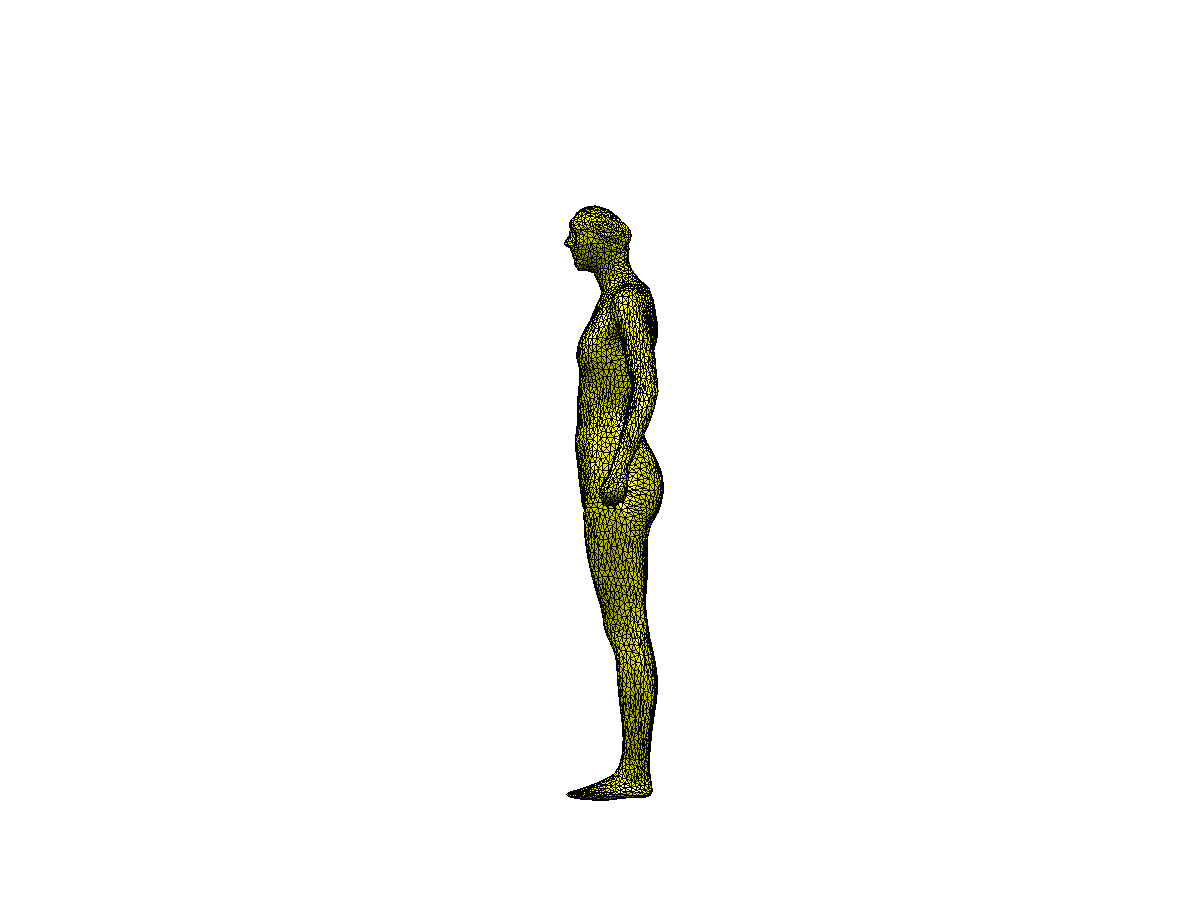}&
\includegraphics[trim=9.0cm 1cm 8.7cm 1cm, clip=true, height=\h\linewidth]{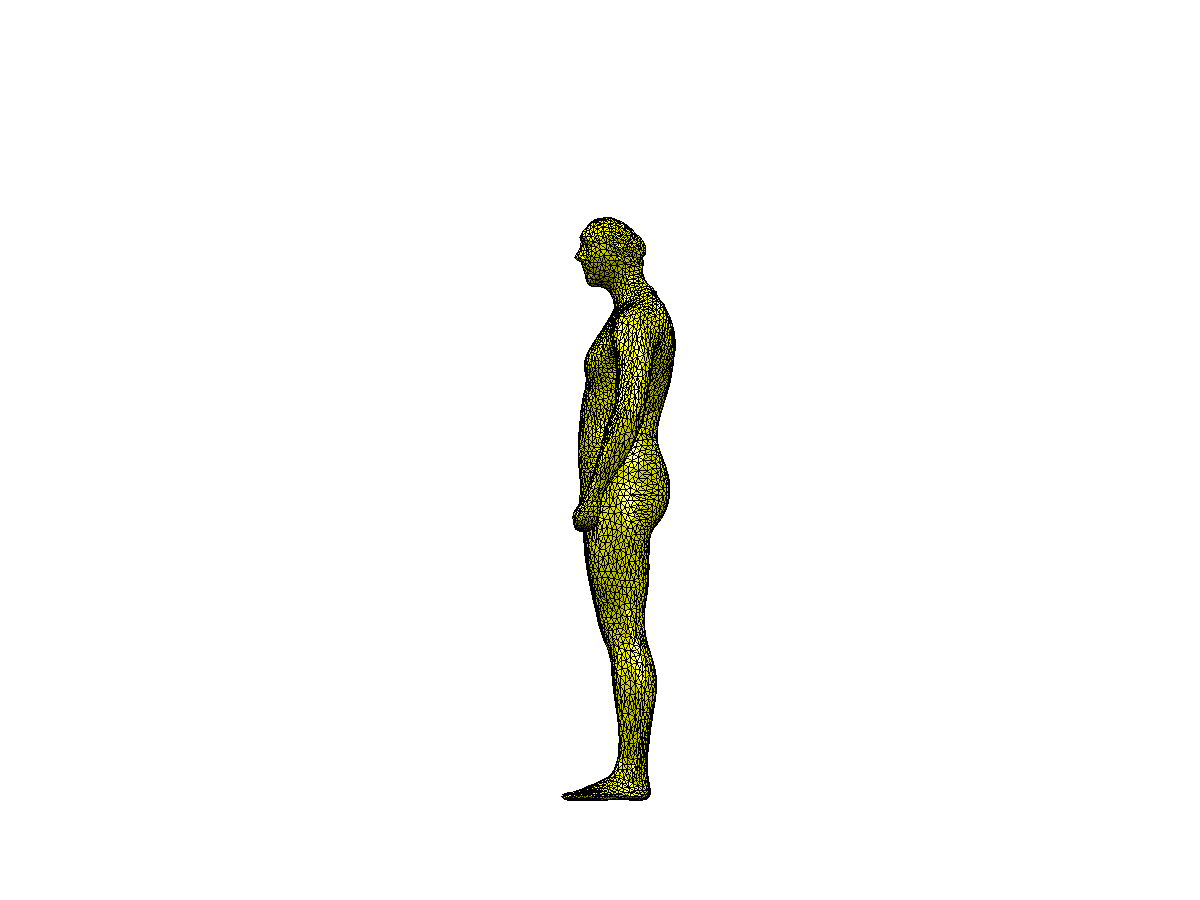}&
\quad\quad&
\includegraphics[trim=9.0cm 1cm 8.7cm 1cm, clip=true, height=\h\linewidth]{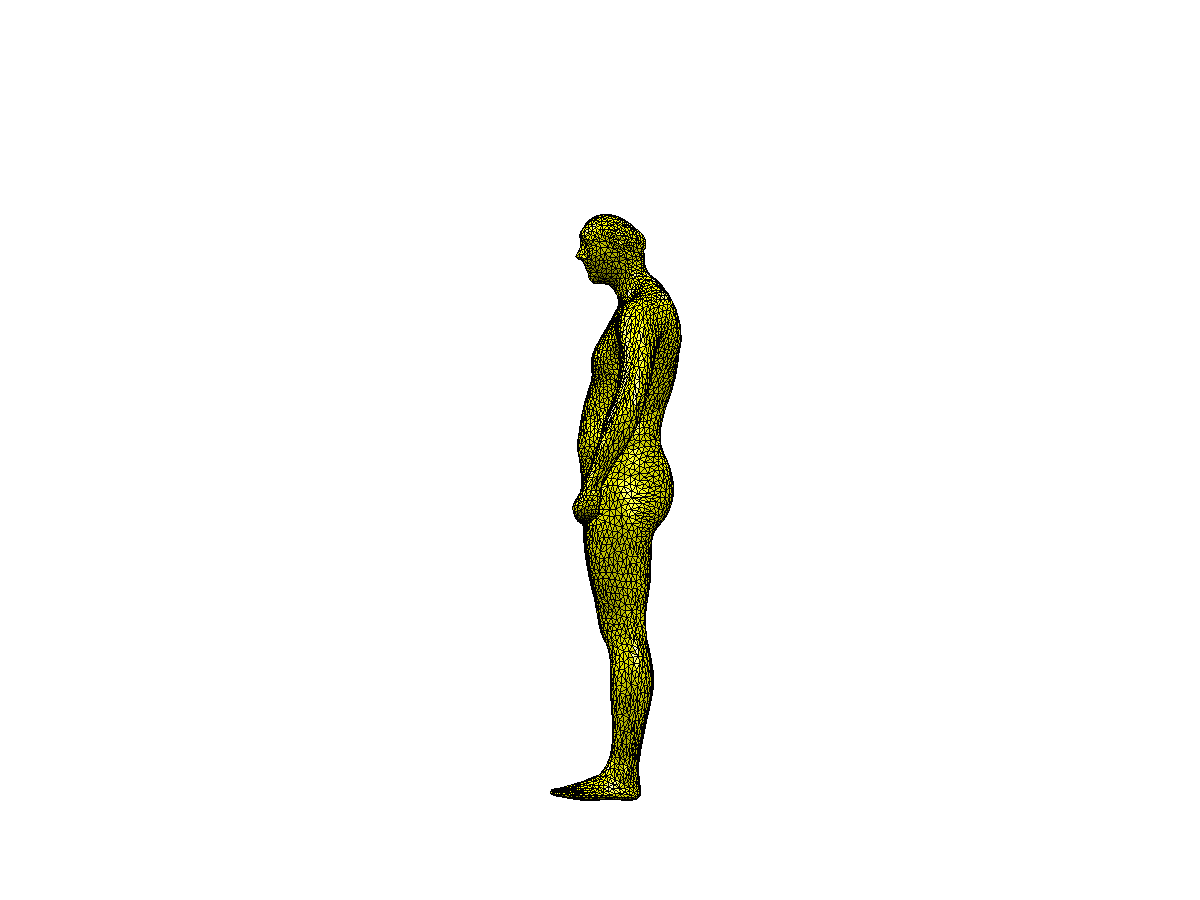}&
\includegraphics[trim=9.0cm 1cm 8.7cm 1cm, clip=true, height=\h\linewidth]{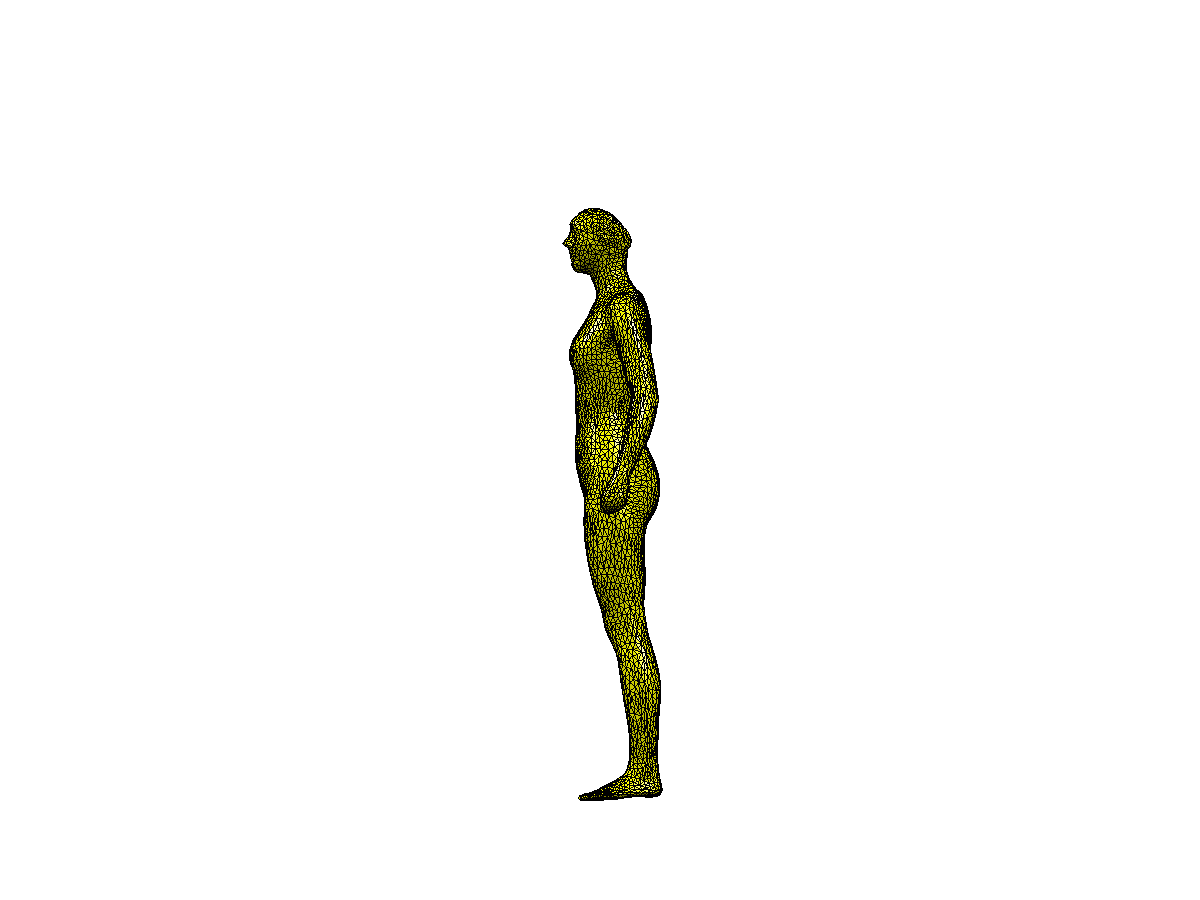}\\

\begin{sideways} \bf \textit \quad\quad\quad\quad \ours \end{sideways}&
\includegraphics[trim=9.0cm 1cm 8.7cm 1cm, clip=true, height=\h\linewidth]{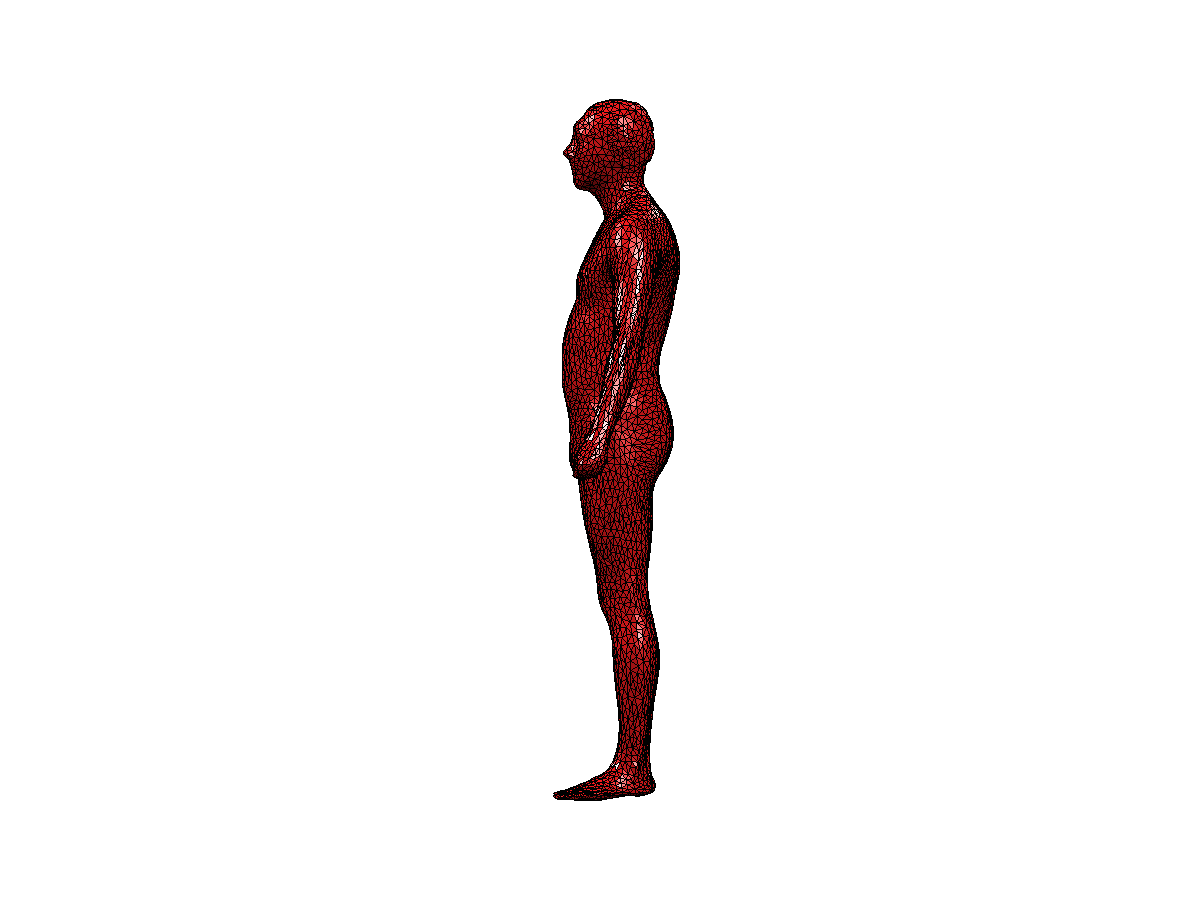}&
\includegraphics[trim=9.0cm 1cm 8.7cm 1cm, clip=true, height=\h\linewidth]{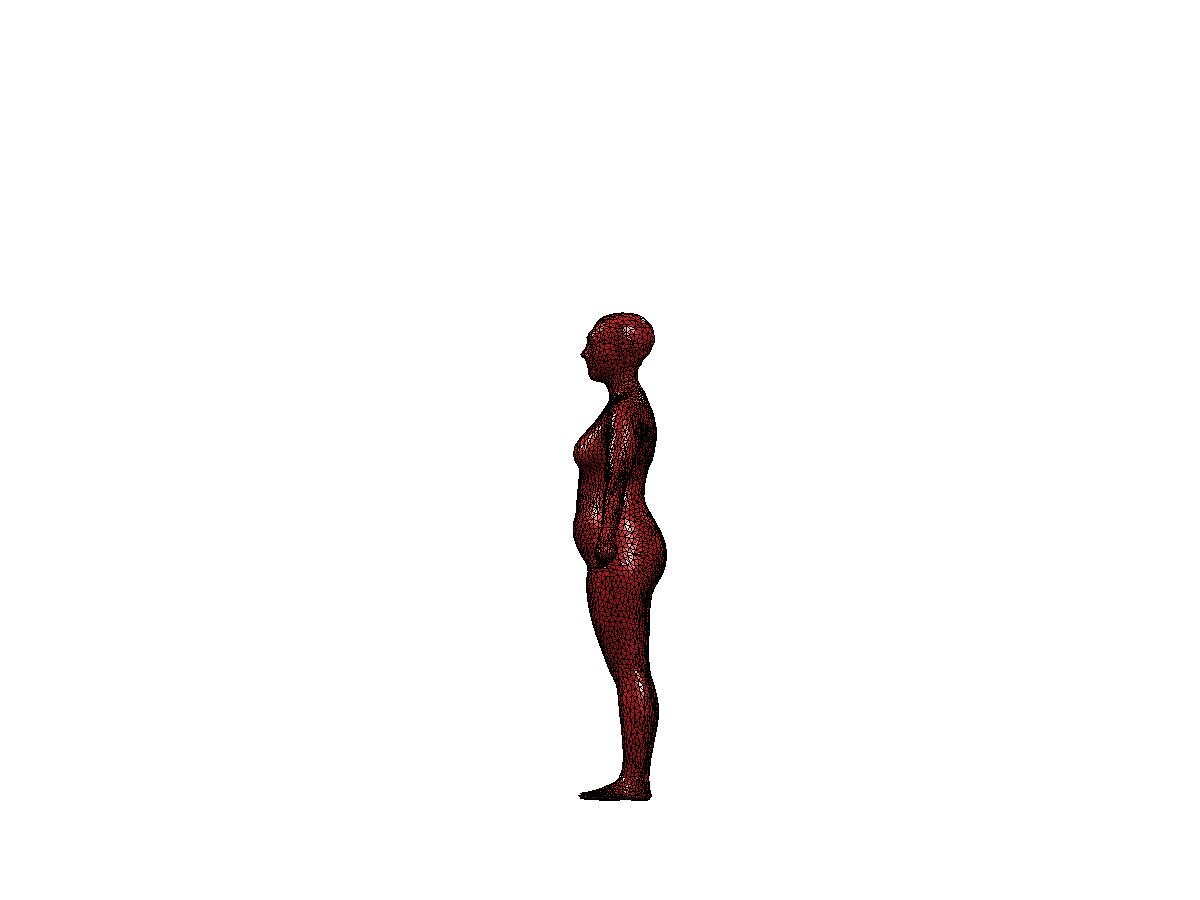}&
\quad\quad&
\includegraphics[trim=9.0cm 1cm 8.7cm 1cm, clip=true, height=\h\linewidth]{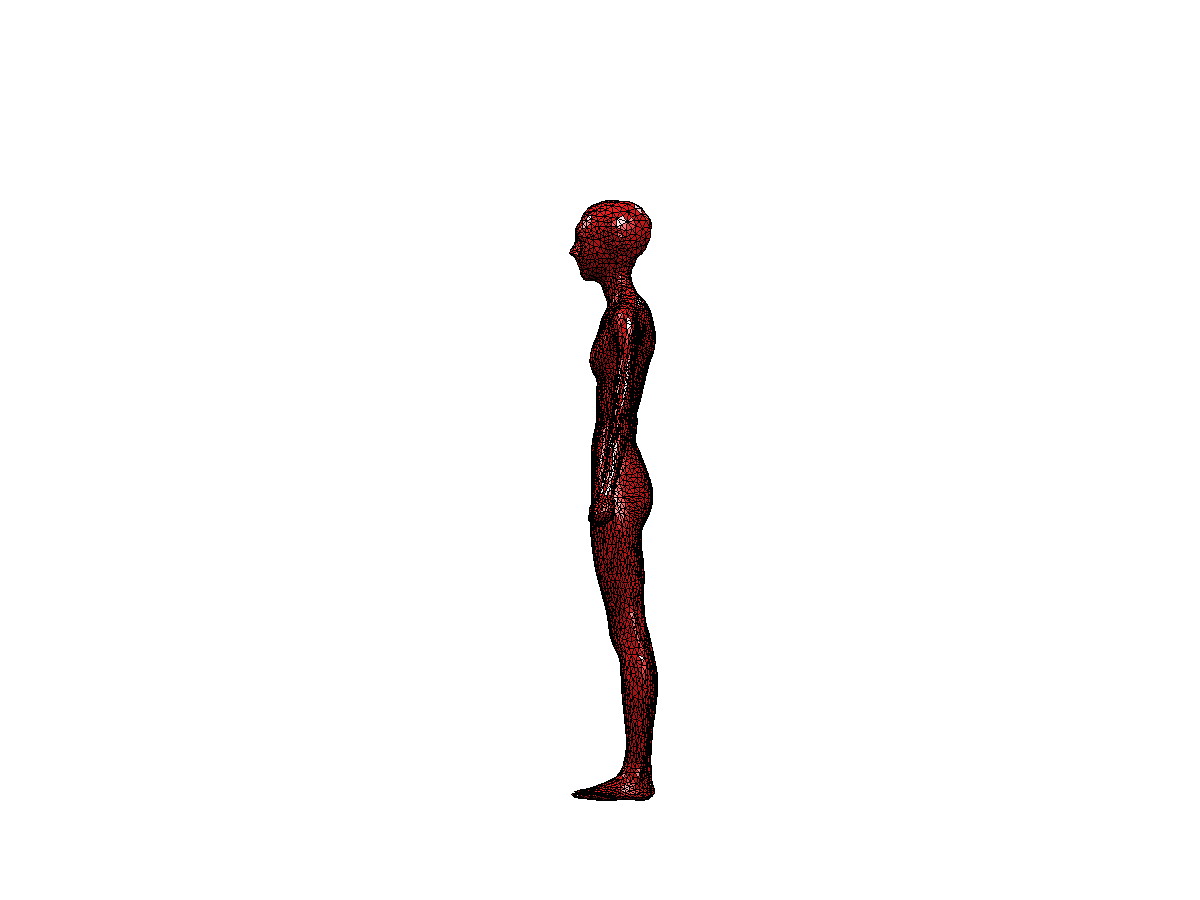}&
\includegraphics[trim=9.0cm 1cm 8.7cm 1cm, clip=true, height=\h\linewidth]{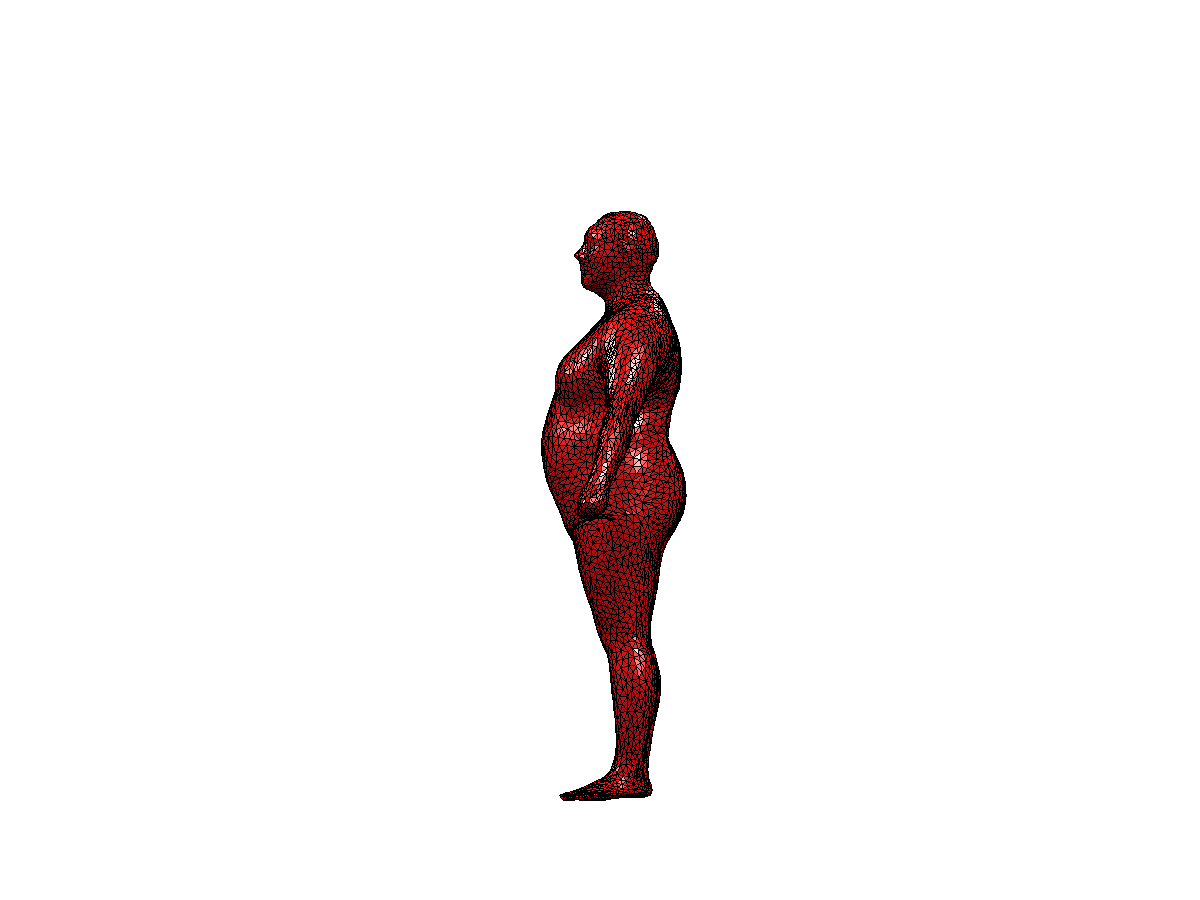}&
\quad\quad&
\includegraphics[trim=9.0cm 1cm 8.7cm 1cm, clip=true, height=\h\linewidth]{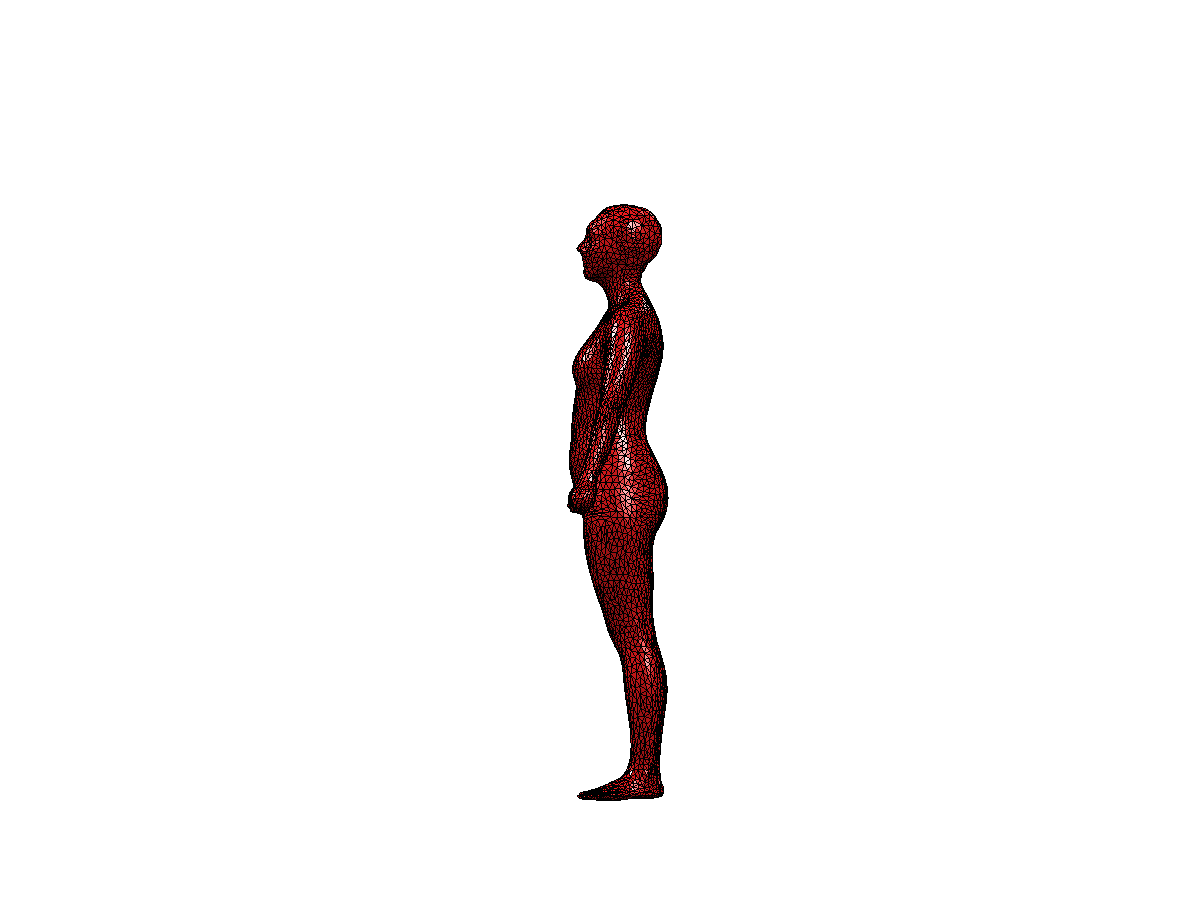}&
\includegraphics[trim=9.0cm 1cm 8.7cm 1cm, clip=true, height=\h\linewidth]{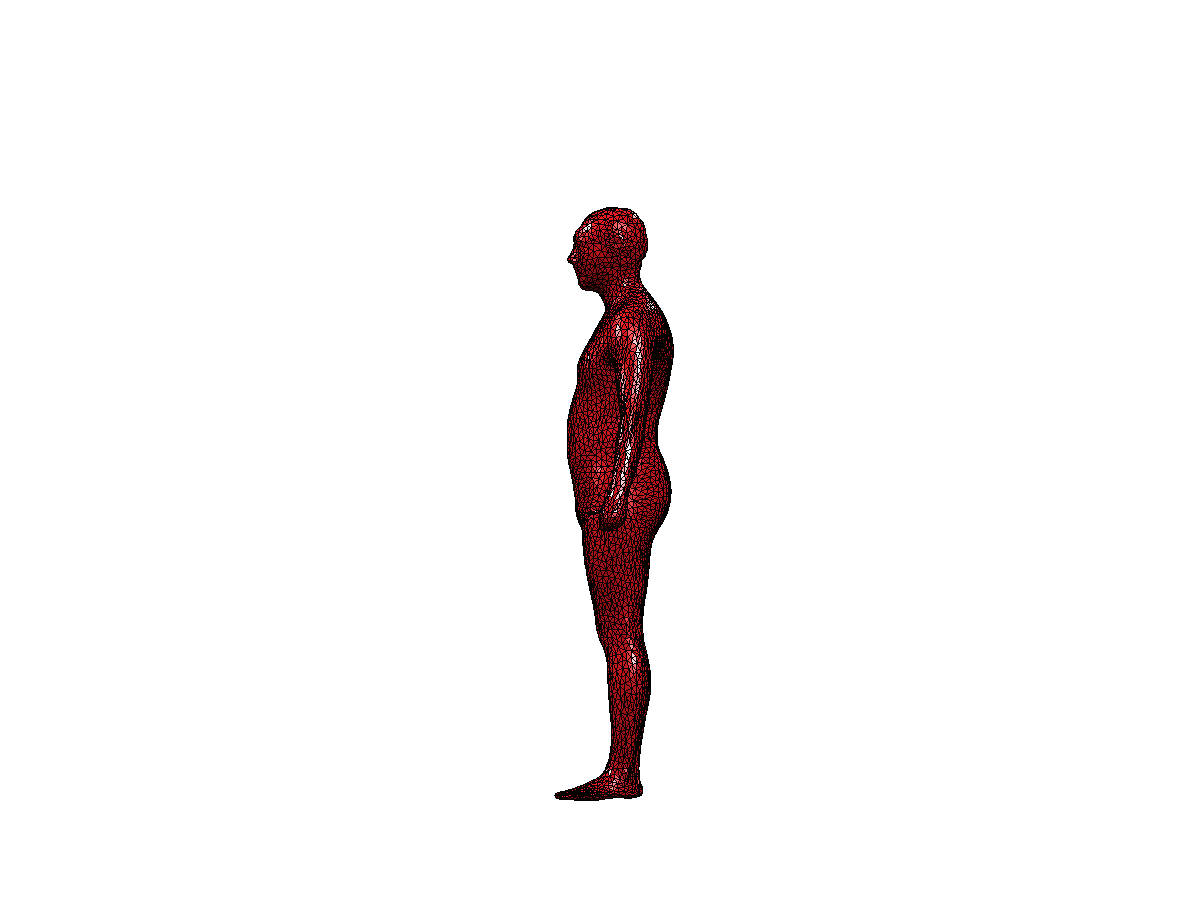}&
\quad\quad&
\includegraphics[trim=9.0cm 1cm 8.7cm 1cm, clip=true, height=\h\linewidth]{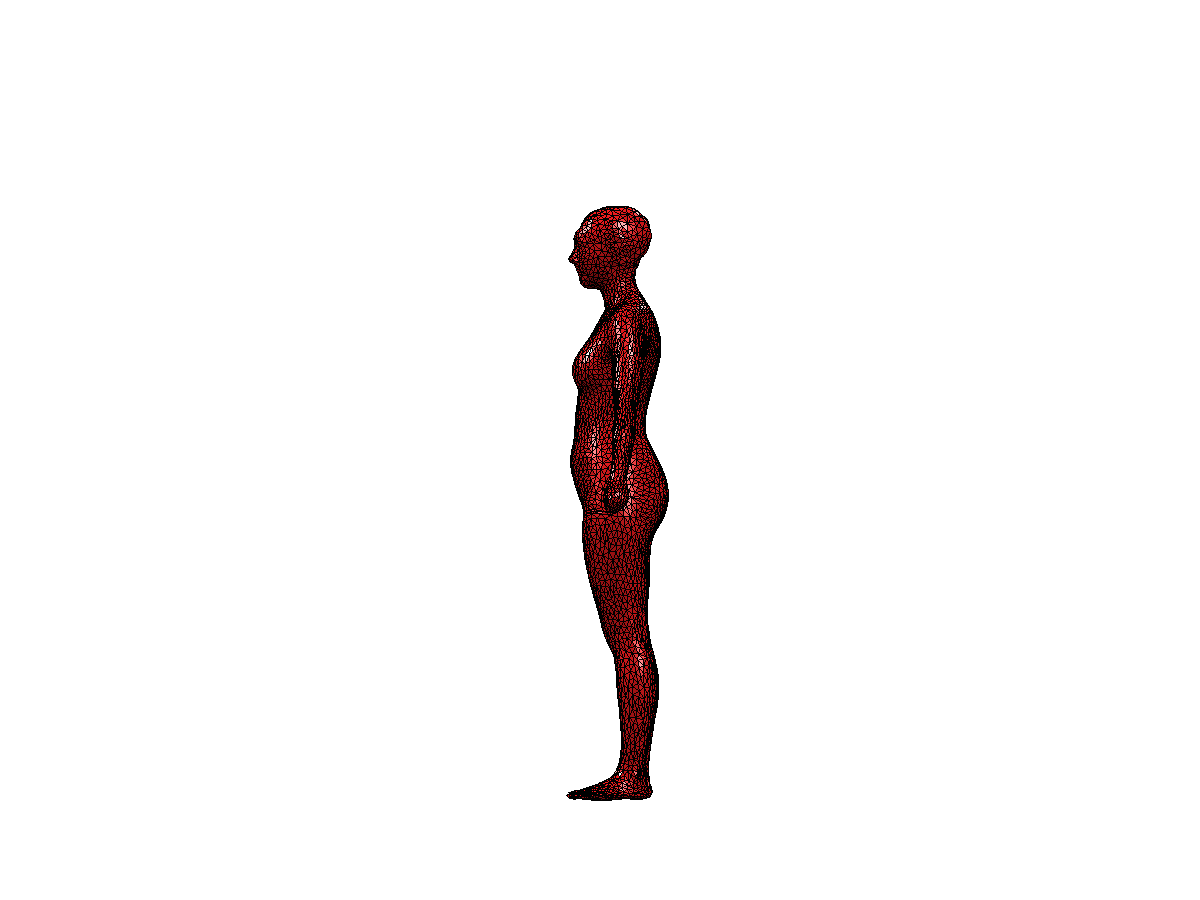}&
\includegraphics[trim=9.0cm 1cm 8.7cm 1cm, clip=true, height=\h\linewidth]{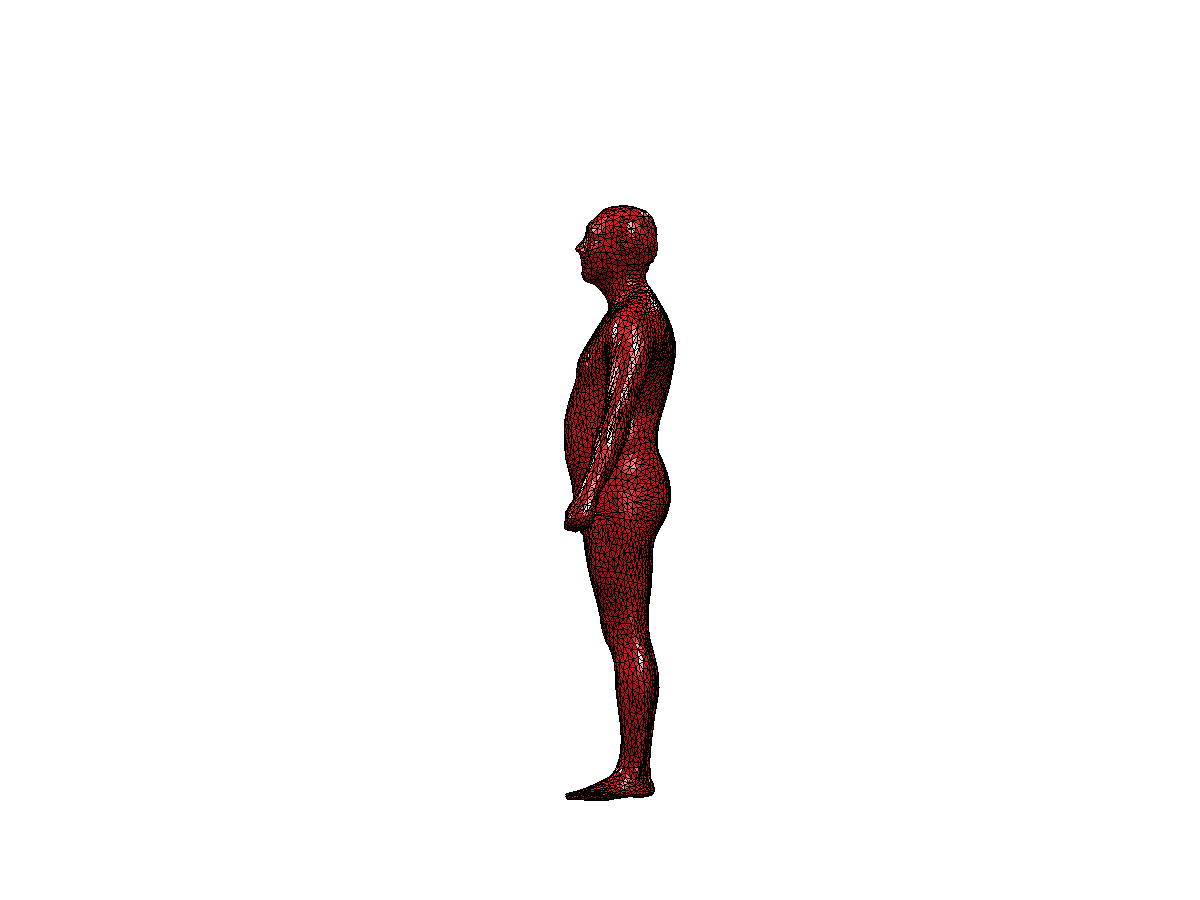}&
\quad\quad&
\includegraphics[trim=9.0cm 1cm 8.7cm 1cm, clip=true, height=\h\linewidth]{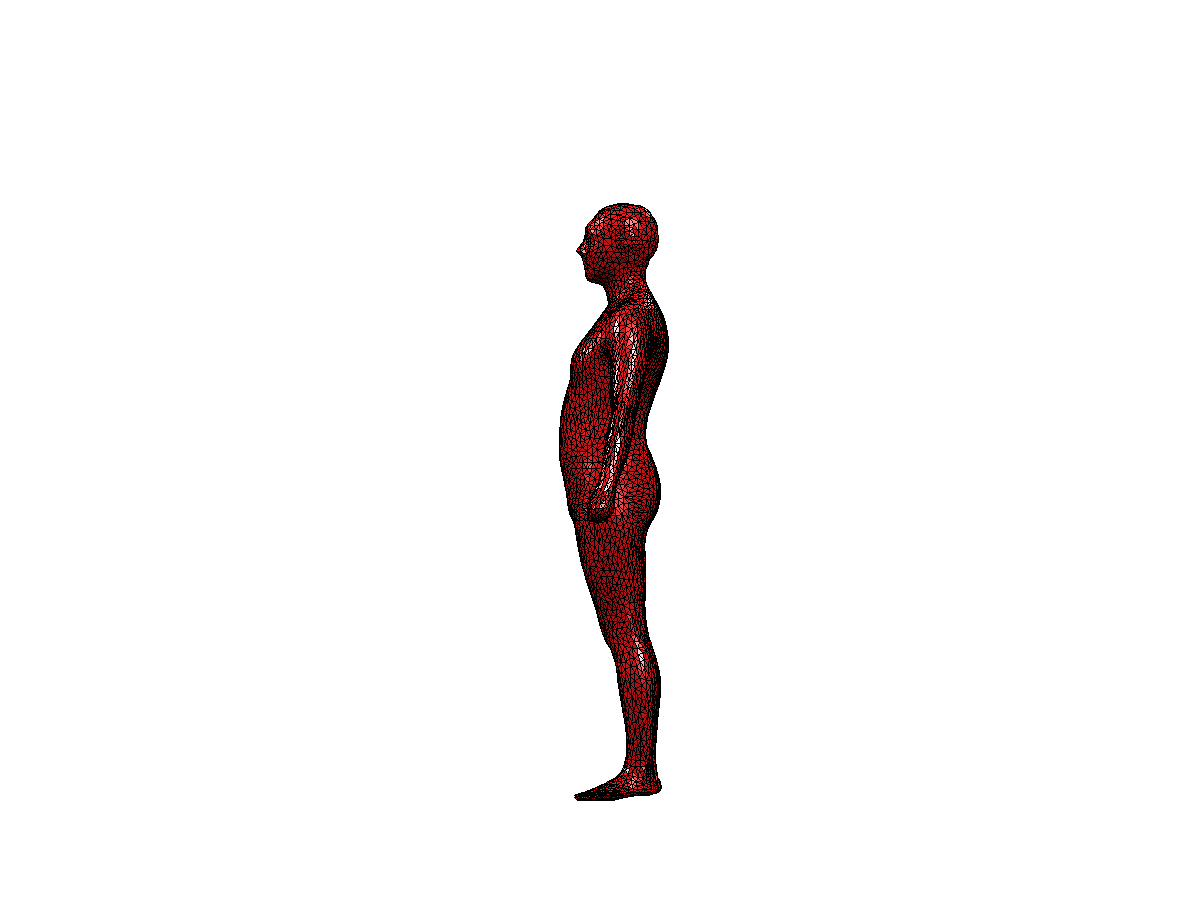}&
\includegraphics[trim=9.0cm 1cm 8.7cm 1cm, clip=true, height=\h\linewidth]{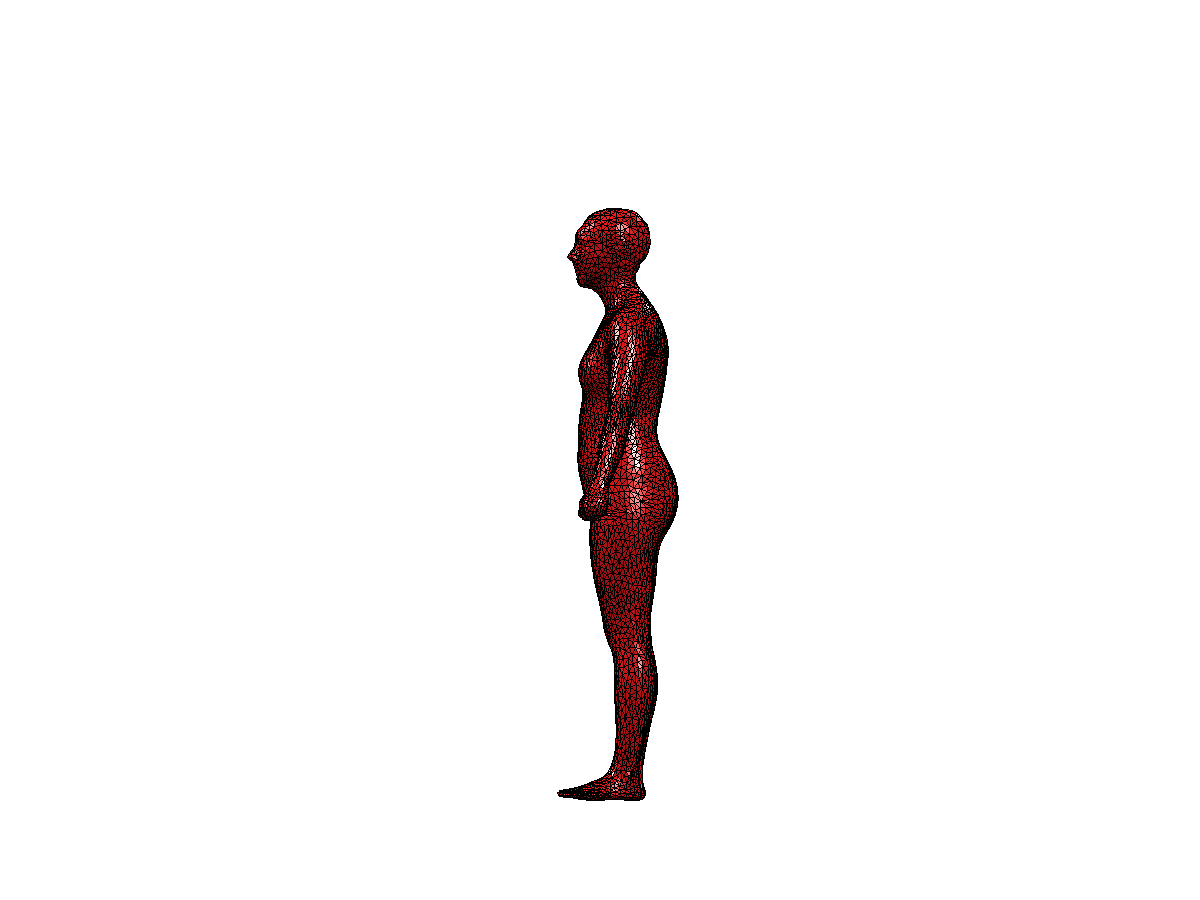}\\

\begin{sideways} \quad\quad\bf \textit \ours + \WSX \end{sideways}&
\includegraphics[trim=9.0cm 1cm 8.7cm 1cm, clip=true, height=\h\linewidth]{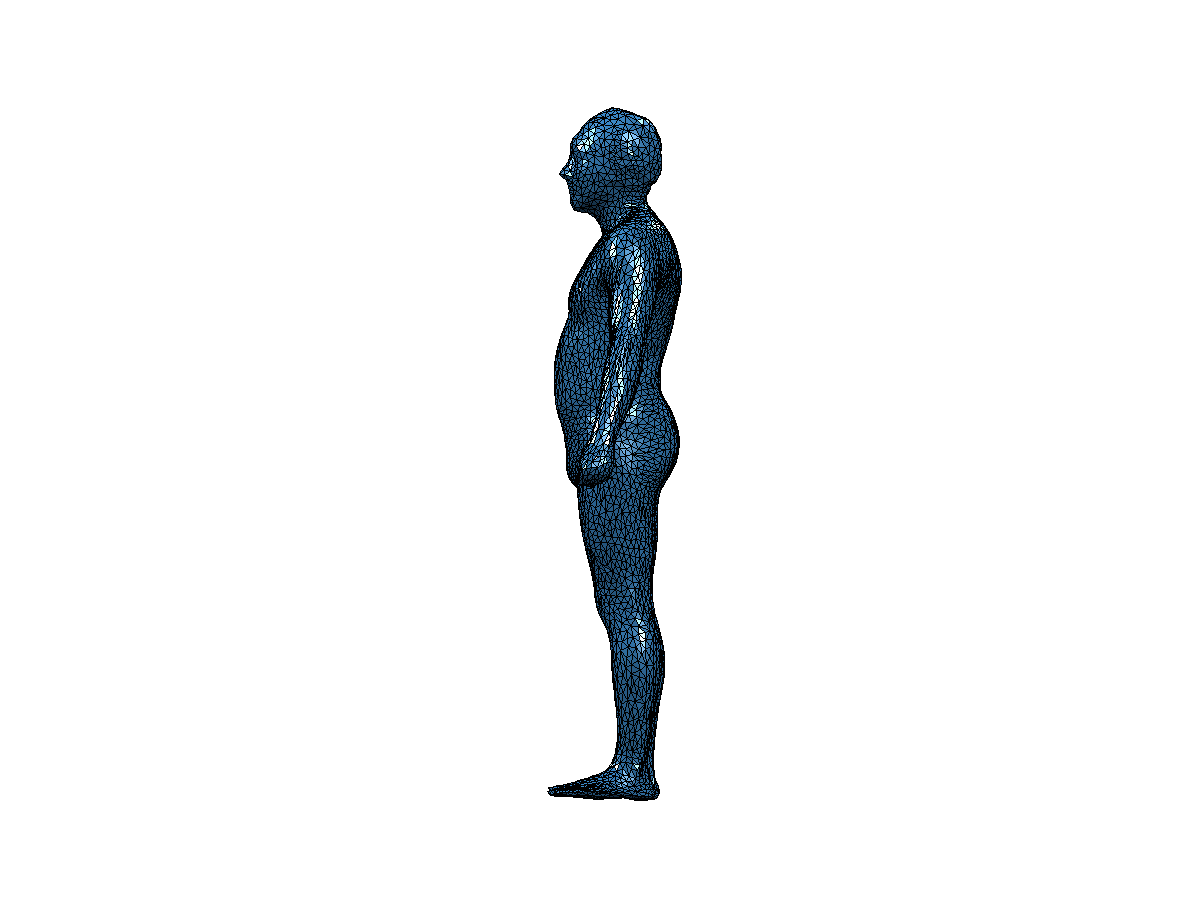}&
\includegraphics[trim=9.0cm 1cm 8.7cm 1cm, clip=true, height=\h\linewidth]{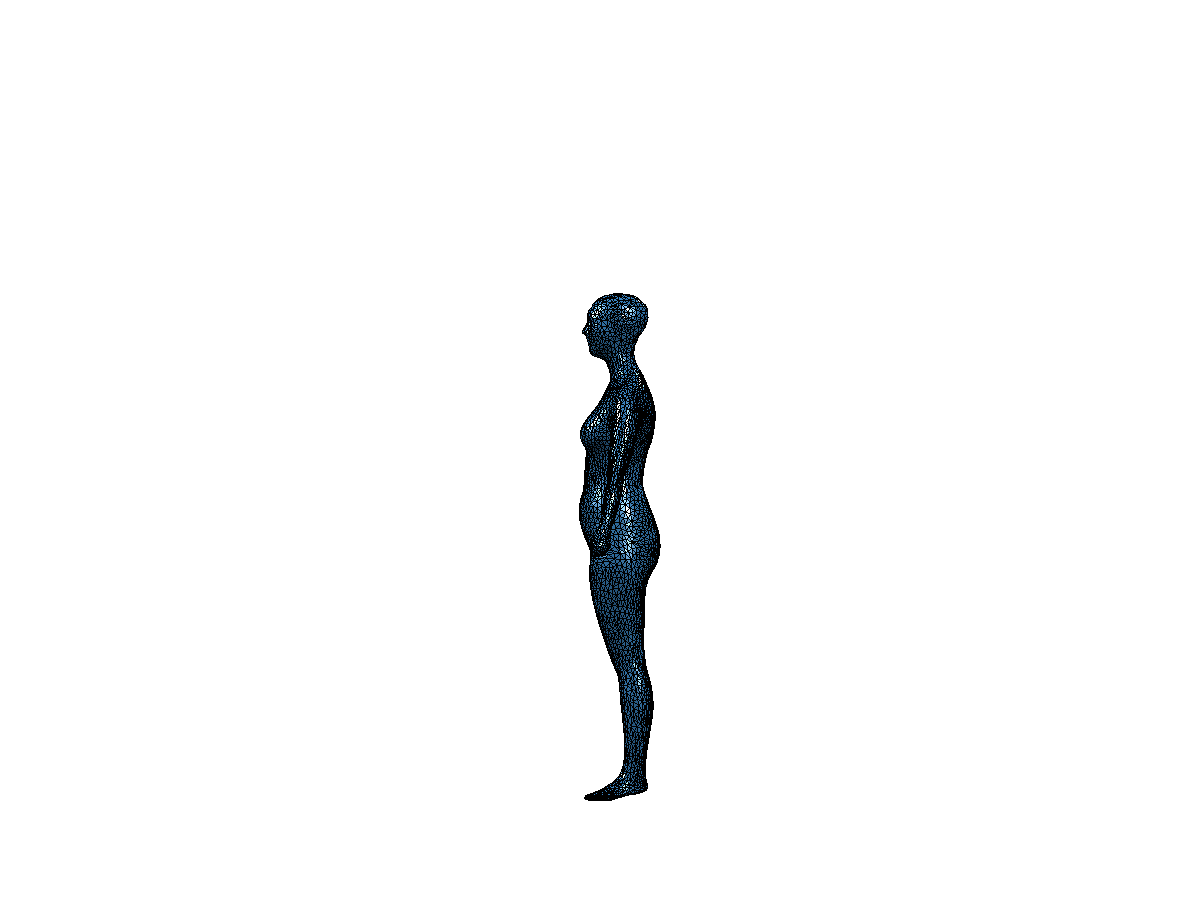}&
\quad\quad&
\includegraphics[trim=9.0cm 1cm 8.7cm 1cm, clip=true, height=\h\linewidth]{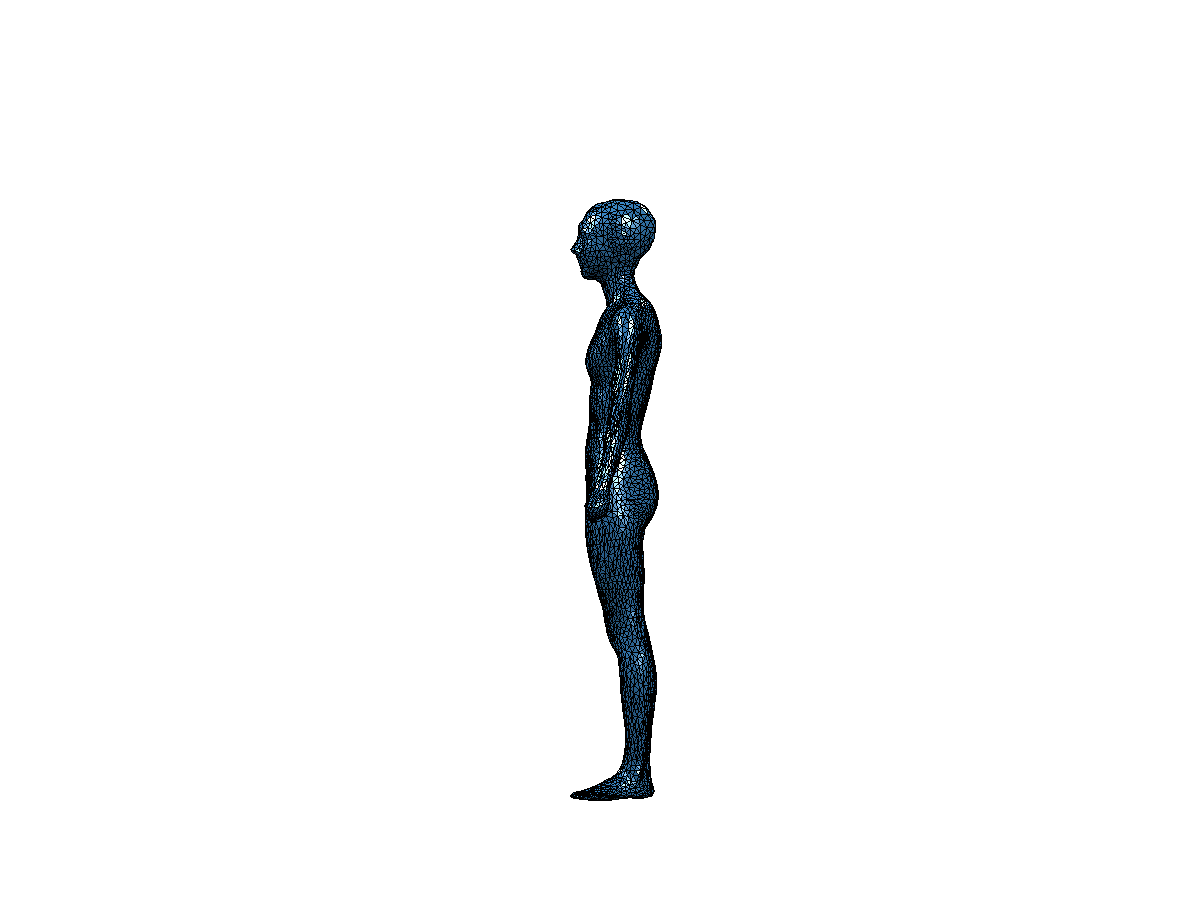}&
\includegraphics[trim=9.0cm 1cm 8.7cm 1cm, clip=true, height=\h\linewidth]{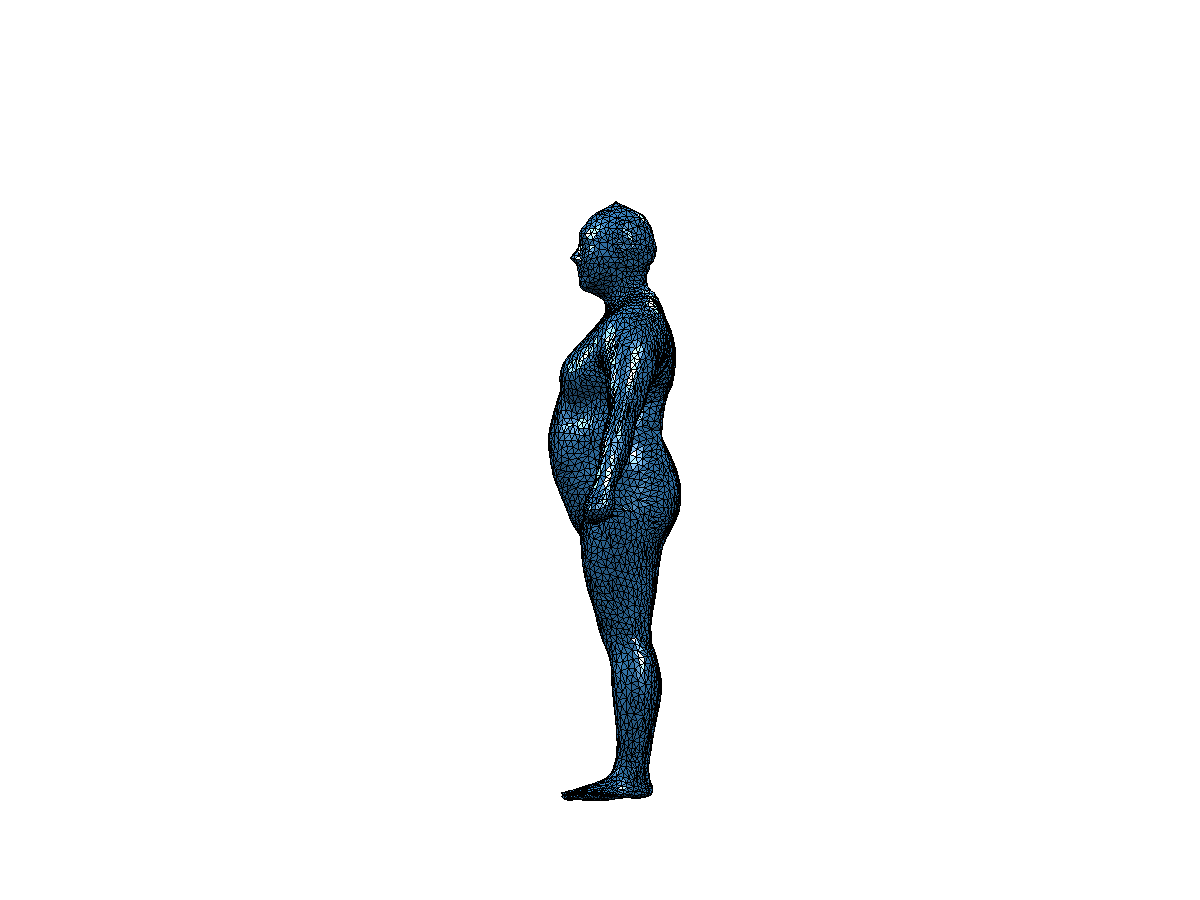}&
\quad\quad&
\includegraphics[trim=9.0cm 1cm 8.7cm 1cm, clip=true, height=\h\linewidth]{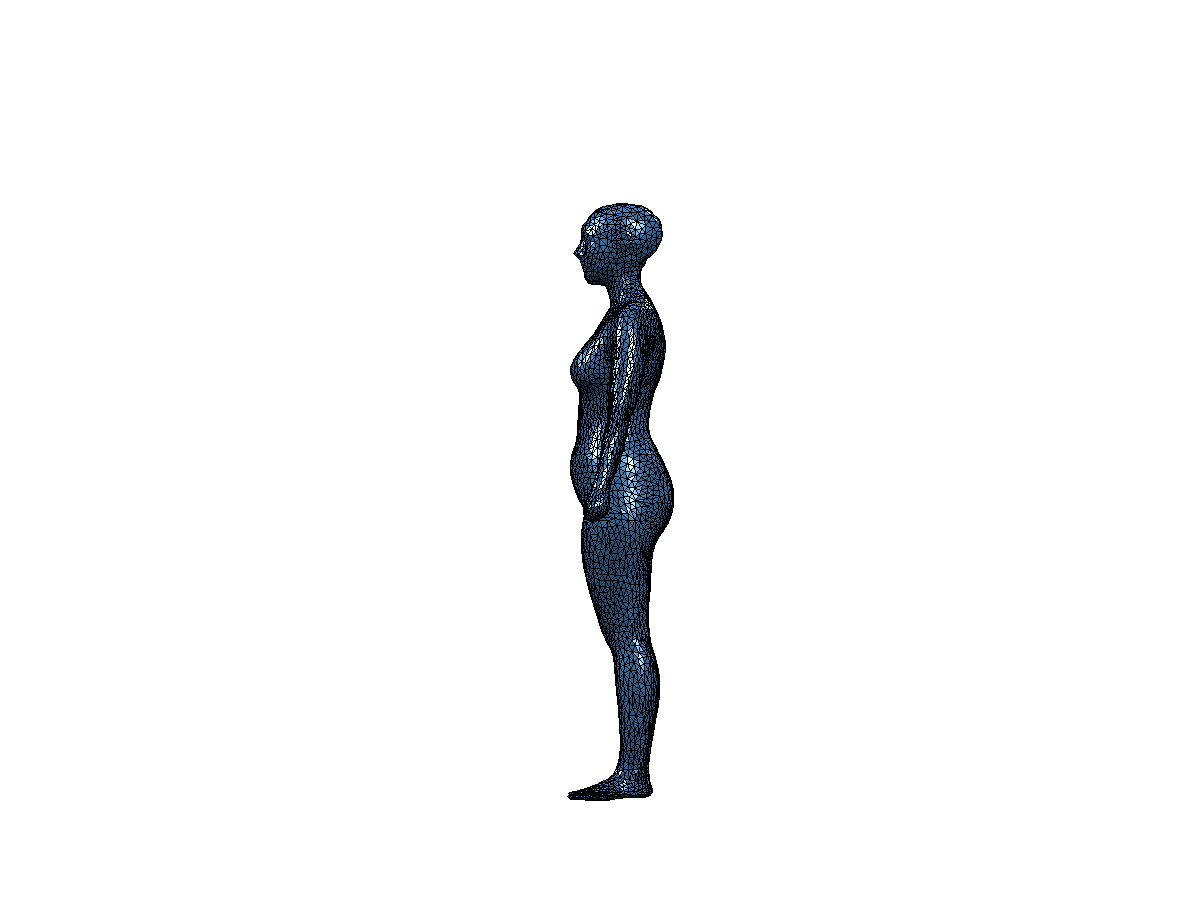}&
\includegraphics[trim=9.0cm 1cm 8.7cm 1cm, clip=true, height=\h\linewidth]{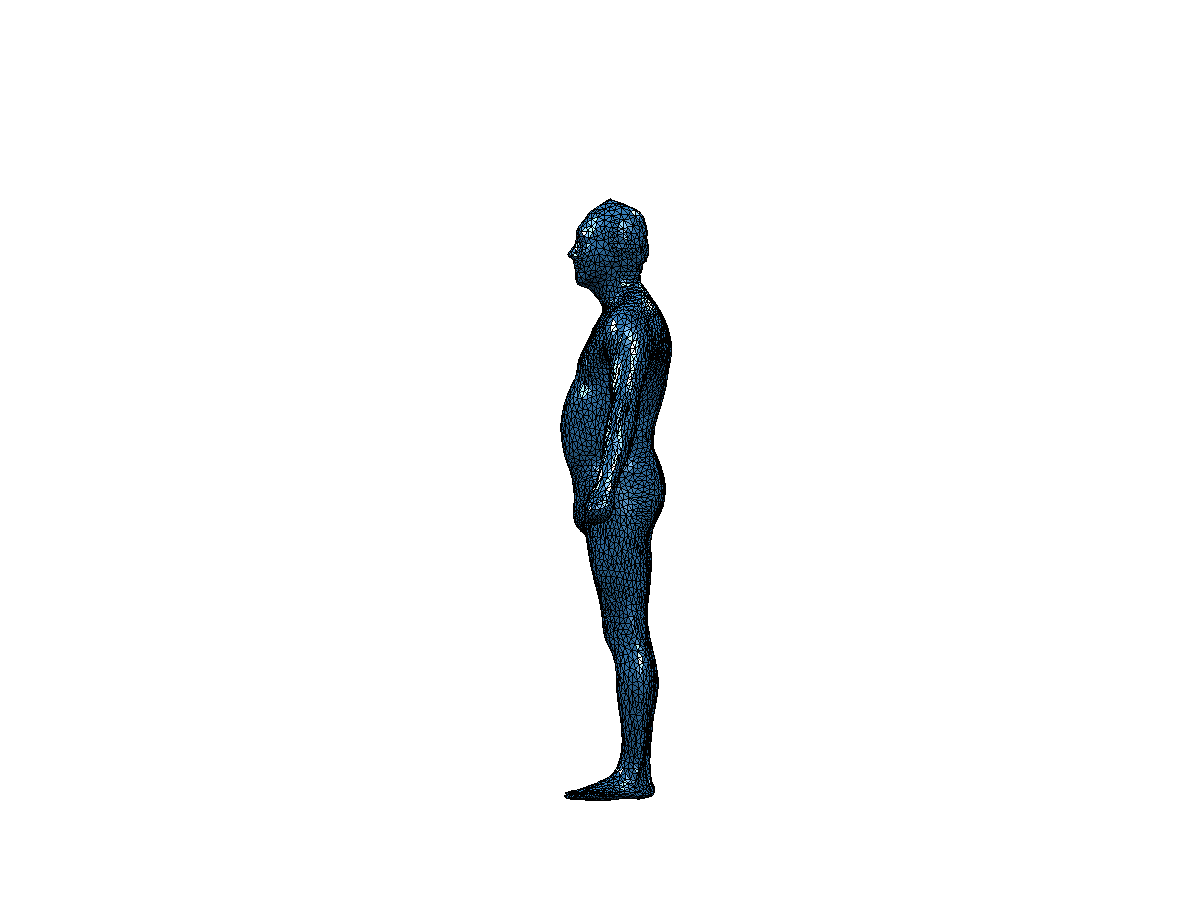}&
\quad\quad&
\includegraphics[trim=9.0cm 1cm 8.7cm 1cm, clip=true, height=\h\linewidth]{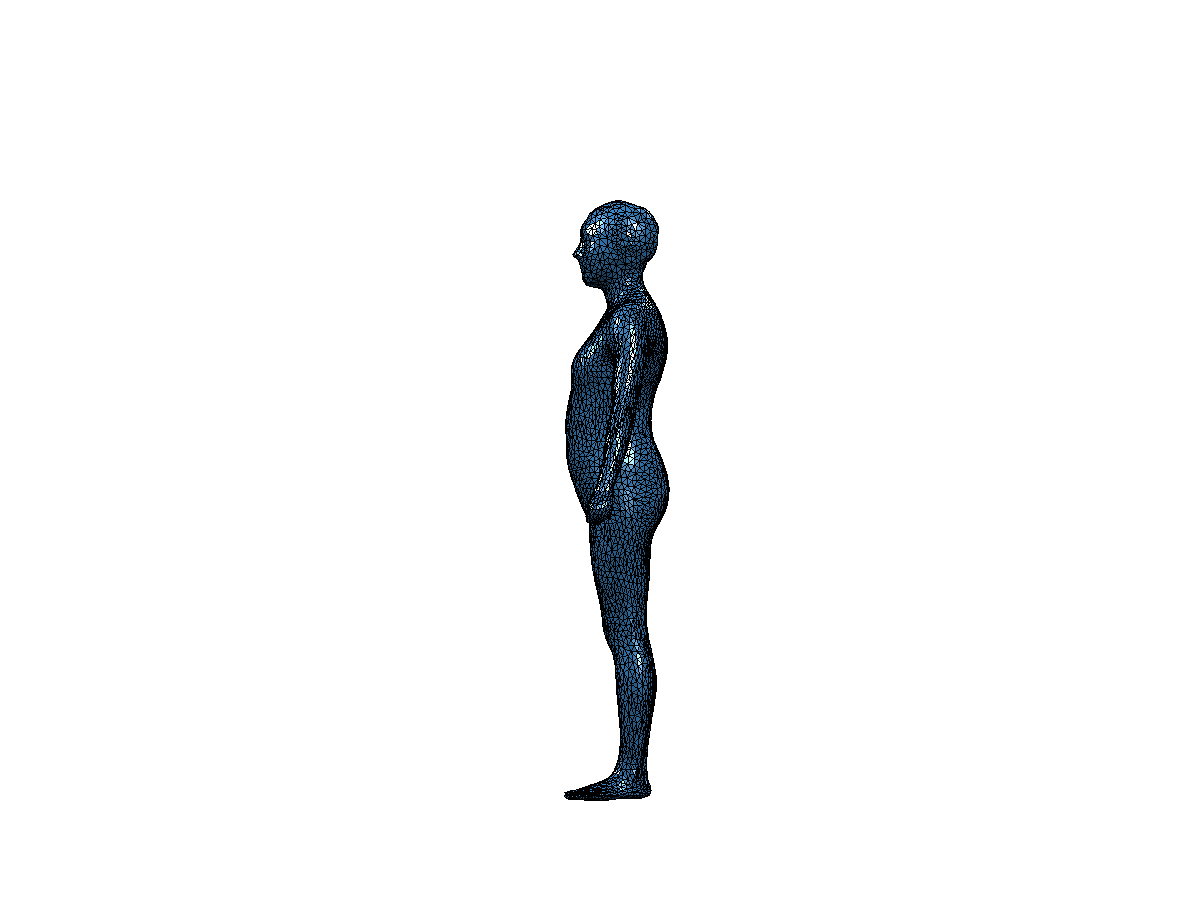}&
\includegraphics[trim=9.0cm 1cm 8.7cm 1cm, clip=true, height=\h\linewidth]{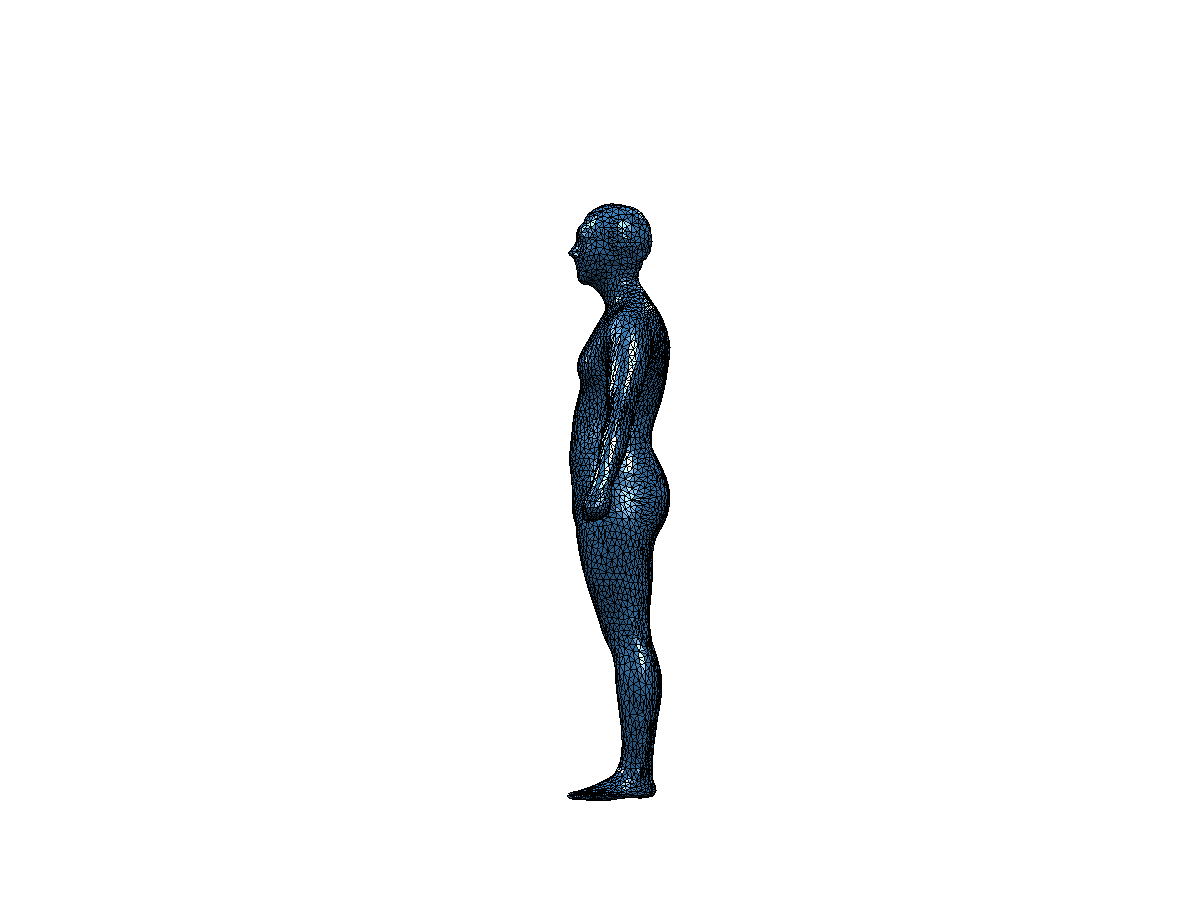}&
\quad\quad&
\includegraphics[trim=9.0cm 1cm 8.7cm 1cm, clip=true, height=\h\linewidth]{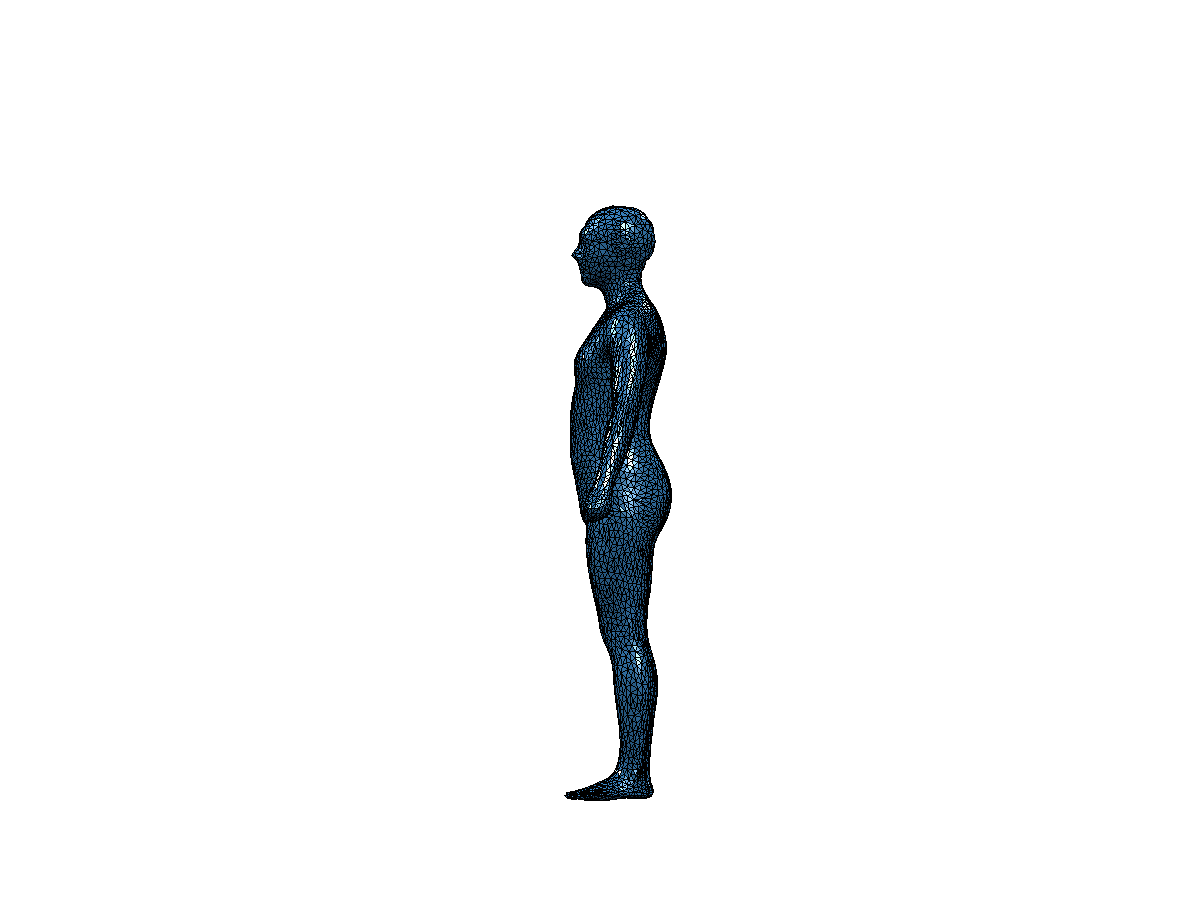}&
\includegraphics[trim=9.0cm 1cm 8.7cm 1cm, clip=true, height=\h\linewidth]{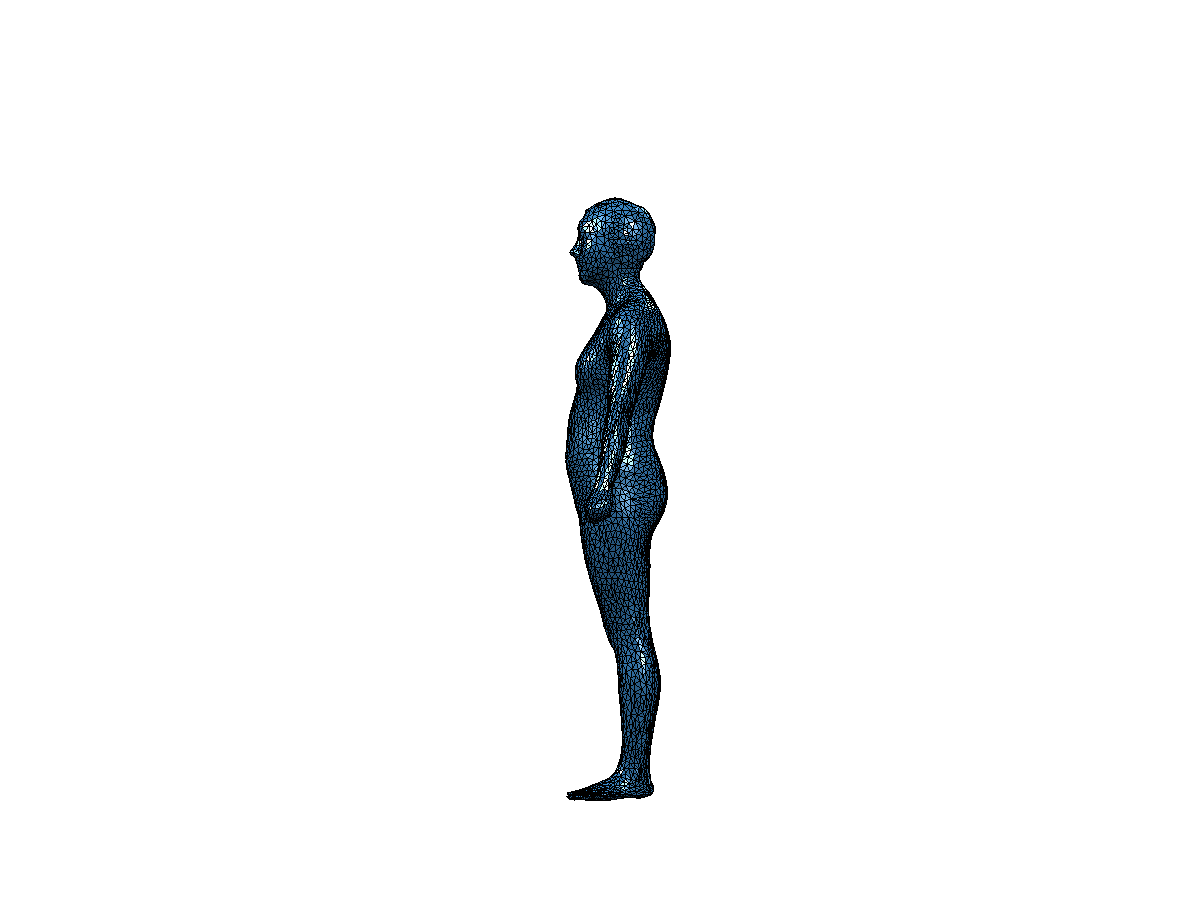}\\

\begin{sideways}\quad\quad\quad \bf \textit \ours + \NH \end{sideways}&
\includegraphics[trim=9.0cm 1cm 8.7cm 1cm, clip=true, height=\h\linewidth]{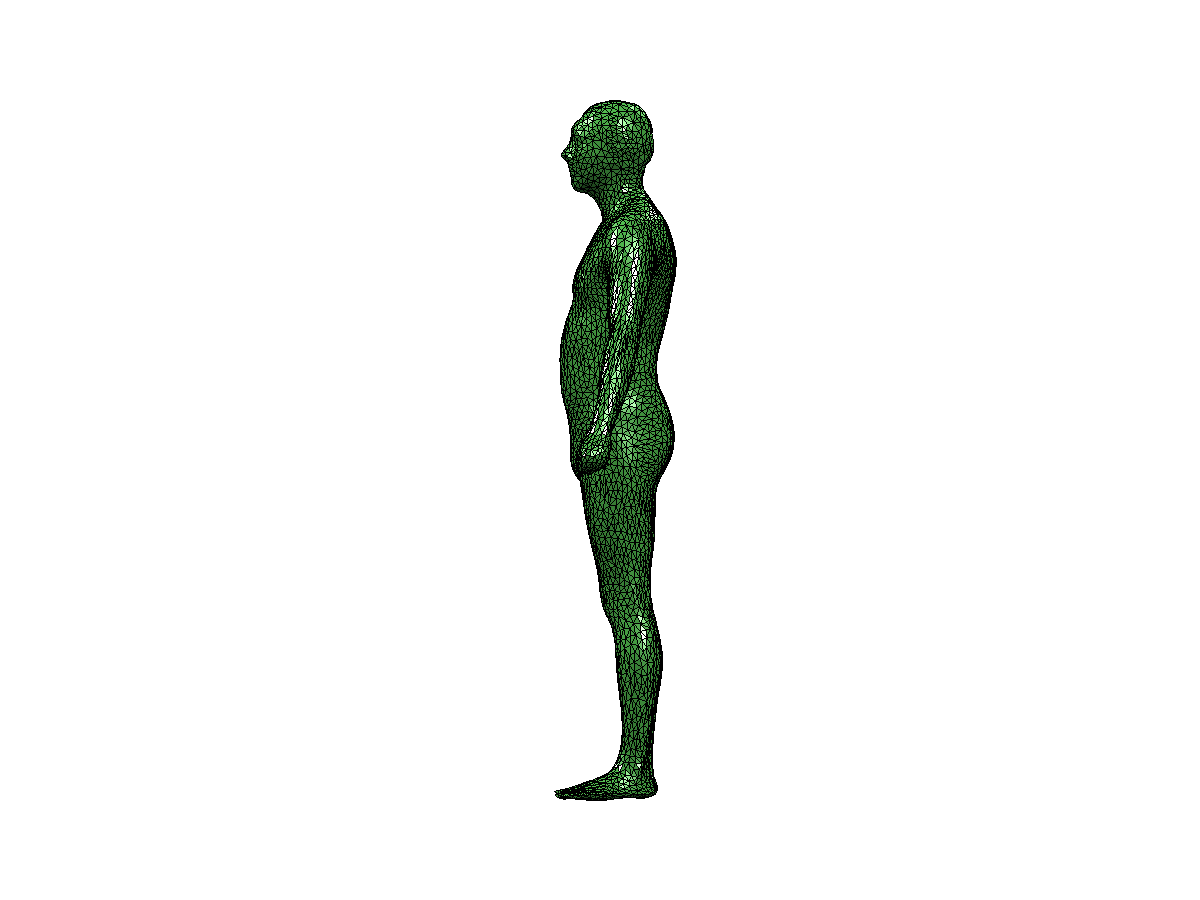}&
\includegraphics[trim=9.0cm 1cm 8.7cm 1cm, clip=true, height=\h\linewidth]{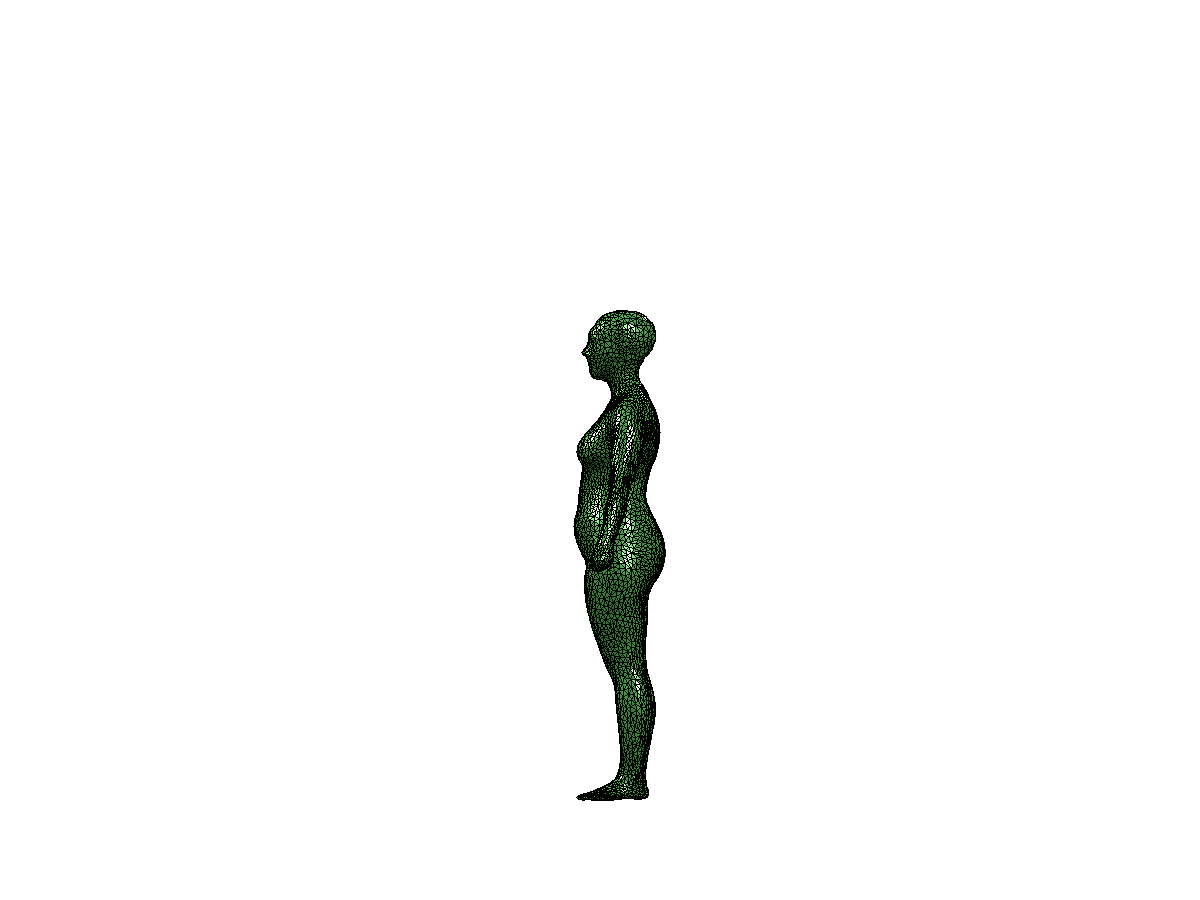}&
\quad\quad&
\includegraphics[trim=9.0cm 1cm 8.7cm 1cm, clip=true, height=\h\linewidth]{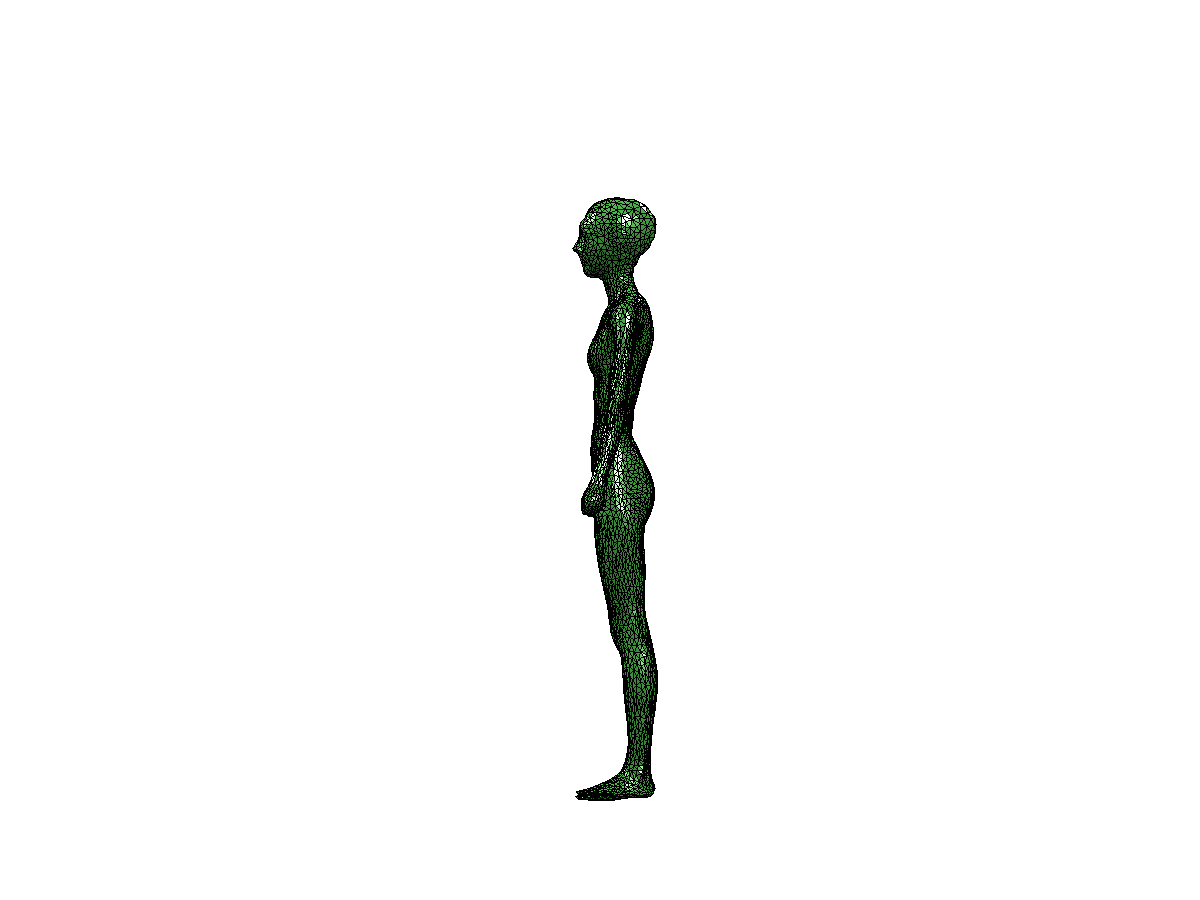}&
\includegraphics[trim=9.0cm 1cm 8.7cm 1cm, clip=true, height=\h\linewidth]{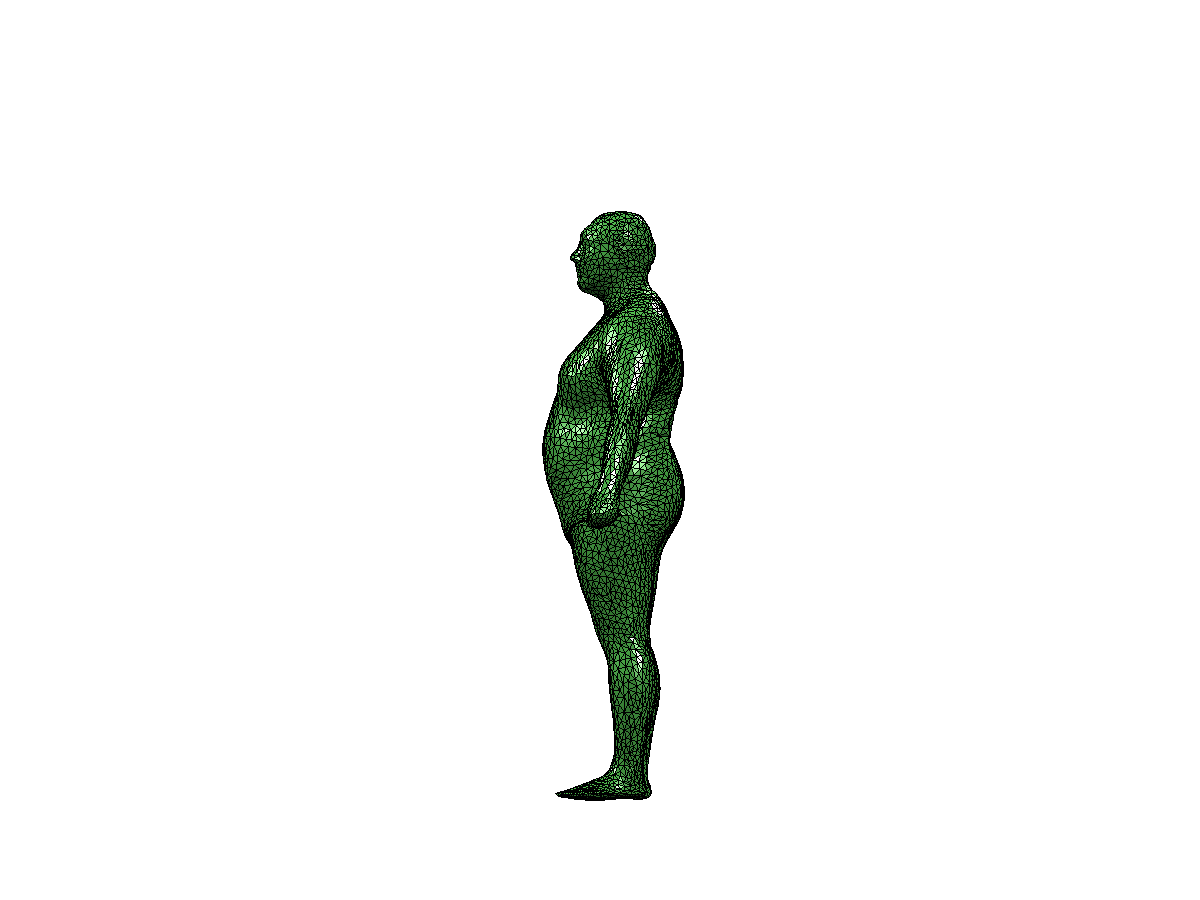}&
\quad\quad&
\includegraphics[trim=9.0cm 1cm 8.7cm 1cm, clip=true, height=\h\linewidth]{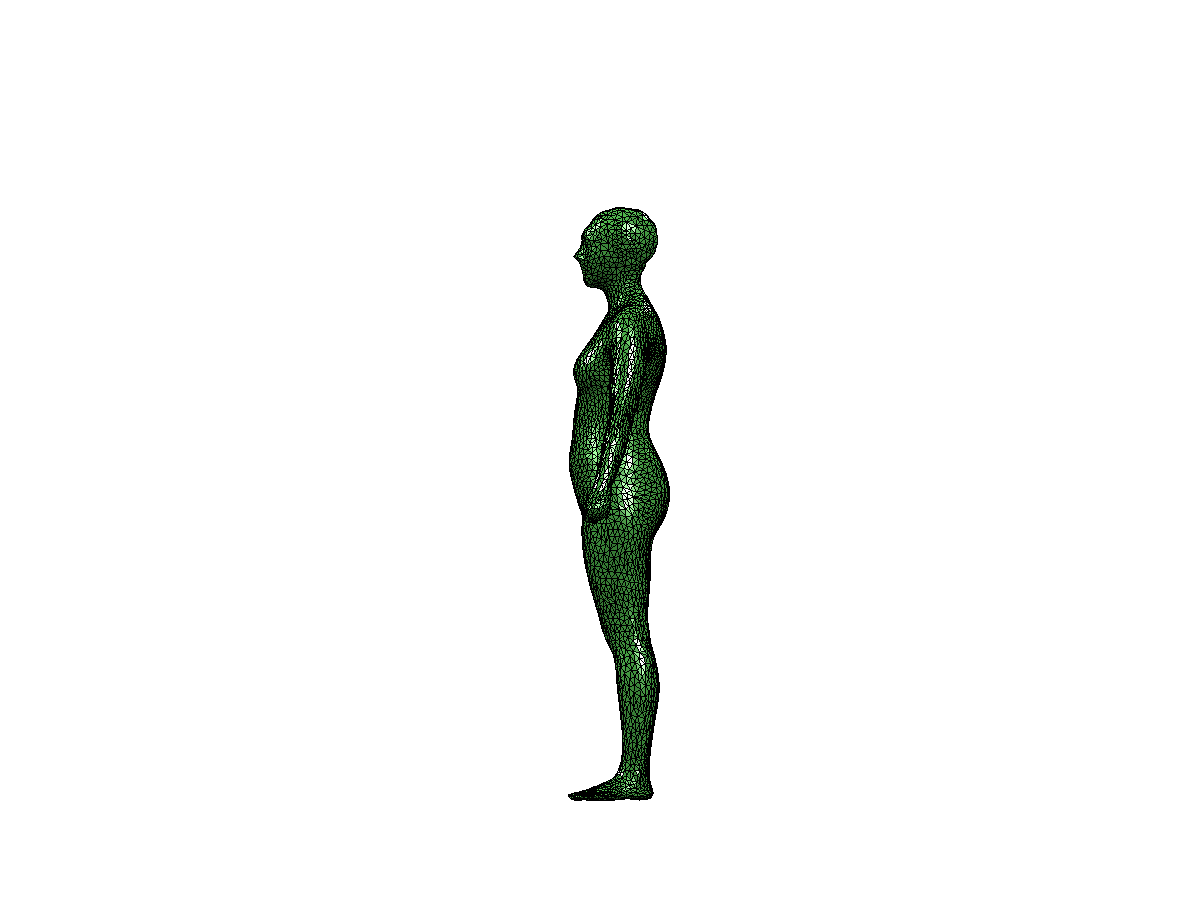}&
\includegraphics[trim=9.0cm 1cm 8.7cm 1cm, clip=true, height=\h\linewidth]{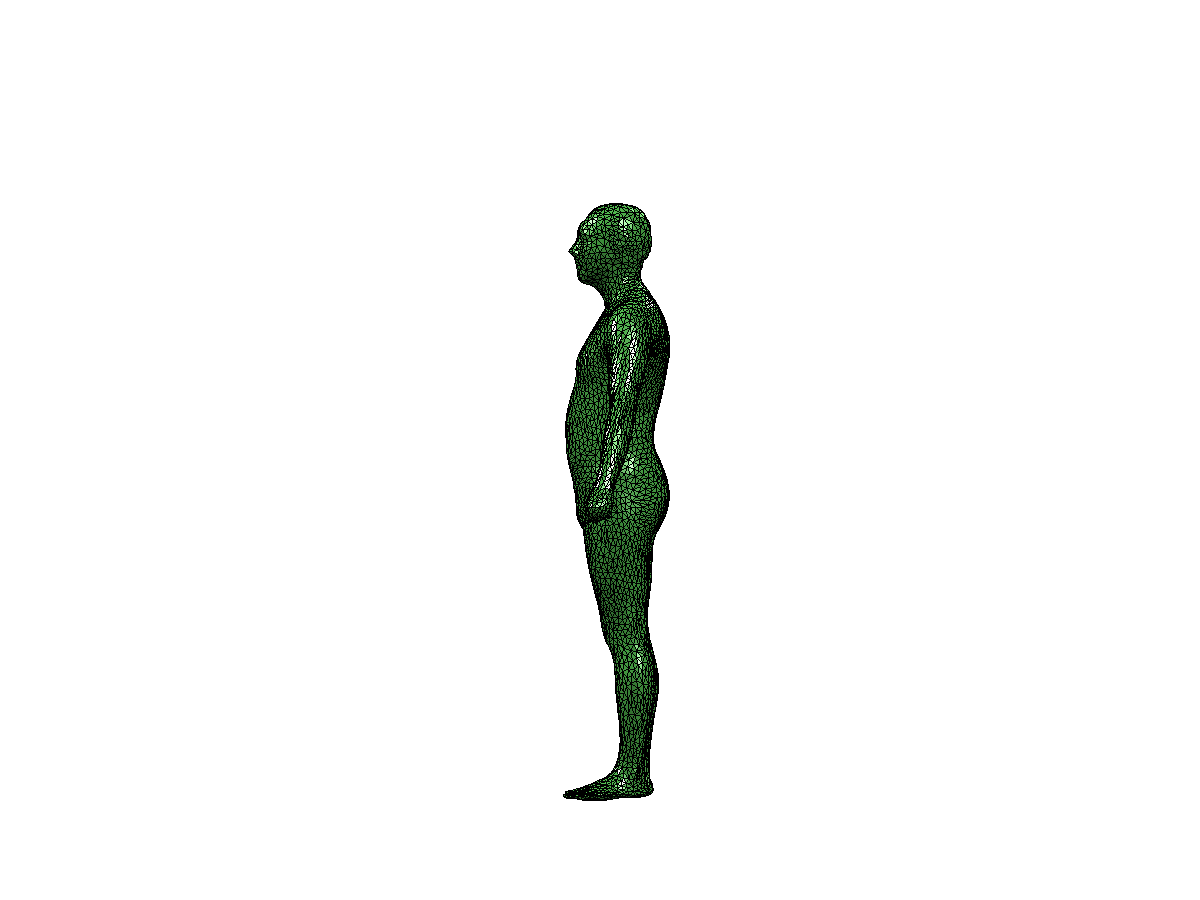}&
\quad\quad&
\includegraphics[trim=9.0cm 1cm 8.7cm 1cm, clip=true, height=\h\linewidth]{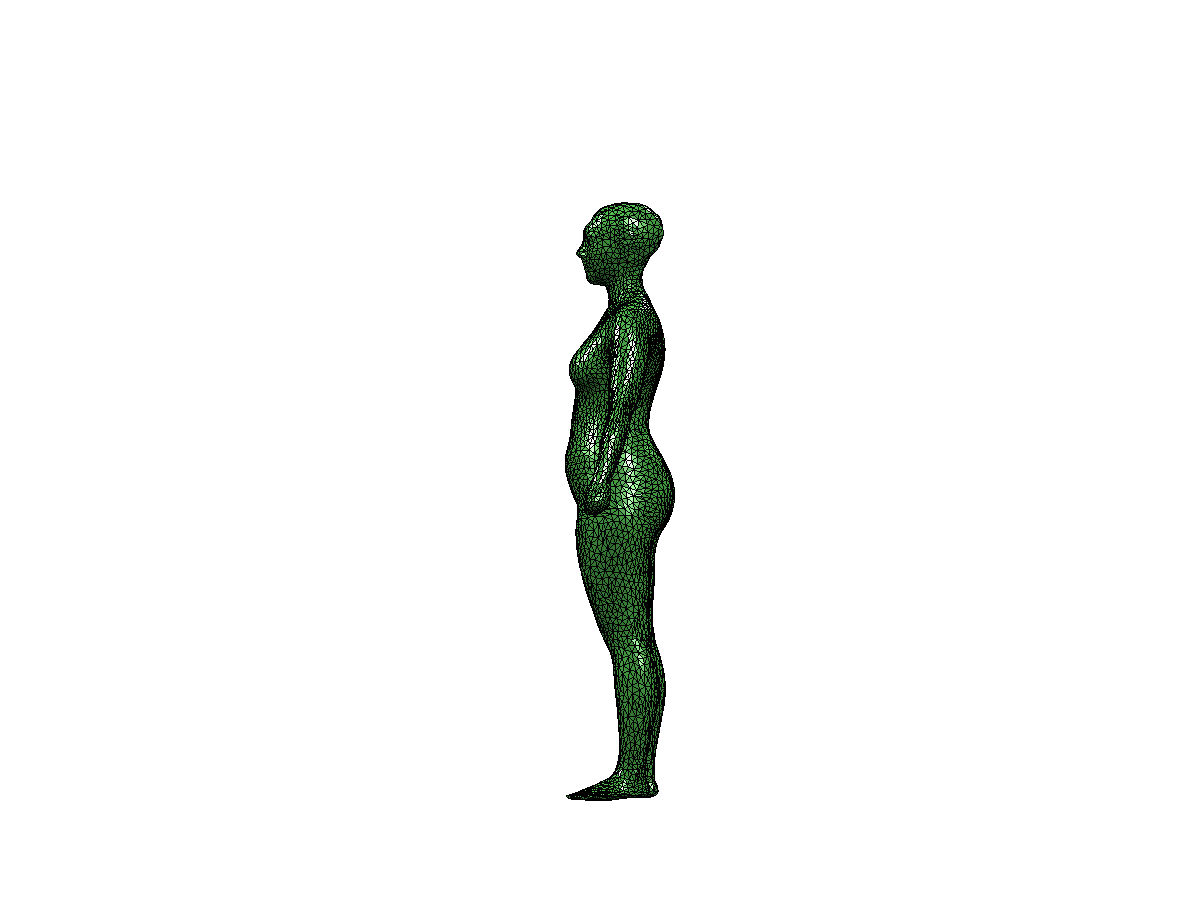}&
\includegraphics[trim=9.0cm 1cm 8.7cm 1cm, clip=true, height=\h\linewidth]{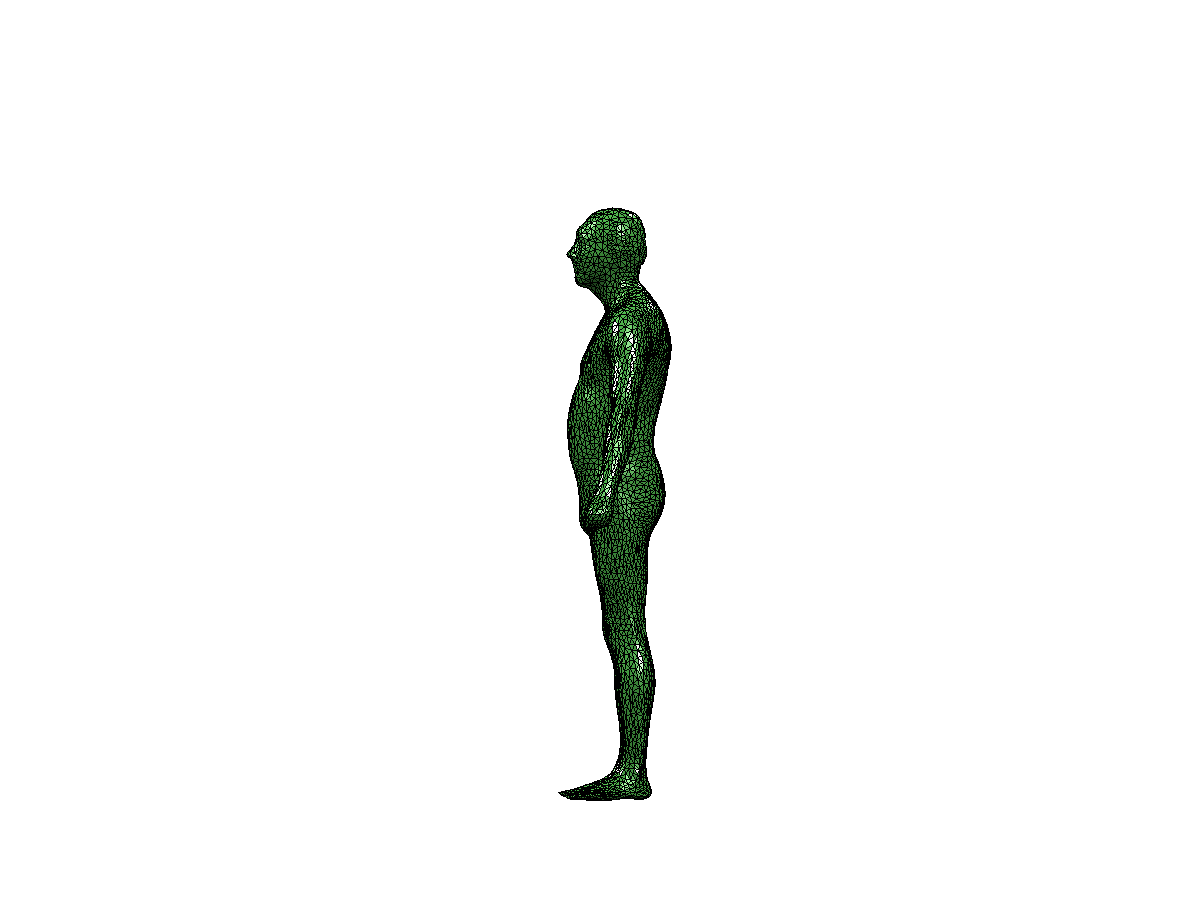}&
\quad\quad&
\includegraphics[trim=9.0cm 1cm 8.7cm 1cm, clip=true, height=\h\linewidth]{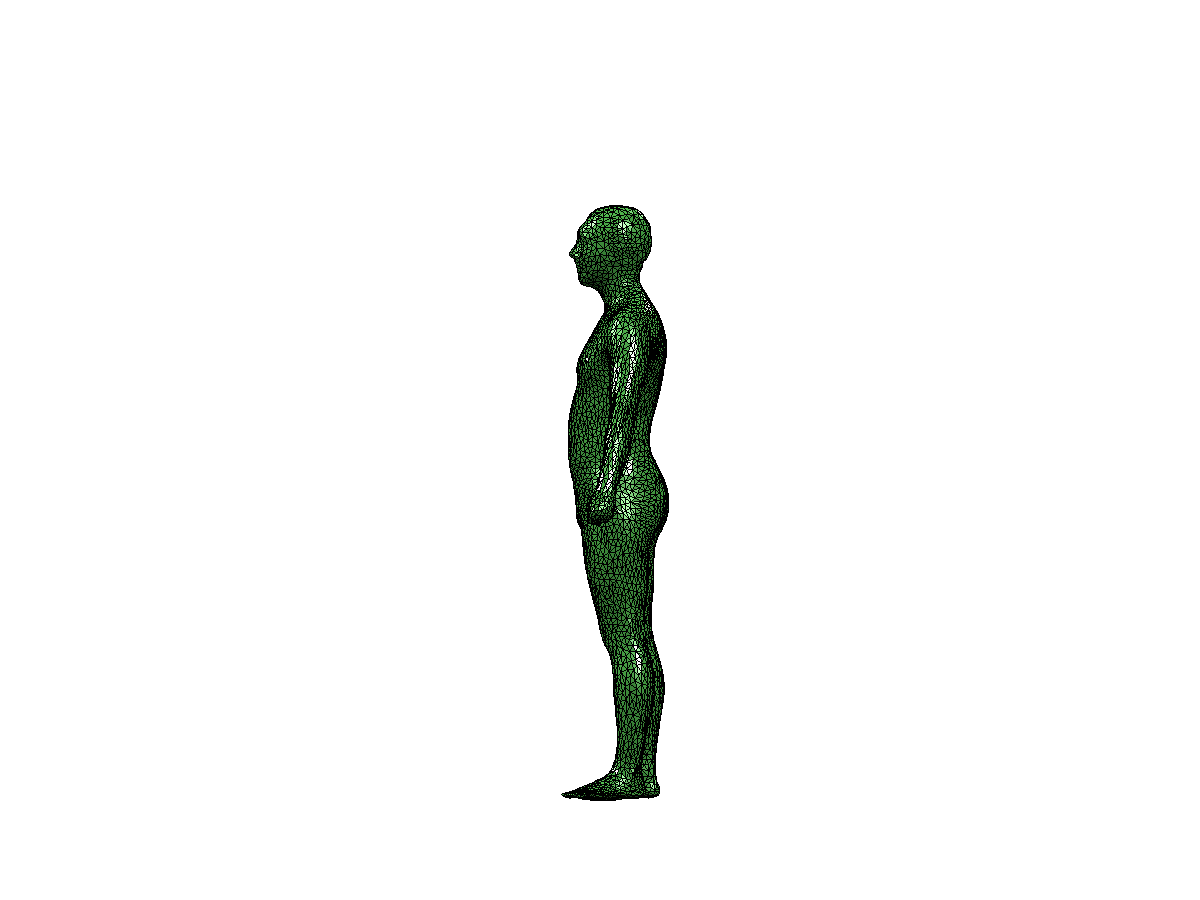}&
\includegraphics[trim=9.0cm 1cm 8.7cm 1cm, clip=true, height=\h\linewidth]{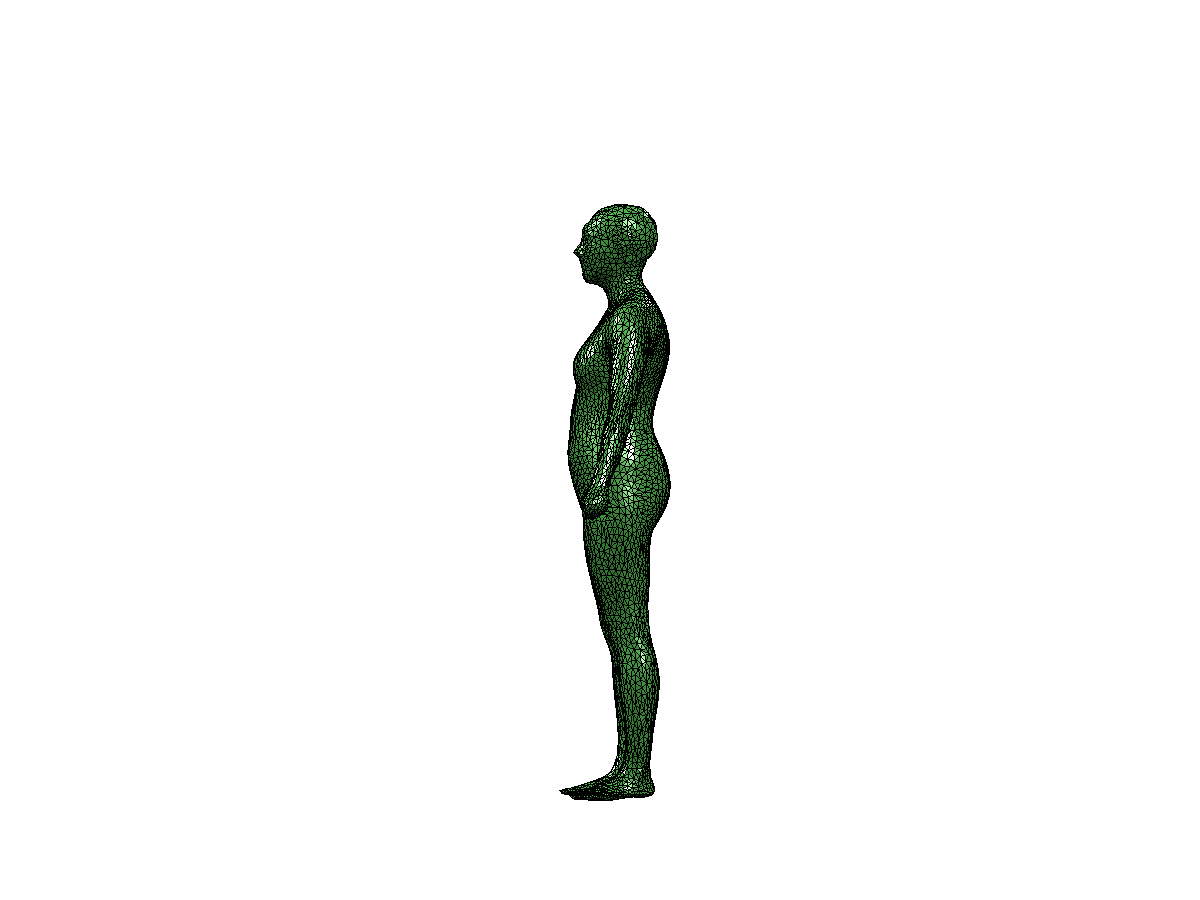}\\

\bf\small st. dev.&
\small$+3 \sigma$& \small$-3 \sigma$&&
\small$+3 \sigma$& \small$-3 \sigma$&&
\small$+3 \sigma$& \small$-3 \sigma$&&
\small$+3 \sigma$& \small$-3 \sigma$&&
\small$+3 \sigma$& \small$-3 \sigma$\\[5pt]

\bf\small PCA id&
\multicolumn{2}{c}{\small I}&&
\multicolumn{2}{c}{\small II}&&
\multicolumn{2}{c}{\small III}&&
\multicolumn{2}{c}{\small IV}&&
\multicolumn{2}{c}{\small V}\\

\end{tabular}
\caption{Visualization of the first five PCA eigenvectors scaled by
  $\pm3 \sigma$ (standard deviation). Shown are eigenvectors of the
  \jain~space~\cite{Jain:2010:MovieReshape} (row 1) and the
  \jain~spaces trained using our pre-processed data without (row 2)
  and with posture normalization using \WSX~\cite{WuhrerPIS12} (row 3)
  and \NH~\cite{Neophytou2013} (row 4).}
  \label{fig:model_vis_pca}
\egroup
\end{figure}

\section{Human body reconstruction}
\label{sec:eval_reconstruction}

Finally, we evaluate our improved \jain~spaces on the task of
estimating human body shape from sparse and noisy visual input. We
follow the approach by Helten et al.~\cite{HeltenPAE13} to estimate
the body shape of a person from two sequentially taken front and back
depth images. First, body shape and posture are fitted independently
to each depth image. Second, the obtained results are used as
initialization of a method that jointly optimizes over shape and
independently optimizes over posture parameters. This optimization
strategy is used because the shape in both depth scans is of the same
person, but the pose may differ.

\subsection{Dataset and experimental setup} 
\begin{figure}[t]
\centering
\bgroup
\tabcolsep 0.0pt
\renewcommand{\arraystretch}{0.0}

%\subfigure[\footnotesize front/back depth scans]{
\begin{tabular}{cccc cccc}
\begin{sideways} \small \bf \quad\quad Female 1 \end{sideways}&$\;$
\includegraphics[height=0.205\linewidth]{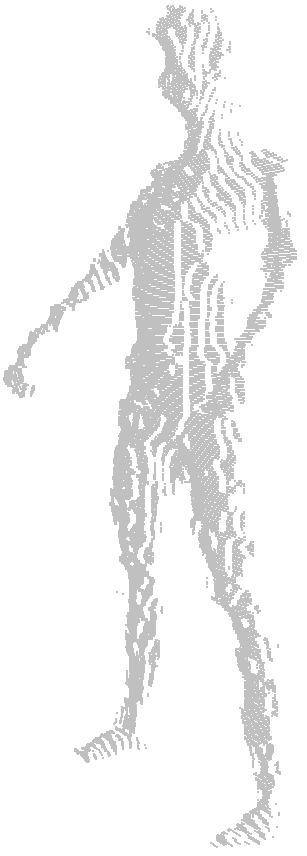}&
\includegraphics[height=0.205\linewidth]{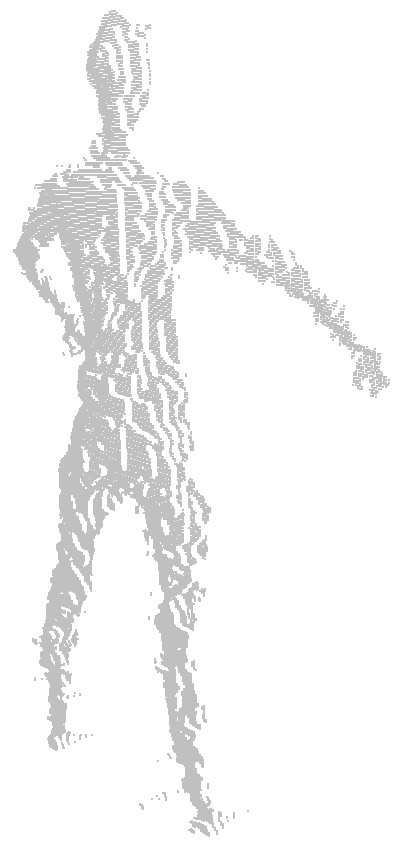}&
\includegraphics[trim=10cm 0cm 0cm
0cm,clip=true,height=0.205\linewidth]{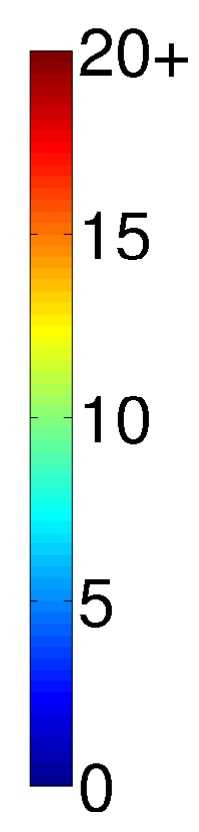}&
\includegraphics[trim=0cm 1.8cm 8.0cm
1.8cm,clip=true,height=0.205\linewidth]{kinect-mpii-no-hip-sho-tho-ankle-scan-new-scanidx-1-thresh1-0-thresh2-20-meanDist-45.pdf}&
\includegraphics[trim=0cm 1.8cm 8.0cm 1.8cm,clip=true,height=0.205\linewidth]{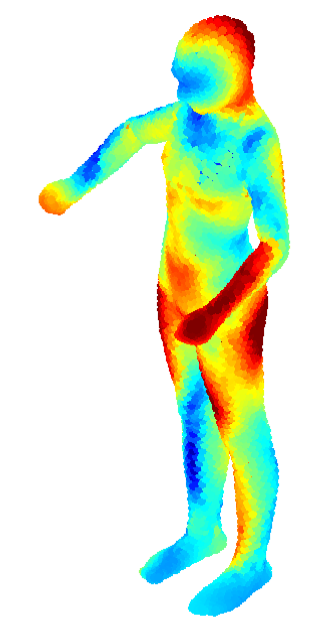}&
\includegraphics[trim=0cm 1.8cm 8.0cm 1.8cm,clip=true,height=0.205\linewidth]{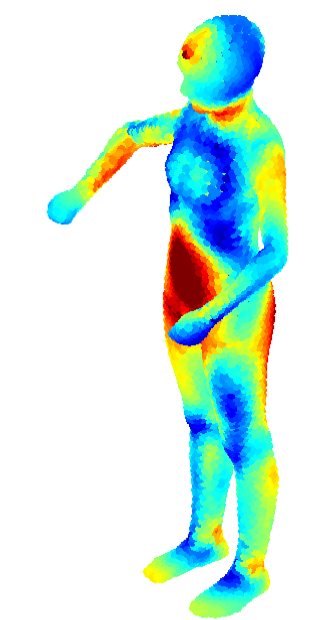}&
\includegraphics[trim=0cm 1.8cm 8.0cm
1.8cm,clip=true,height=0.205\linewidth]{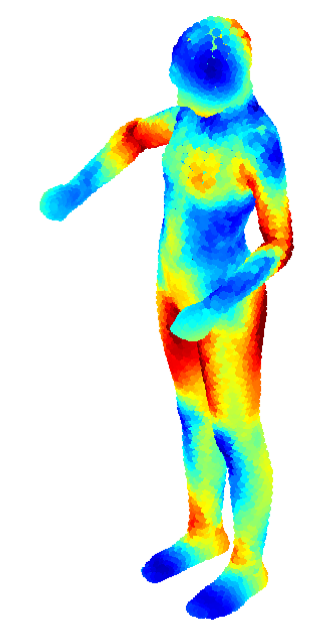}\\

\begin{sideways} \small \bf \quad\quad Male 1 \end{sideways}&$\;$
\includegraphics[height=0.205\linewidth]{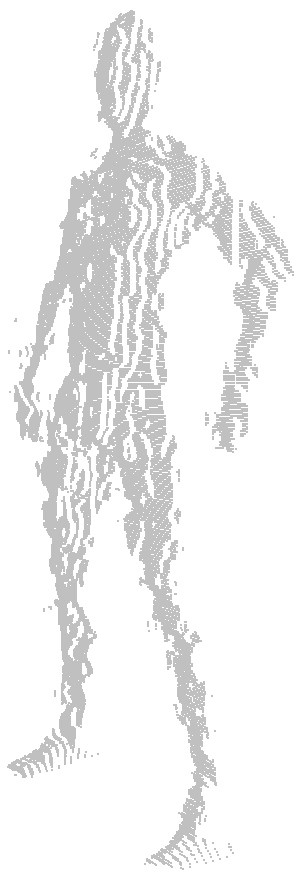}&
\includegraphics[height=0.205\linewidth]{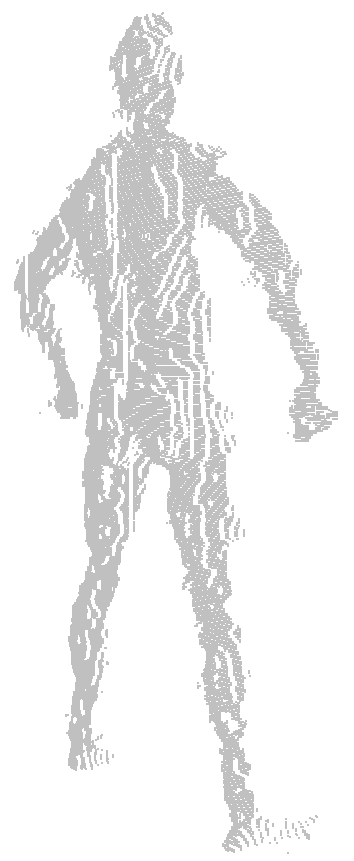}&
\includegraphics[trim=10cm 0cm 0cm
0cm,clip=true,height=0.205\linewidth]{kinect-mpii-no-hip-sho-tho-ankle-scan-new-scanidx-1-thresh1-0-thresh2-20-meanDist-45.pdf}&
\includegraphics[trim=0cm 1.8cm 8.0cm 1.8cm,clip=true,height=0.205\linewidth]{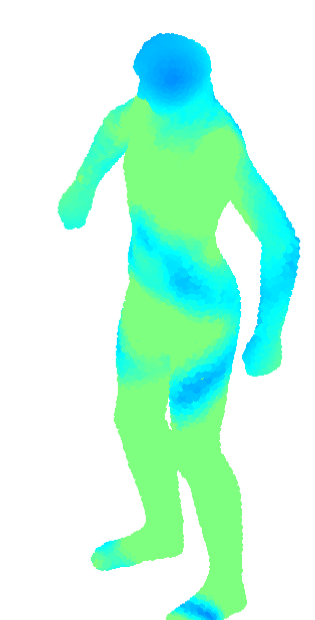}&
\includegraphics[trim=0cm 1.8cm 8.0cm
1.8cm,clip=true,height=0.205\linewidth]{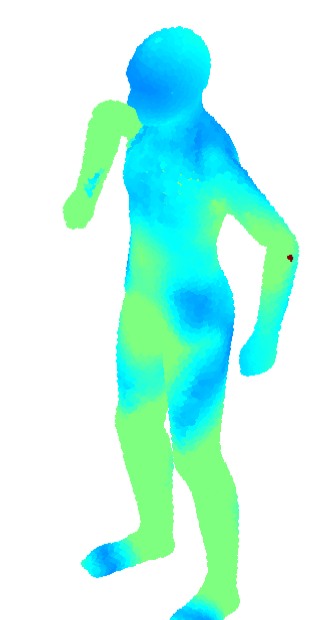}&
\includegraphics[trim=0cm 1.8cm 8.0cm 1.8cm,clip=true,height=0.205\linewidth]{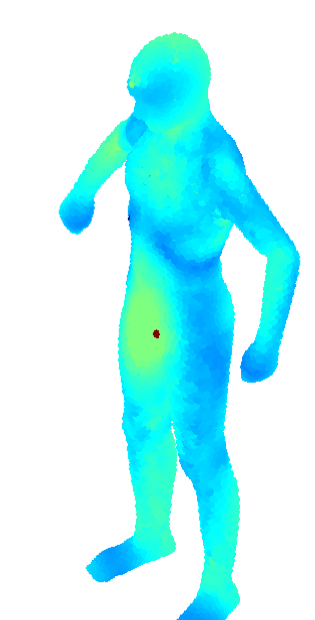}&
\includegraphics[trim=0cm 1.8cm 8.0cm 1.8cm,clip=true,height=0.205\linewidth]{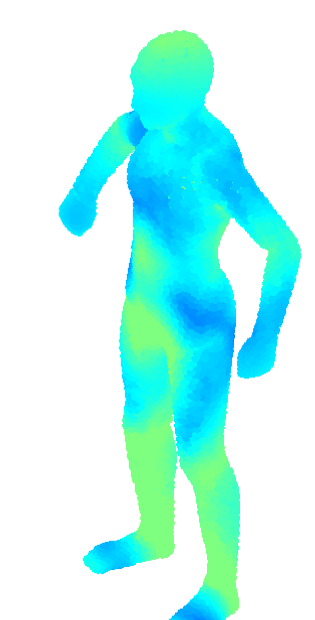}\\[5pt]
&\multicolumn{2}{c}{\footnotesize front/back scans}
&&
\footnotesize \jain~\cite{Jain:2010:MovieReshape}&
\footnotesize \ours&
\footnotesize \ours&
\footnotesize \ours\\
&&
&&
&
&
\footnotesize +\WSX&
\footnotesize +\NH\\
\end{tabular}
\caption{Per-vertex shape fitting error (mm) of multiple methods on
  sample individuals from the dataset of Helten et al.~\cite{HeltenPAE13}.}
  \label{fig:kinect-pose}
\egroup
\end{figure}

We use a publicly available dataset~\cite{HeltenPAE13} containing
Kinect body scans of three males and three females. Examples of the
Kinect scans are shown in Fig.~\ref{fig:kinect-pose}(a). For each
subject, a high-resolution laser scan was captured to determine
``ground truth'' body shape. Following the evaluation protocol of
Helten et al.~\cite{HeltenPAE13} we first fit a shape space to the
depth data, then fit shape space to the ground truth scan, and finally
compute the fitting error as a vertex-to-vertex Euclidean distance
between the vertices of the depth-fitted mesh and the ground
truth-fitted mesh. As the required landmarks are not available for
this dataset, we manually placed $14$ landmarks on each depth and
laser scan.

\subsection{Quantitative evaluation} 
For quantitative evaluation, we compare the following four shape
models presented above: the current state-of-the-art shape
space~\cite{Jain:2010:MovieReshape}, our shape space without posture
normalization and with posture normalization using \WSX~and \NH. In
our experiments, we vary the number of shape space parameters and the
number of training samples. To evaluate the fitting accuracy, we
report the proportion of vertices below a certain threshold.

\begin{figure}[t]
\centering
%% %\bgroup
\tabcolsep 1.5pt
\renewcommand{\arraystretch}{0.2}
\begin{tabular}{ccccc}
\# train &\multicolumn{4}{c}{\# PCA coefficients}\\
 sampl. &\\

  & 20 & 30 & 40 & 50 \\
\begin{sideways} \quad\quad all \end{sideways}
&
  \includegraphics[width=0.22\linewidth]{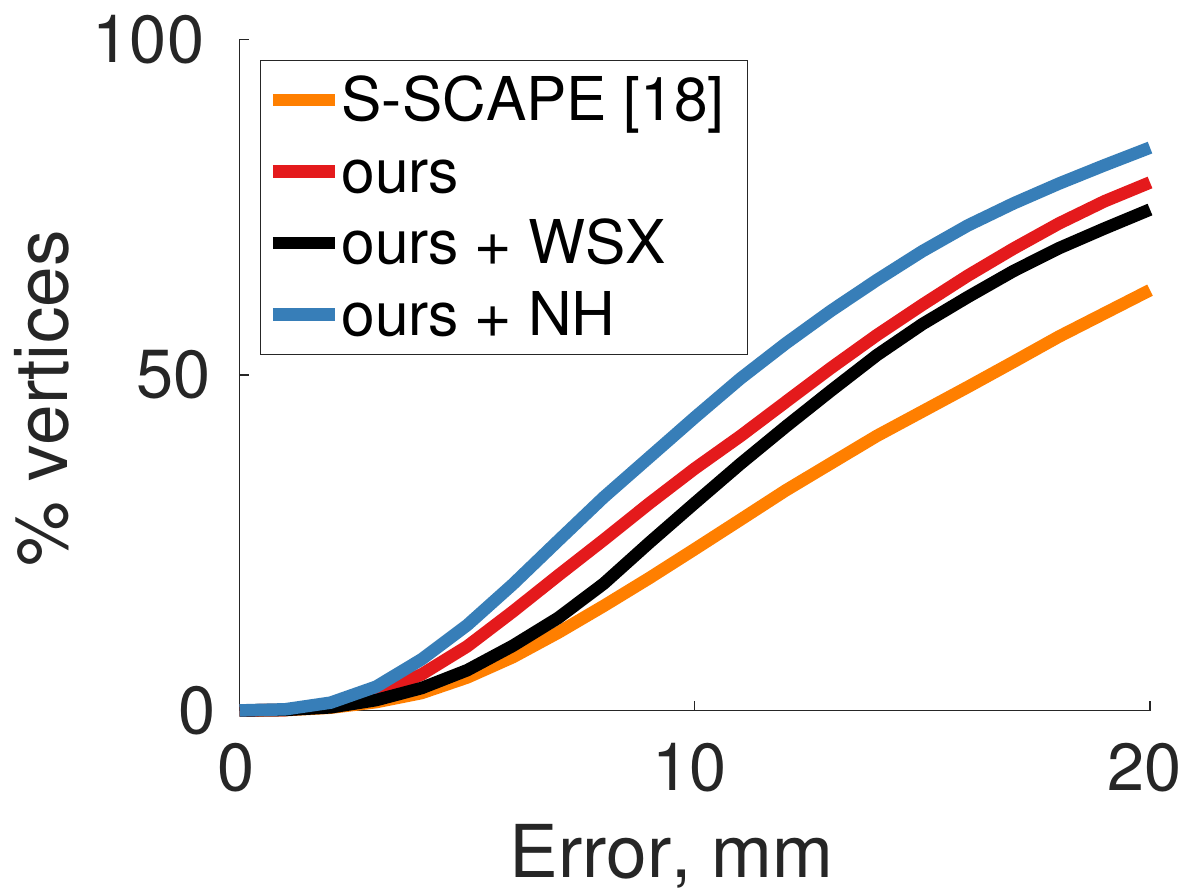}
&
\includegraphics[width=0.22\linewidth]{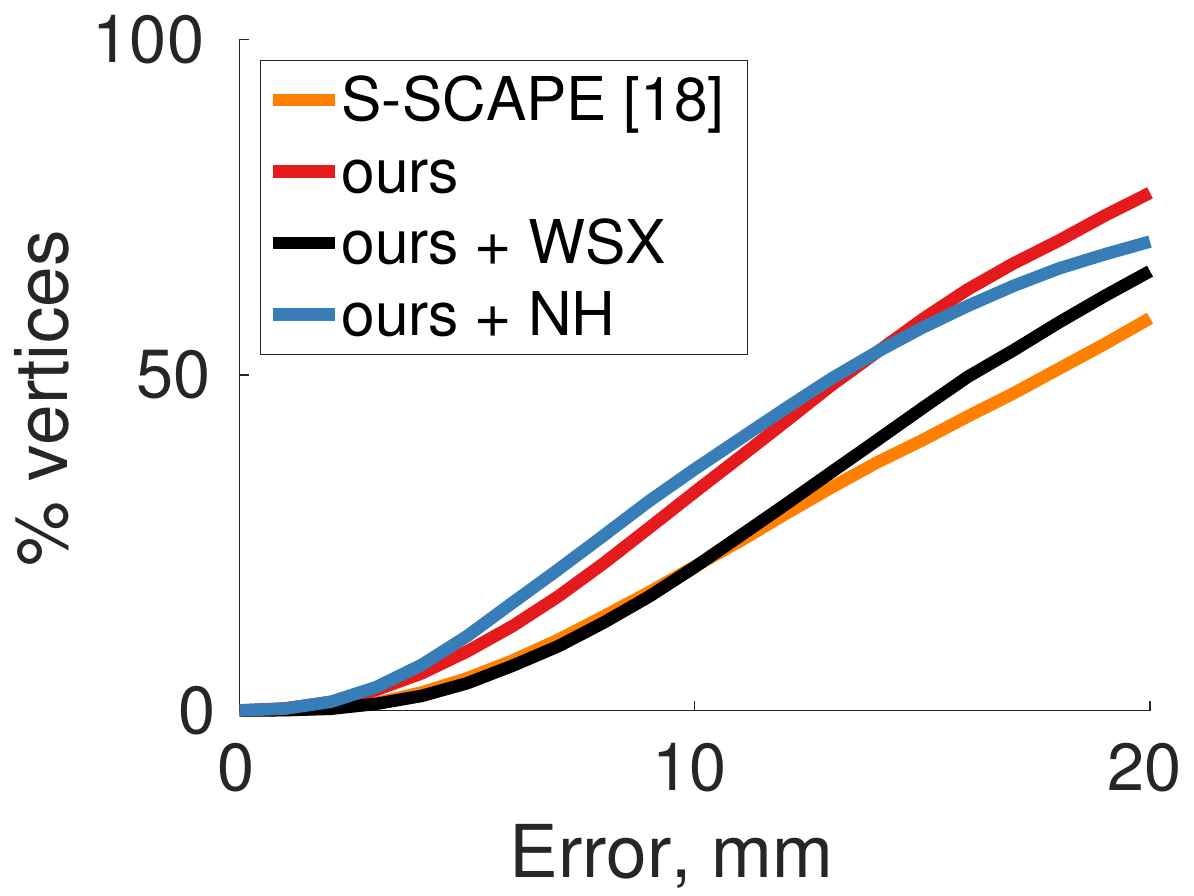}
&
\includegraphics[width=0.22\linewidth]{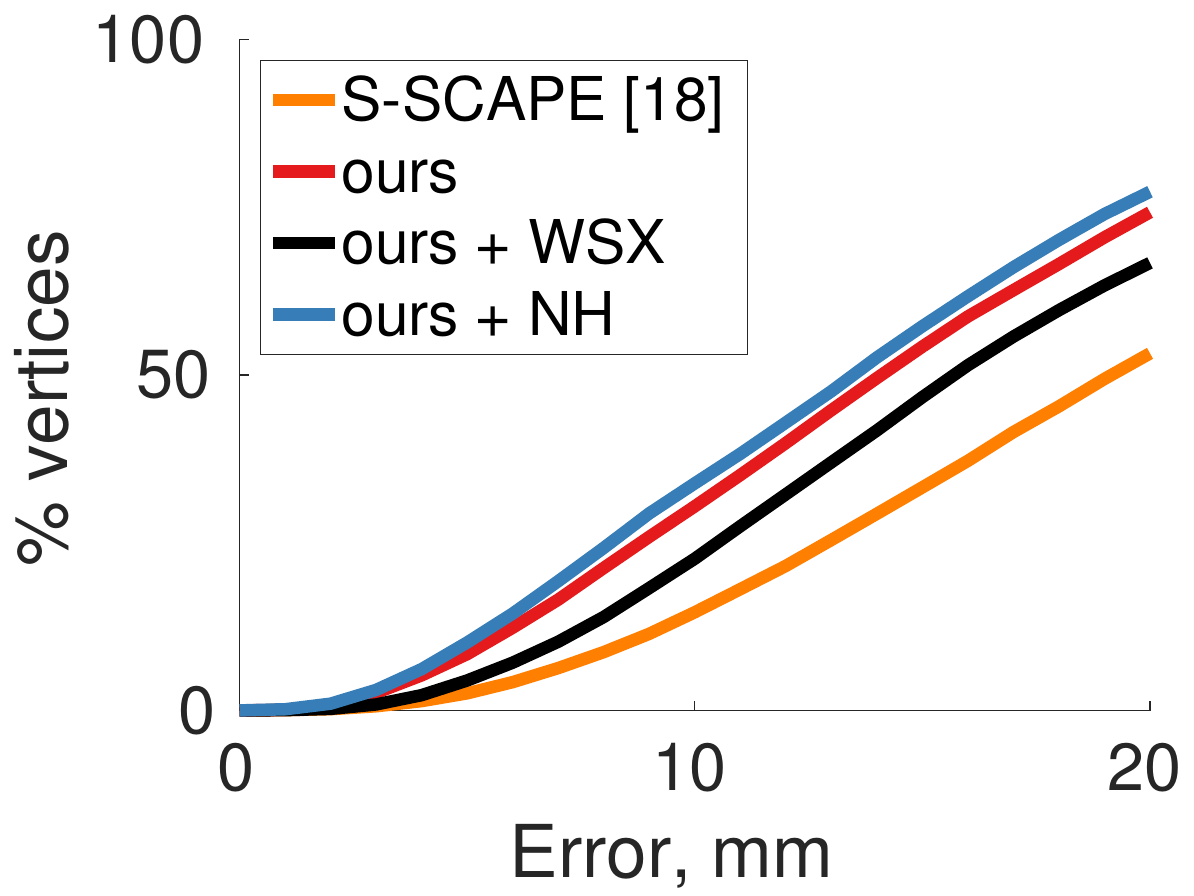}
&
\includegraphics[width=0.22\linewidth]{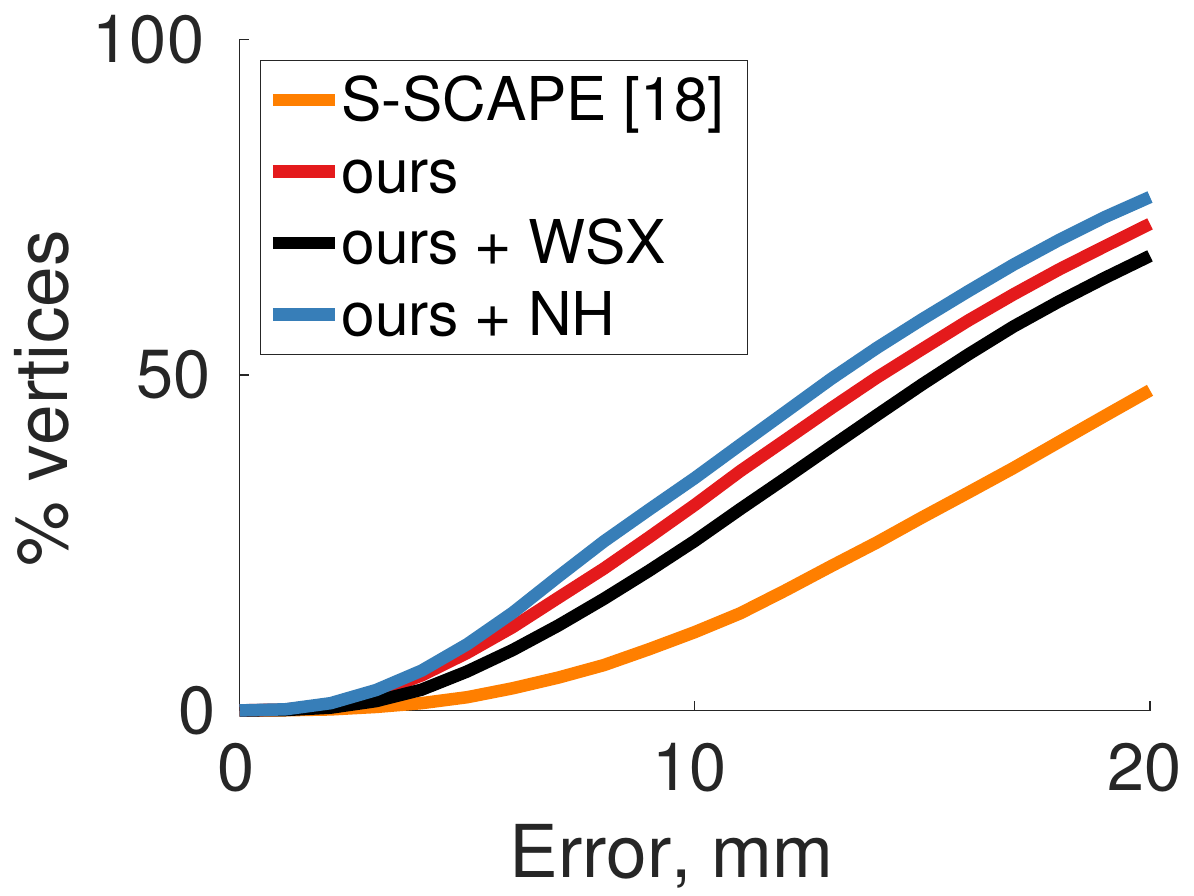}\\[2pt]

\begin{sideways} \quad\quad 1000 \end{sideways}
&
  \includegraphics[width=0.22\linewidth]{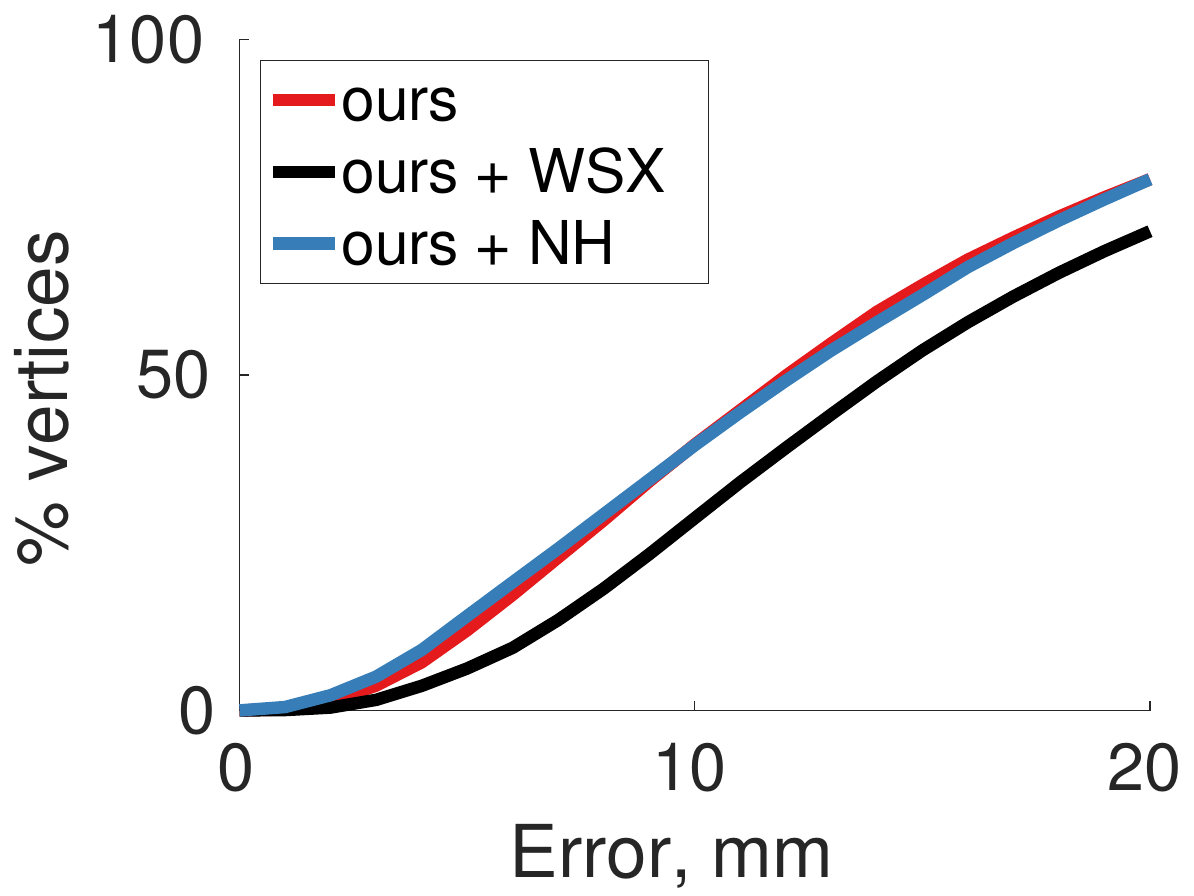}
&
\includegraphics[width=0.22\linewidth]{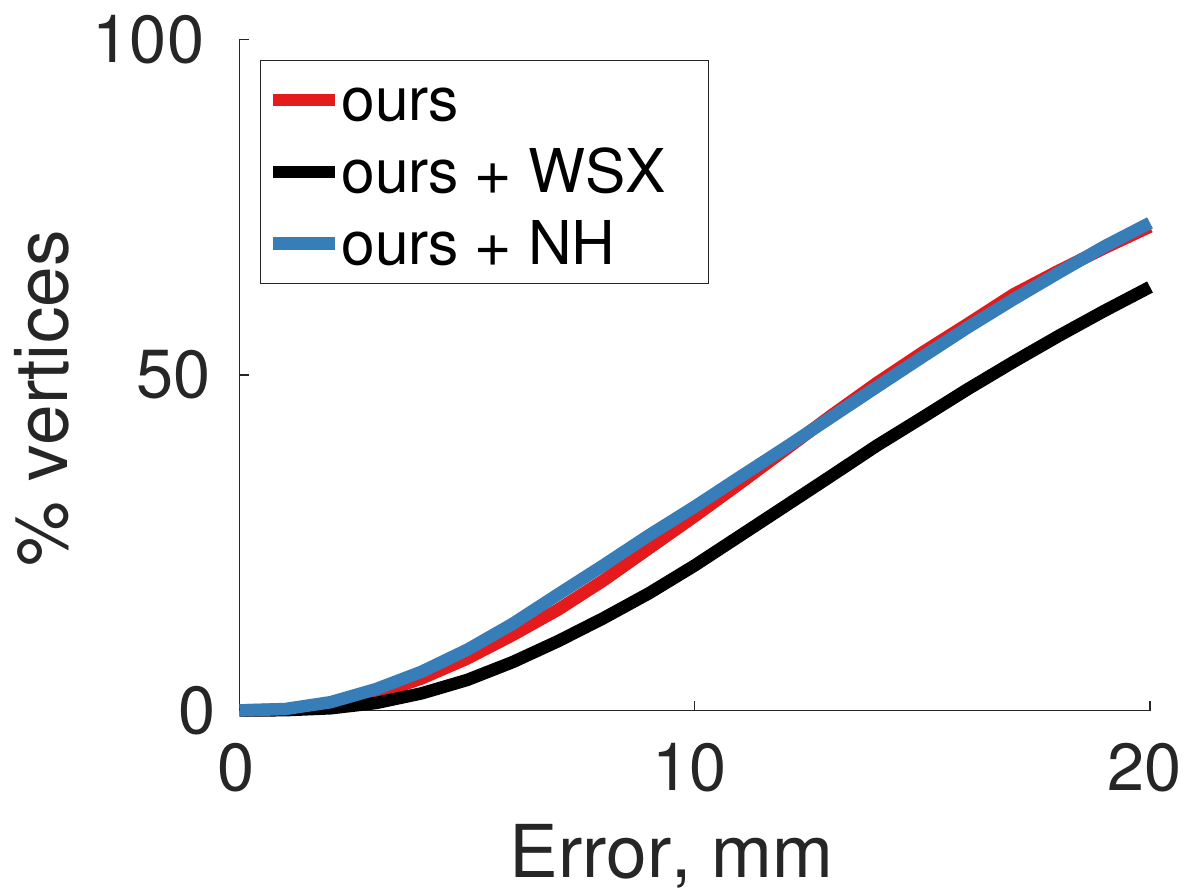}
&
\includegraphics[width=0.22\linewidth]{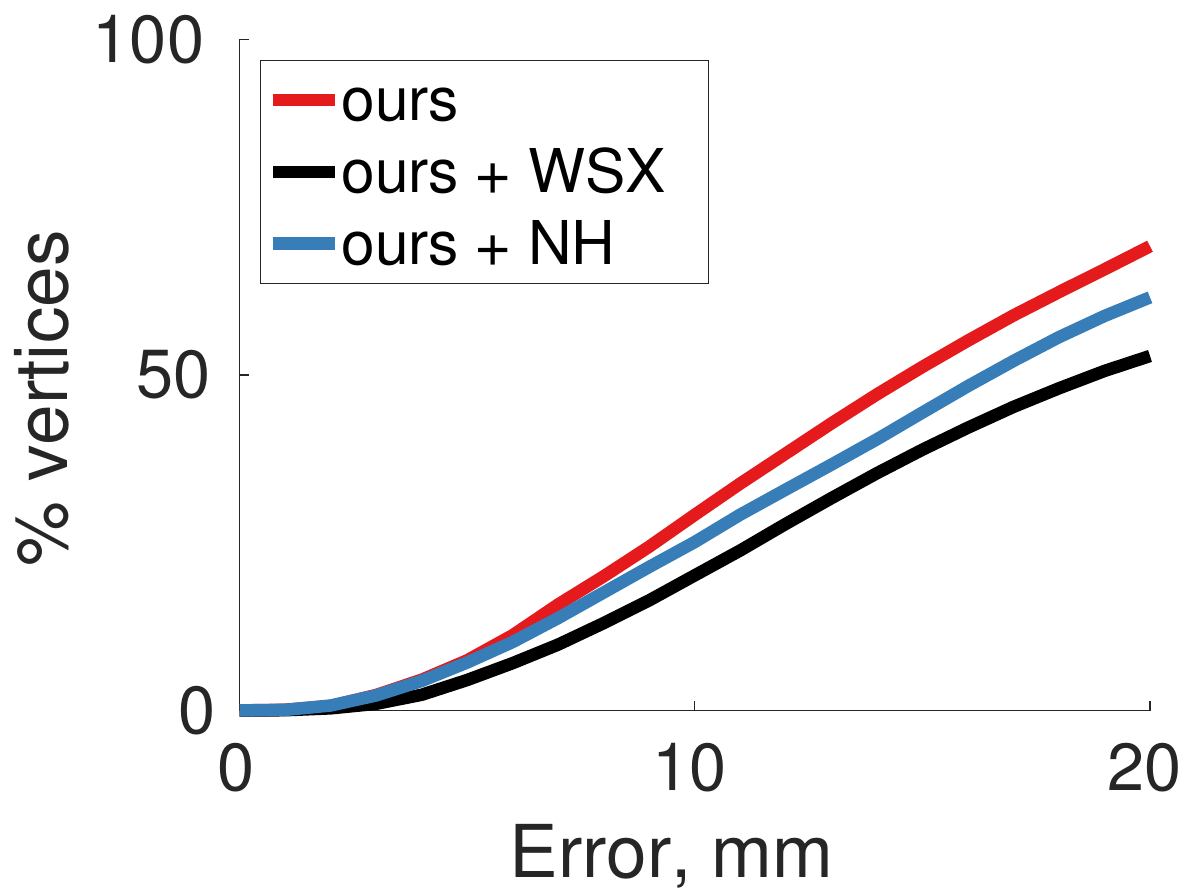}
&
\includegraphics[width=0.22\linewidth]{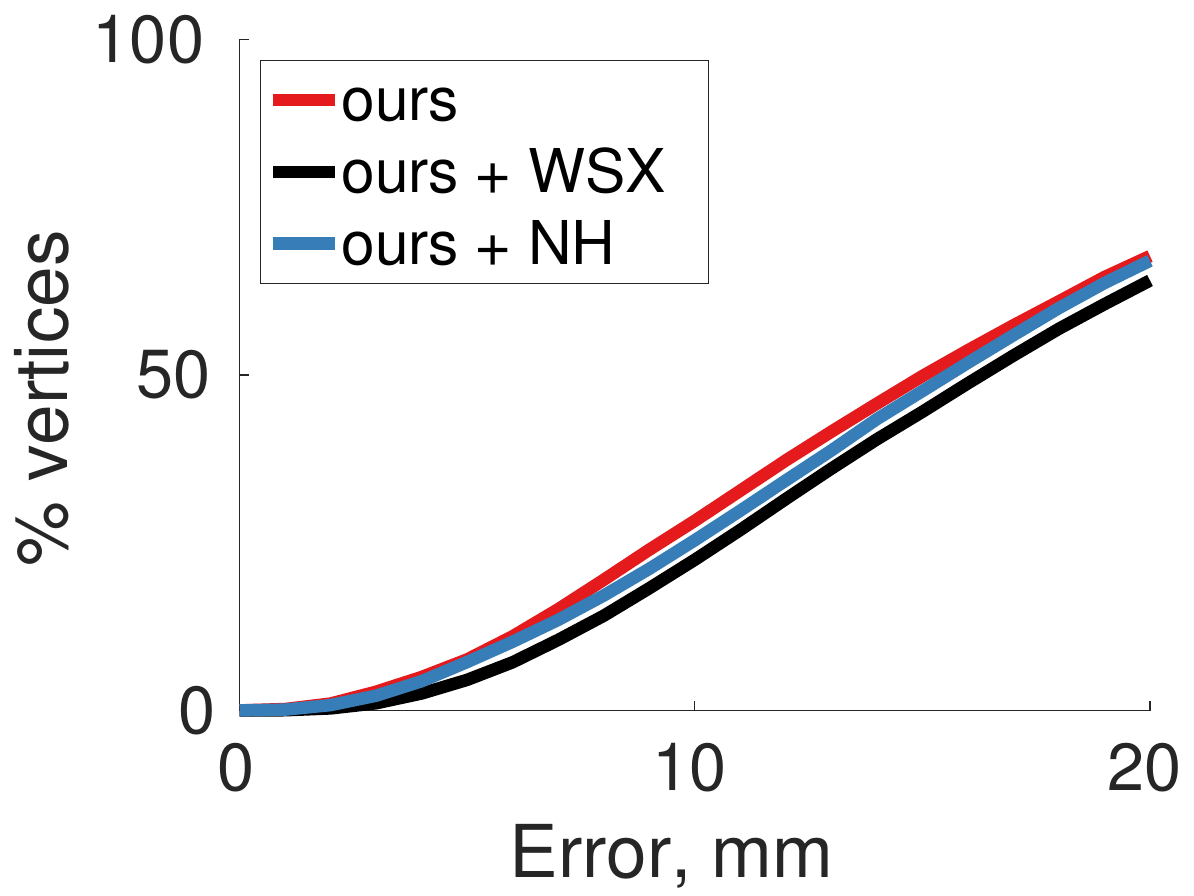}\\[2pt]

\begin{sideways} \quad\quad 100 \end{sideways}
&
\includegraphics[width=0.22\linewidth]{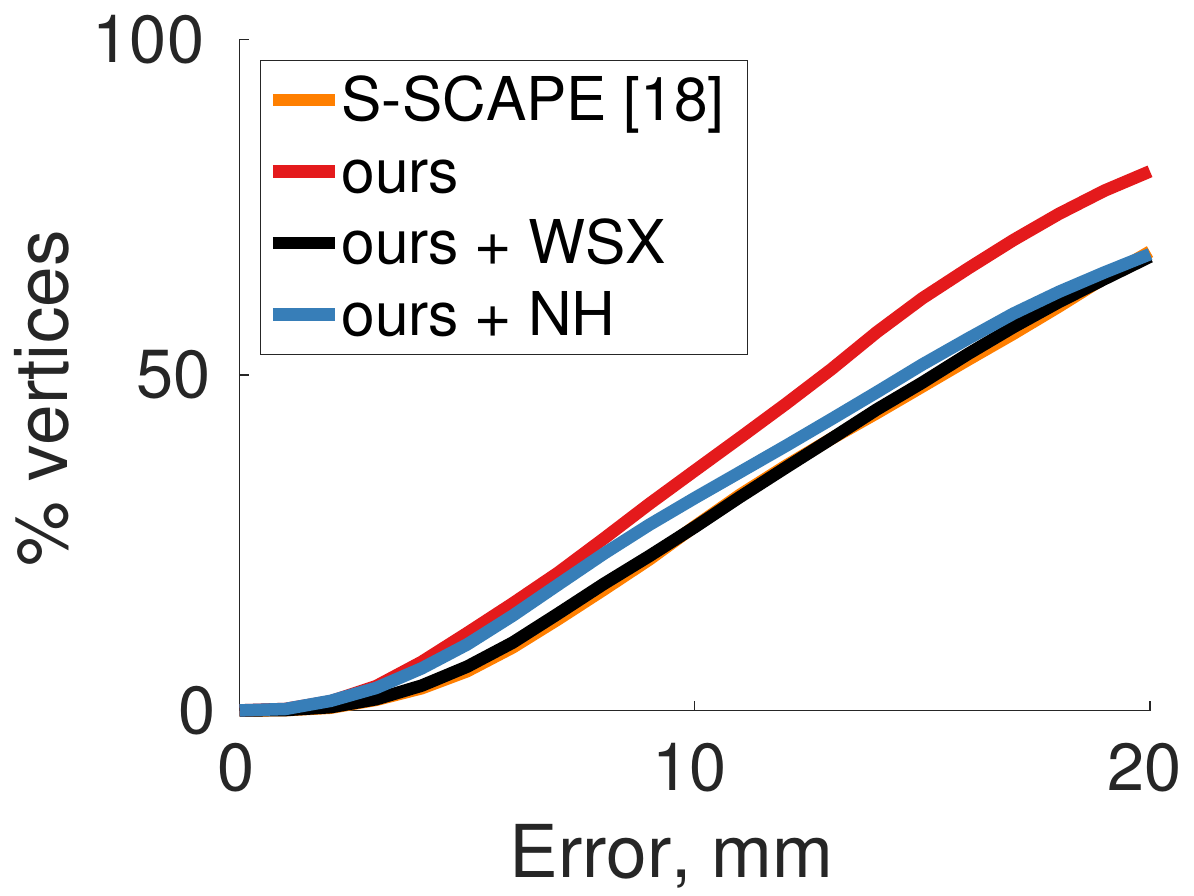}
&
\includegraphics[width=0.22\linewidth]{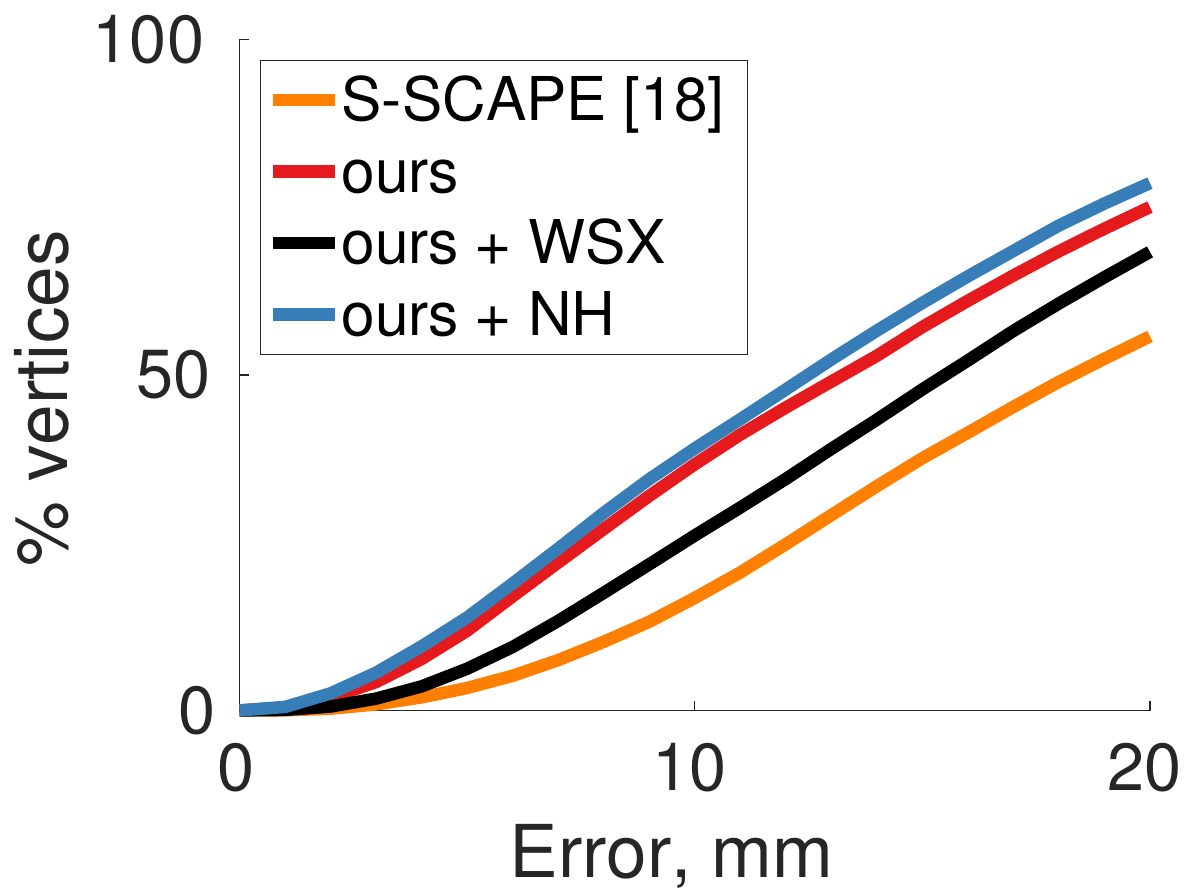}
&
\includegraphics[width=0.22\linewidth]{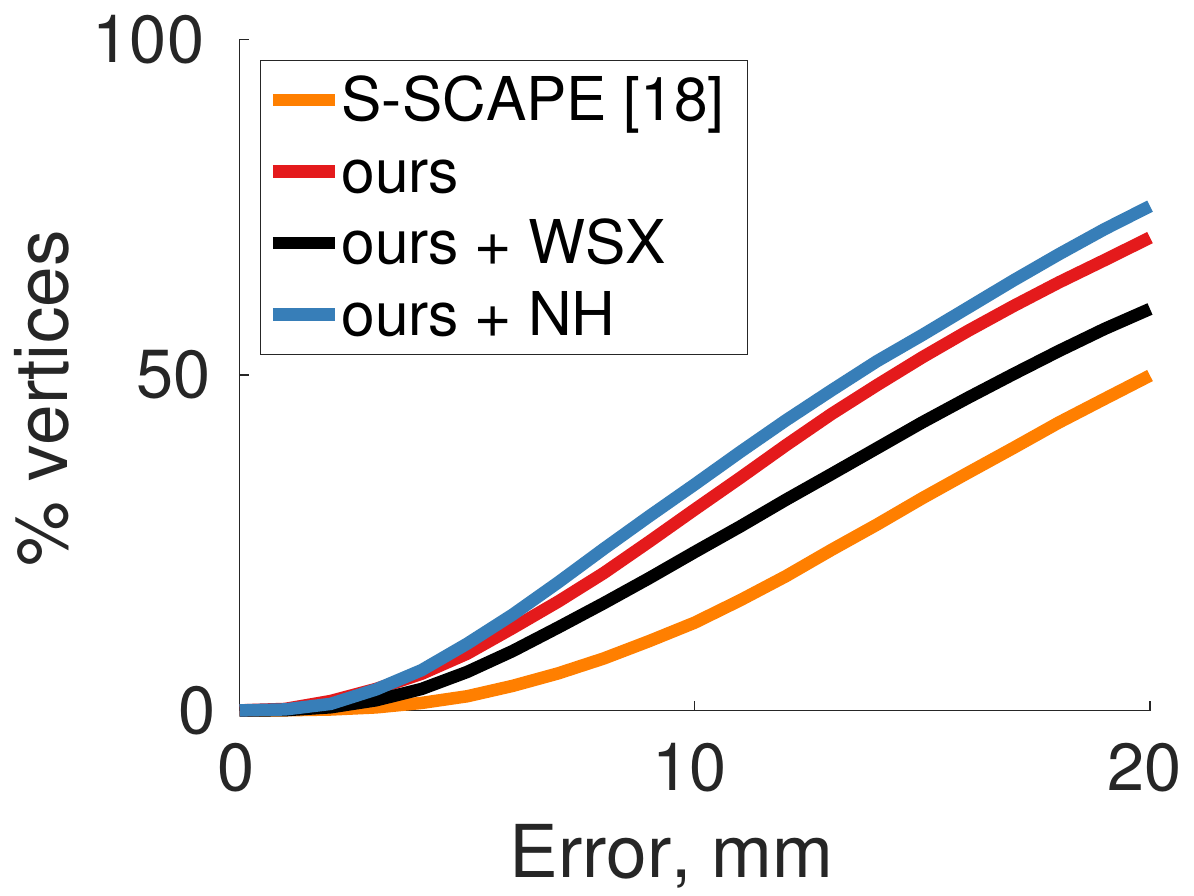}
&
\includegraphics[width=0.22\linewidth]{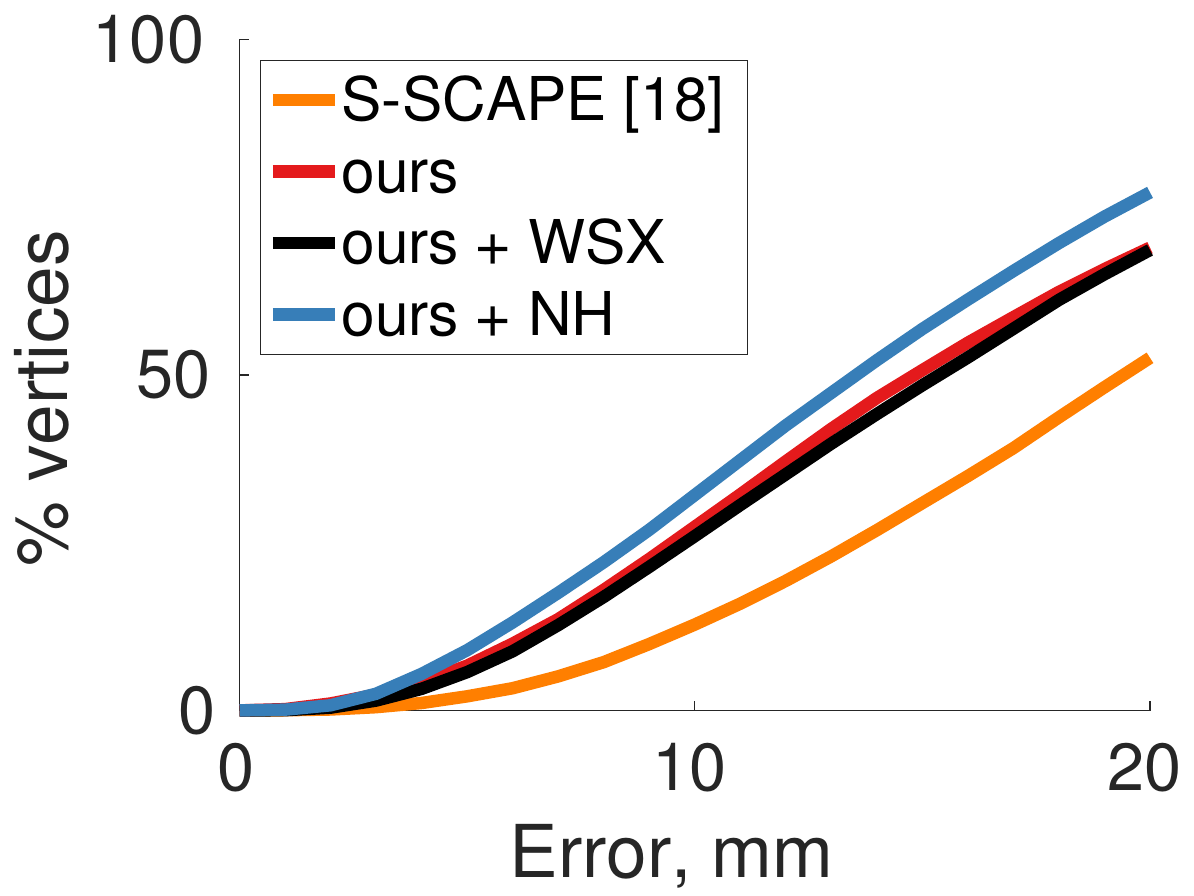}\\[2pt]

\begin{sideways} \quad\quad 50 \end{sideways}
&
\includegraphics[width=0.22\linewidth]{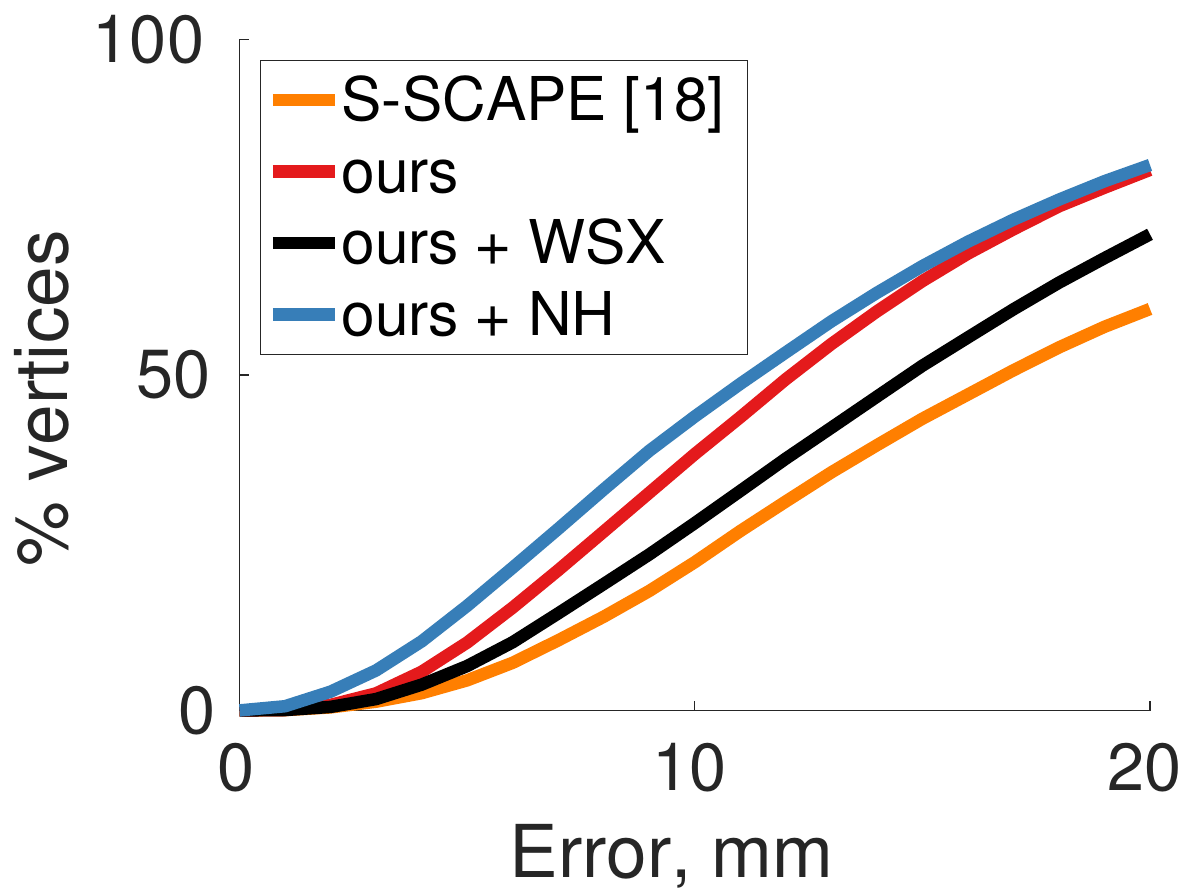}
&
\includegraphics[width=0.22\linewidth]{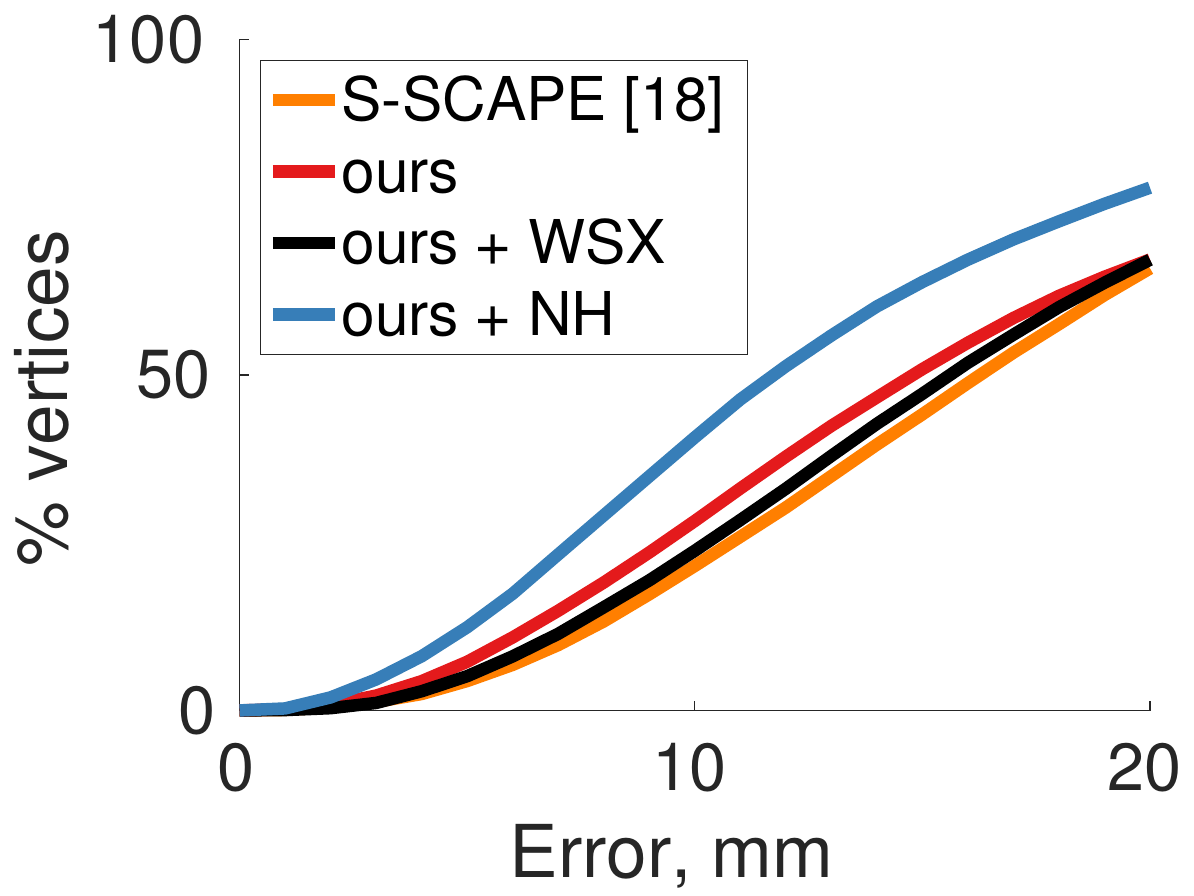}
&
\includegraphics[width=0.22\linewidth]{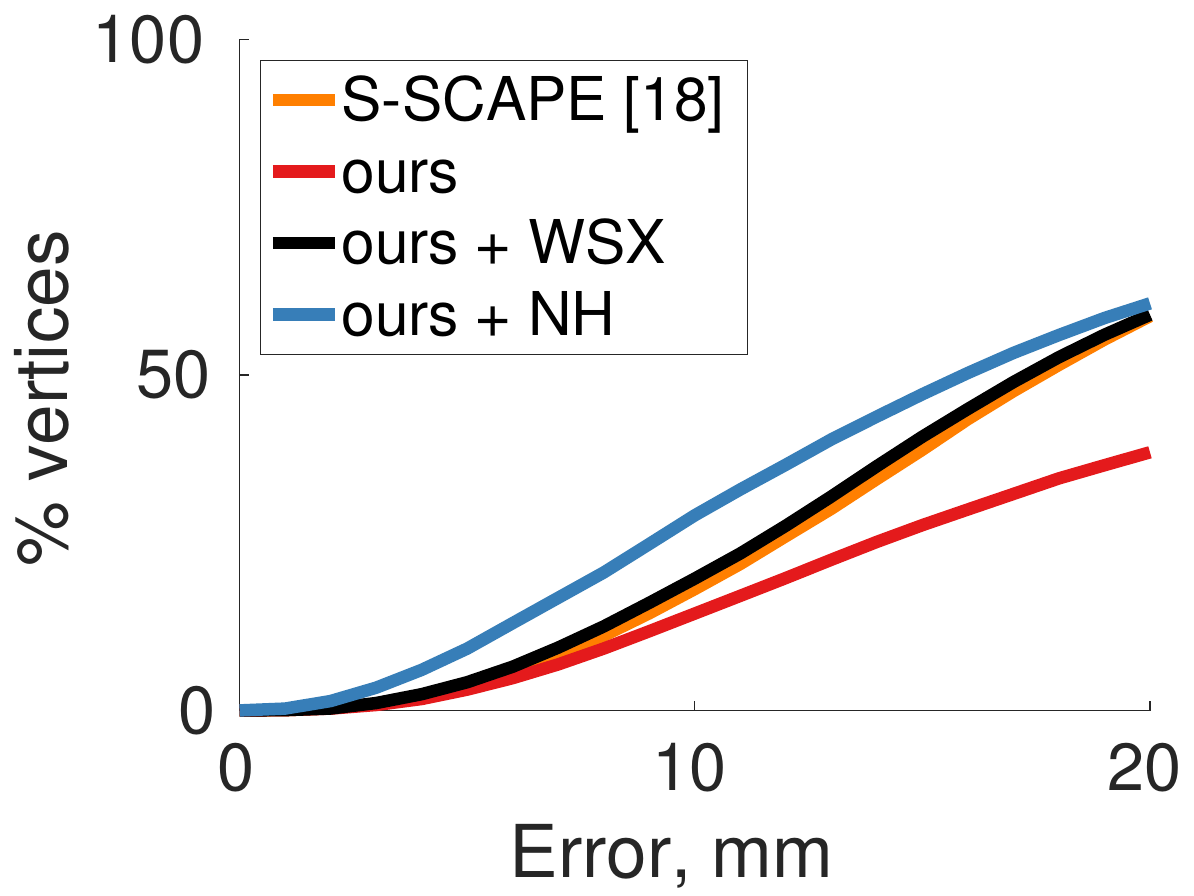}\\[2pt]
\end{tabular}
%}
\caption{Fitting error on the dataset of depth
  scans~\cite{HeltenPAE13} of \jain~spaces by Jain et
  al.~\cite{Jain:2010:MovieReshape} and \jain~spaces trained using our
  processed data without and with posture normalization using WSX and
  NH. Shown is the proportion of vertices [\%] for which the fitting
  error falls below a threshold.}
\label{fig:kinect-dist-all}
%% %\vspace{-0.5em}
%% %\egroup
\end{figure}

The results are shown in Fig.~\ref{fig:kinect-dist-all}, where the
number of shape space parameters varies in the columns and the number
of training samples varies in the rows. In all cases our \jain~spaces
learned from the CAESAR dataset significantly outperform the shape
space by Jain et al., which is learned from the far less
representative MPI Human Shape dataset. Our models achieve good
fitting accuracy when using as few as $20$ shape parameters, and the
performance stays stable when increasing the number of shape
parameters up to $50$ (first row). In contrast, the performance of the
shape space by Jain et al. drops, possibly due to overfitting to
unrealistic shape deformations in noisy depth data. Interestingly,
better performance by our models is evident even in the case when all
models are learned from the same number of training samples (third and
fourth rows). This shows that the CAESAR data has higher shape
variability than the MPI Human Shape data. In the majority of cases,
the shape space learned from the posture-normalized samples with
\NH~outperforms the shape space learned from samples without posture
normalization. This shows that the posture normalization method of
Neophytou and Hilton~\cite{Neophytou2013} helps to improve the
accuracy of fitting to noisy depth data. Surprisingly, the shape space
learned from samples without posture normalization outperforms the
shape space learned from the posture-normalized samples with \WSX~in
most cases. %\todo{Why?}
Overall, the quantitative results show the advantages of our approach
of building \jain~spaces learned from a large representative set of
training samples with additional posture normalization.

\subsection{Qualitative evaluation} 
To qualitatively evaluate the fitting, we visualize the per-vertex
fitting errors. We consider the \jain~spaces learned from
all available training samples and use $20$ shape space
parameters. For visualization we choose two subjects, male and female,
where the differences among the shape spaces are most pronounced.

Results are shown in Fig.~\ref{fig:kinect-pose}. Our shape spaces
better fit the data, in particular in the areas of belly and
chest. This is to be expected, as we learn from the larger and more
representative CAESAR dataset. Both shape spaces trained from posture
normalized models can better fit the arms compared to non-normalized
models.

\section{Conclusion}
In this work we address the challenging problem of building an
efficient and expressive 3D body shape space from the largest
commercially available 3D body scan dataset~\cite{RobinetteTCP99}. We
carefully design and evaluate different data preprocessing steps
required to obtain high-quality body shape models. To that end we
evaluate different template fitting procedures. We observe that shape
and posture fitting of an initial shape space to a scan prior to
non-rigid deformation considerably improves the fitting results. Our
findings indicate that multiple passes over the dataset improve
initialization and thus increase the overall fitting accuracy and
statistical shape space qualities. Furthermore, we show that posture
normalization prior to learning a shape space leads to significantly
better generalization and specificity of the \jain~spaces. Finally, we
demonstrate the advantages of our learned shape spaces over the
state-of-the-art shape space of Jain et
al.~\cite{Jain:2010:MovieReshape} learned on largest publicly
available dataset~\cite{HaslerBM2009} on the task of human body
shape reconstruction from noisy depth data.

We release our \jain~spaces, registered CAESAR scans, raw scan
preprocessing code, code to fit a \jain~space to a raw scan and
evaluation code for public usage\footnote{Available at
  \emph{\url{http://humanshape.mpi-inf.mpg.de}}}. We believe this
contribution is required for future development in human body
modeling.

%For future work, we plan to further explore how to compute correspondences and statistical models jointly. A recent approach by Hirshberg et al.~\cite{Hirshberg2012} takes a first step in this direction by treating the problems of learning a SCAPE model and computing point-to-point correspondences between a set of training data simultaneously by solving a single variational problem.

\section*{Acknowledgements}

We thank Alexandros Neophytou and Adrian Hilton for sharing their
posture normalization code. We also thank M\'onica Vidriales and
Gautham Adithya for their contributions to model fitting and
evaluation code. This work was partially funded by the Cluster of
Excellence \emph{MMCI}.

{\small
\bibliographystyle{plain}
\bibliography{refs-short}
}

\end{document}